\documentclass{article}

\PassOptionsToPackage{numbers, compress}{natbib}
\usepackage[final]{neurips_2022}

\usepackage[intlimits]{amsmath}
\usepackage{mathtools}
\usepackage{amsmath}
\usepackage{dsfont}
\usepackage{stmaryrd}
\usepackage{comment}
\usepackage{setspace}

\usepackage[unicode,psdextra]{hyperref}

\usepackage[capitalize,noabbrev]{cleveref}

\usepackage{algorithm}
\usepackage{algorithmic}
\usepackage{amssymb}
\usepackage{amsfonts}
\usepackage{amsthm}
\usepackage{graphics}
\usepackage{graphicx}
\usepackage{setspace}
\usepackage{subcaption}

\providecommand{\definitionname}{Definition}
\providecommand{\examplename}{Example}
\providecommand{\lemmaname}{Lemma}
\providecommand{\corrolaryname}{Corollary}
\providecommand{\propositionname}{Proposition}
\providecommand{\conditionsname}{Conditions}
\providecommand{\theoremname}{Theorem}
\providecommand{\assumptionname}{Assumption}

\usepackage{xcolor}

\usepackage{url}            
\usepackage{booktabs}       
\usepackage{amsfonts}       
\usepackage{nicefrac}       
\usepackage{microtype}      
\usepackage{xcolor}         
\usepackage{graphicx}
\usepackage{colortbl}
\usepackage{blindtext}
\usepackage{multicol}

\usepackage[T1,T2A]{fontenc}
\usepackage[utf8]{inputenc}
\usepackage[russian,english]{babel}

\title{Authorship attribution for Differences between Literary Texts by Bilingual Russian-French and Non-Bilingual French Authors}

\author{%
  Margarita Makarova \\
  SLAS, University of Lausanne,\\
  Lausanne, Switzerland\\
  \texttt{margarita.makarova@unil.ch} \\
}

\begin{document}
\maketitle
\begin{abstract}
    Do bilingual Russian-French authors of the end of the twentieth century such as Andreï Makine, Valéry Afanassiev, Vladimir Fédorovski, Iegor Gran, Luba Jurgenson have common stylistic traits in the novels they wrote in French? Can we distinguish between them and non-bilingual French writers' texts? Is the phenomenon of interference observable in French texts of Russian authors? This paper applies authorship attribution methods including Support Vector Machine (SVM), $K$-Nearest Neighbors (KNN), Ridge classification, and Neural Network to answer these questions. 
\end{abstract}

\section{Introduction}
\label{sec:Introduction}
A number of French literary works written by authors of Russian origins have been published in the twentieth century but have not been given much scholarly attention. There are studies on Russian-French bilingualism, but none of them applied authorship attribution to detect ``traces of Russian language'' in French texts or to find differences between texts by bilingual and non-bilingual authors. This article aims to answer two questions. Firstly, are there differences between bilingual and non-bilingual French writers’ texts? Secondly, is it possible to find traces of Russian language in literary texts written by bilingual Russian-French writers?  

The expression ``traces of Russian language'' can be replaced with the term ``interference'', that was defined by American linguist Uriel Weinreich in regard to bilingual non-literary speech, as follows:

\begin{quote}
Those instances of deviation from the norms of either language which occur in the speech of bilinguals as a result of their familiarity with more than one language, i. e. as a result of language contact, will be referred to as INTERFERENCE phenomena. […].
The term interference implies the rearrangement of patterns that result from the introduction of foreign elements into the more highly structured domains of language, such as the bulk of the phonemic system, a large part of the morphology and syntax, and some areas of the vocabulary (kinship, color, weather, etc.) (\cite{weinreich1970}:1).
\end{quote}

In this paper, interference is understood to mean a change in literary speech in French (writing language of Russian-French group of authors) under the influence of elements of Russian (their native language) on the phonetic, lexical, morphological, grammatical, and/or syntactic levels. Our computational approach focuses first of all on morphological and lexical aspects. We differentiate between interference and ``calque'', which represents an exact translation of a lexeme, an idiom, or a phrase.

The corpus consists of the following texts in French:

A. ``Russian French'' bilingual writers: {\itshape Confession d'un porte-drapeau déchu}, {\itshape Au temps du fleuve Amour}, {\itshape Le Testament français}, {\itshape Le Crime d'Olga Arbélina}, {\itshape Requiem pour l'Est} by Andreï Makine, {\itshape Lettres sonores} by Valéry Afanassiev, {\itshape Les Deux sœurs} by Vladimir Fédorovski, {\itshape Éducation nocturne} by Luba Jurgenson, {\itshape Acné festival} by Iegor Gran. They were all written between 1990 and 2000.

B. ``French-French'' classic writers: {\itshape Le Côté de Guermantes (I)} by Marcel Proust, {\itshape Germinal} by Émile Zola, {\itshape Bouvard et Pécuchet} by Gustave Flaubert, {\itshape La Maison du chat-qui-pelote} by Honoré de Balzac (19th-early 20th centuries).

C. ``French-French'' contemporary writers: {\itshape Du plus loin de l’oubli} by Patrick Modiano, {\itshape Terrasse à Rome} by Pascal Quignard, {\itshape L’Amant de la Chine du Nord} by Marguerite Duras, {\itshape La Honte}, {\itshape L’Événement} by Annie Ernaux, {\itshape Les Particules élémentaires} by Michel Houellebecq. They were published between 1990–2000, as the texts of group A, that is why we prefer ``contemporary'' to ``modern''.

The choice of Makine's novels from group A is due to the fact that they are analyzed by K. Baleevskikh in her thesis. There are few studies about Russian-French literary bilingualism with a textual analysis. Baleevskikh's thesis, in addition to a thesis by M. \cite{shakhnovitch} on Elsa Triolet’s novels, are the most comprehensive. Baleevskikh finds in Makine’s books the frequency of subject-predicate inversions, impersonal constructions, nominative sentences, dashes, words ``soul'', ``fate'', ``friend'', conjunction ``and'' in the beginning of the sentence, words expressing surprise (``soudain'', ``brusquement'', etc.), suffixes, diminutives, adverbs ending in ``ment'', etc. (\cite{baleevskikh}: 160-165). According to Baleevskikh, these traits are characteristic of Makine as a Russian native speaker. The texts by other authors belong to the same time period (1990-2000). French writers of Russian origins of the same period could have a common background (birth, in some cases study and work in the USSR, emigration, or career in literature) and therefore form a kind of community. Among the features we noted while reading the novels in group A, the most common are suffixes of adjectives denoting colors, diminutives, repetitions, in some cases inversions, and calques.

As for the second group, which includes texts by Proust, Zola, Flaubert, and Balzac, their choice is related to the fact that researchers have noted similarities between Makine's style and that of Proust (\cite{shishkina}, \cite{baleevskikh}, \cite{mccall}, \cite{rubins}, \cite{wanner}, \cite{clement}, \cite{jongeneel}, \cite{ausoni}), and to the fact that Makine makes references to Proust, Balzac, and Flaubert in his novels (\cite{wanner}: 39). It is unknown whether other French writers of Russian origins of the 1990s share common features with Proust, Zola, Balzac, and Flaubert. At first glance, the language of the texts by Russian-French authors is bookish, referring to the ``classics'' of the nineteenth century. The styles of Makine, Fédorovski, Jurgenson, and Afanassiev seem more similar to those of the ``classics'' than to those of the French writers of the 1990s, that is, their contemporaries.

The third group includes the texts of 1990-2000 by French authors of French origins — Patrick Modiano, Pascal Quignard, Marguerite Duras, Annie Ernaux, and Michel Houellebecq. This group is compared with the Russian-French group. The main criterion for the selection of texts by French authors of French origins is their year of publication — the period chosen was 1990-2000. We include non-Russian-speaking authors. In addition, we include texts by Nobel Prize winners (Modiano, Ernaux), Goncourt Prize winners (Quignard, Duras, Houellebecq) and other French prizes (all authors) as guarantors of the ``correctness'' of the French language. The books of all the above mentioned contemporary prose writers, except for M. Houellebecq, were selected as the best books in French of the twentieth century by the readers of the Swiss newspaper {\itshape Le Temps} in 2019. The authors of the Russian-French group wrote more voluminous texts, referring rather to the French prose of the late nineteenth and early twentieth centuries. For this reason, there are more texts in this group than in previous groups.

Some novels are excluded because of their length, for example, Flaubert’s {\itshape Madame Bovary}, although it is more famous and representative of Flaubert than {\itshape Bouvard et Pécuchet}. We needed to balance the datasets.

Our research is conducted from a stylometric perspective. Stylometry refers to the process of classifying literary texts using statistical and machine learning methods in order to find similarities between the styles of different authors. Authorship attribution as a part of stylometry aims to find the real author of an anonymous text. All of the texts in our corpus are known. We select each of them, name it ``Disputed'' as if it were unknown, and use authorship attribution to find differences or similarities between groups.

\section{Methods}

Two authorship attribution approaches are used in our paper: \textit{the multivariate analysis approach} (the closest text has the smallest distance measure) and \textit{the machine learning approach} (\cite{koppeletal2009}: 10-11). Our goal is to test whether every text in the Russian-French group is more similar to other texts in the same group or to the texts in a French-French group. An additional goal is to detect interference, if possible. In order to do that, we use multivariate Chi-squared method and four machine learning supervised techniques: Ridge classification, Support Vector Machine (SVM), $K$-Nearest Neighbors (KNN) and Artificial Neural Network. In addition, we use Principal Component Analysis (PCA), an unsupervised technique that allows the visualization of the structure of the Neural Network inner layers.

The algorithm for Chi-squared is the following: 1) remove stop words; 2) form a train set (groups A and B or A and C) and a test set (Disputed); 3) remove punctuation; 4) find the most frequent 500 (or 100, 1000, etc.) words in both train set and test set; 5) calculate Chi-squared between Disputed and groups A and B or groups A and C. If the distance between Disputed and group A is smaller than the distance between Disputed and group B(C), then Disputed is classified in group A and vice versa. This indicates that the text Disputed is similar to the group to which it is attributed.

The algorithms for machine learning methods are the following: 1) remove stop words and punctuation; 2) do tokenizing; 3) find the most frequent 500 (or 100, 1000, etc.) words in both train set and test set; 4) divide each text in segments of 1700 words; 5) create train and test sets containing segments obtained at point 4; 6) for the train test define the target value that is 0 for group A and 1 for group B or C; 7) the test set targets are not revealed to the algorithm; 8) training the models (Ridge, SVM, KNN and Neural Network) using the data and the targets in the training set; 9) using the trained model, predict the unknown target of the test set.

\section{Experiments and Results}
\subsection{Chi-squared Method}

Test 1: Is the lexicon of Makine’s novels similar to the lexicon of group A writers or to the lexicon of group B writers? 
Train set: the novels of Afanassiev, Fédorovski, Jurgenson, Gran. Test set: the texts of Balzac, Flaubert, Zola. Each of Makine's novels becomes Disputed one by one. See Table \ref{tab:chi_squared_1}.

\begin{table}[h!]
\begin{tabular}{|l|c|c|c|c|}
\hline
\rowcolor[HTML]{C0C0C0} 
\multicolumn{1}{|c|}{\cellcolor[HTML]{C0C0C0}Disputed} & \begin{tabular}[c]{@{}c@{}}Russian-French\\ Chi-squared\end{tabular} & \begin{tabular}[c]{@{}c@{}}French-French\\ Chi-squared\\ (classic)\end{tabular} & True group & \multicolumn{1}{l|}{\cellcolor[HTML]{C0C0C0}\begin{tabular}[c]{@{}l@{}}Difference \\ between \\ groups in \%\end{tabular}} \\ \hline
\textit{Confession d’un port-drapeau déchu}            & {\color[HTML]{0076BA} \textbf{5273}}                                 & 6299                                                                            & Ru-Fr      & 16\%                                                                                                                       \\ \hline
\textit{Au temps du fleuve Amour}                      & {\color[HTML]{0076BA} \textbf{11154}}                                & 12990                                                                           & Ru-Fr      & 14\%                                                                                                                       \\ \hline
\textit{Le Testament français}                         & {\color[HTML]{0076BA} \textbf{11814}}                                & 15571                                                                           & Ru-Fr      & 24\%                                                                                                                       \\ \hline
\textit{Le Crime d’Olga Arbélina}                     & {\color[HTML]{0076BA} \textbf{12231}}                                & 15638                                                                           & Ru-Fr      & 22\%                                                                                                                       \\ \hline
\textit{Requiem pour l’Est}                            & {\color[HTML]{0076BA} \textbf{10693}}                                & 14246                                                                           & Ru-Fr      & 25\%                                                                                                                       \\ \hline
\end{tabular}
\caption{\textbf{Chi-squared score for Makine's novels (Disputed), group A, and group B.} \label{tab:chi_squared_1}}
\end{table}

Test 2: Is the lexicon of Makine’s novels similar to the lexicon of group A writers or to the lexicon of group C writers? 
Train set: the novels of Afanassiev, Fédorovski, Jurgenson, Gran. Test set: the novels of Modiano, Ernaux, Duras, Quignard, Houellebecq. Each of Makine's novels becomes Disputed one by one. See Table \ref{tab:chi_squared_2}.

\begin{table}[h!]
\begin{tabular}{|l|c|c|c|c|}
\hline
\rowcolor[HTML]{C0C0C0} 
\multicolumn{1}{|c|}{\cellcolor[HTML]{C0C0C0}Disputed} & \begin{tabular}[c]{@{}c@{}}Russian-French\\ Chi-squared\end{tabular} & \begin{tabular}[c]{@{}c@{}}French-French\\ Chi-squared\\ (contemp.)\end{tabular} & True group & \multicolumn{1}{l|}{\cellcolor[HTML]{C0C0C0}\begin{tabular}[c]{@{}l@{}}Difference \\ between \\ groups in \%\end{tabular}} \\ \hline
\textit{Confession d’un port-drapeau déchu}            & {\color[HTML]{0076BA} \textbf{5272}}                                 & 6359                                                                             & Ru-Fr      & 17\%                                                                                                                       \\ \hline
\textit{Au temps du fleuve Amour}                      & {\color[HTML]{0076BA} \textbf{11153}}                                & 12932                                                                            & Ru-Fr      & 14\%                                                                                                                       \\ \hline
\textit{Le Testament français}                         & {\color[HTML]{0076BA} \textbf{11814}}                                & 14890                                                                            & Ru-Fr      & 21\%                                                                                                                       \\ \hline
\textit{Le Crime d’Olga Arbélina}                     & {\color[HTML]{0076BA} \textbf{12230}}                                & 15013                                                                            & Ru-Fr      & 19\%                                                                                                                       \\ \hline
\textit{Requiem pour l’Est}                            & {\color[HTML]{0076BA} \textbf{10693}}                                & 14545                                                                            & Ru-Fr      & 26\%                                                                                                                       \\ \hline
\end{tabular}
\caption{\textbf{Chi-squared score for Makine's novels (Disputed), group A, and group C.} \label{tab:chi_squared_2}}
\end{table}

As a result, all of Makine’s novels in the corpus are closer to other Russian-French authors than to French-French authors. In the first test Makine’s novel {\itshape Au temps du fleuve Amour} is closer to the Russian-French group by 14\% and his novel {\itshape Requiem pour l'Est} by 25\%. In the second test Makine's {\itshape Au temps du fleuve Amour} is also closer to the Russian-French group by 14\% and his novel {\itshape Requiem pour l'Est} by 26\%. This indicates similarities between the texts by Makine, Afanassiev, Fédorovski, Jurgenson and Gran. 

It is unclear whether similarities are due to the fact that there is lexico-morphological interference in all Russian-French texts. We do not know which lexemes played a decisive role in obtaining the conclusion of similarity. Without answering the question which lexemes make the texts of the Russian-French authors similar to each other and without seeing whether they are subject to interference, at this stage we cannot claim that we have confirmed the existence of lexical and/or morphological interference in Makine’s texts using the Chi-square method.

\subsection{Machine Learning Methods}
\subsubsection{Ridge and SVM classifications}

\paragraph{Groups A and B}

Ridge allows us to make a binary classification taking into account the weight of each word. We first compare groups A and B. We consider the content words, therefore, the text segments are compiled in a way that avoids function words.

\begin{figure}
\advance\leftskip-1cm
\begin{tabular}{ccccc}
\subfloat[range $500$ - $600$]{%
       \includegraphics[scale=0.2]{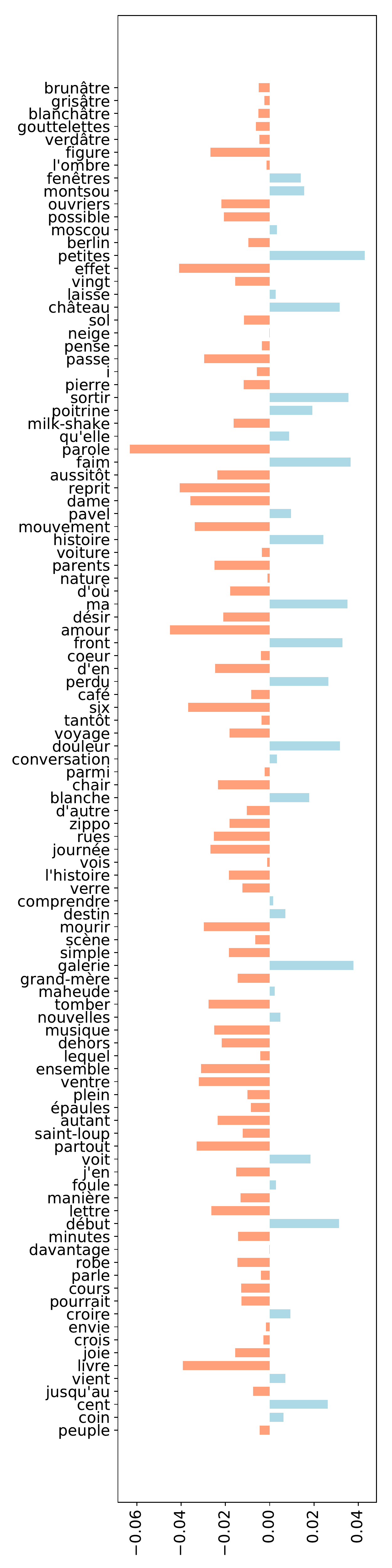}
     } &
\subfloat[range $600$ - $700$]{%
       \includegraphics[scale=0.2]{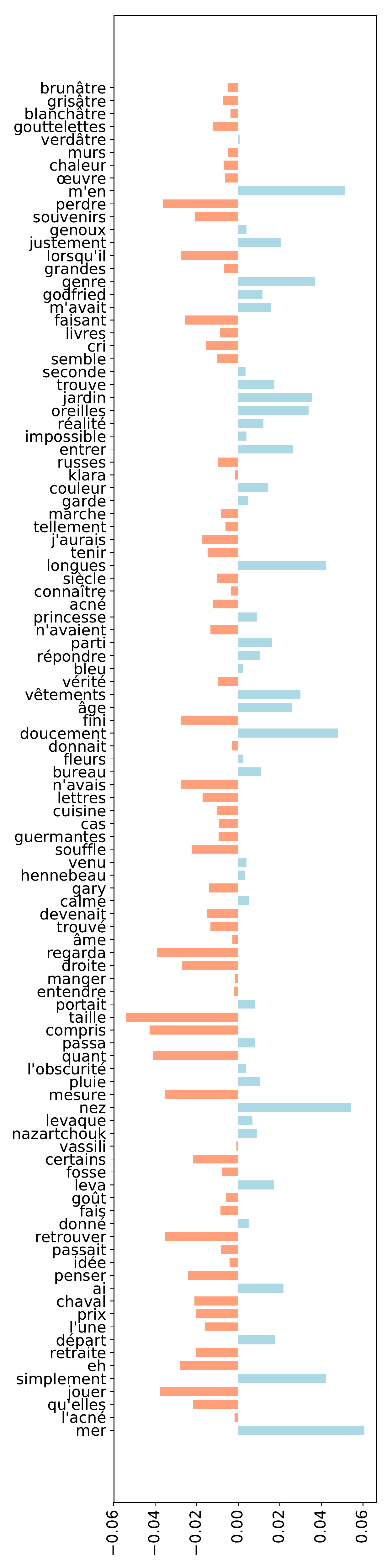}
     }  &
\subfloat[range $700$ - $800$]{%
       \includegraphics[scale=0.2]{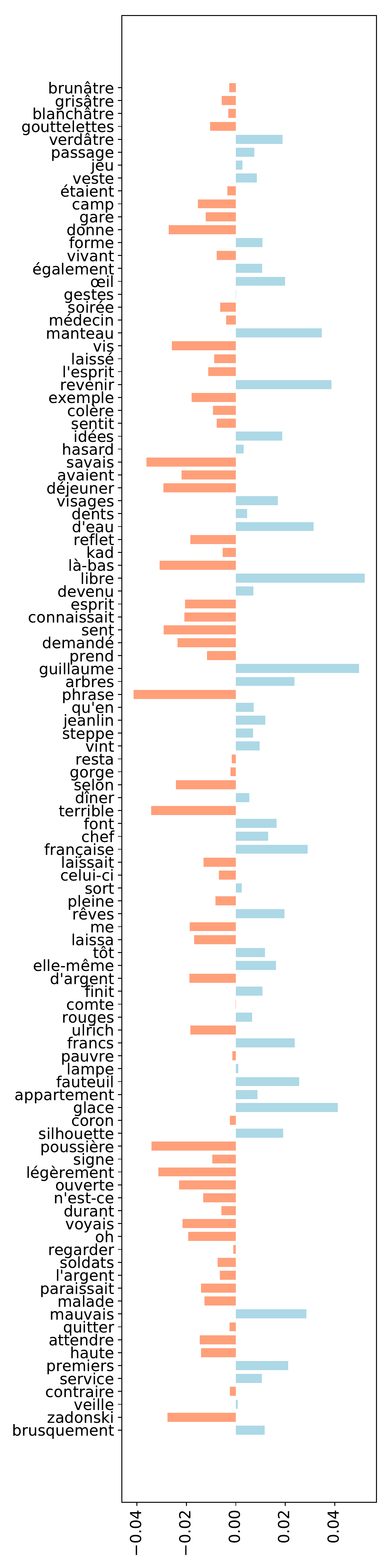}
     }  &
\subfloat[range $800$ - $900$]{%
       \includegraphics[scale=0.2]{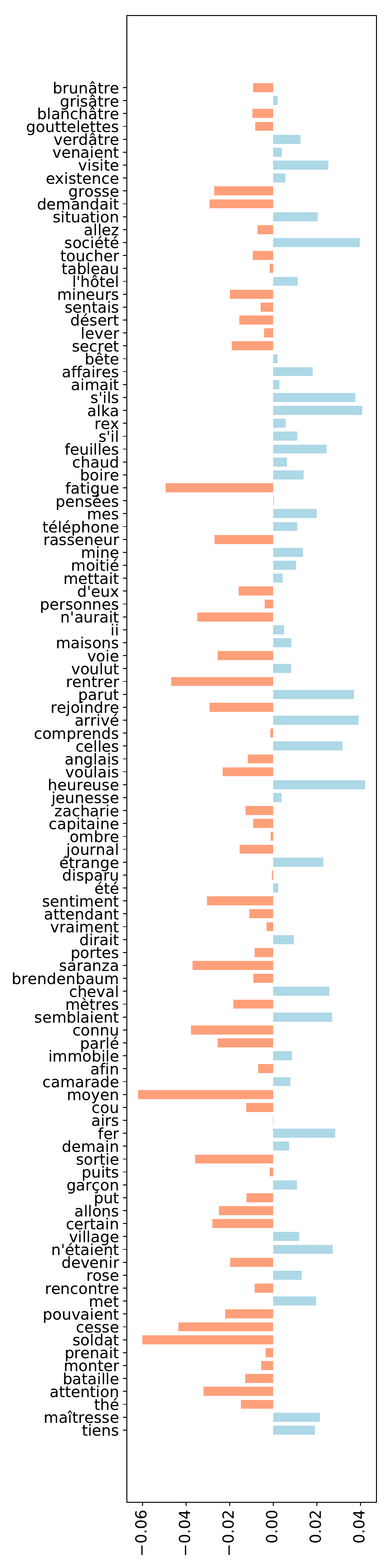}
     }  &
\subfloat[range $900$ - $1000$]{%
       \includegraphics[scale=0.2]{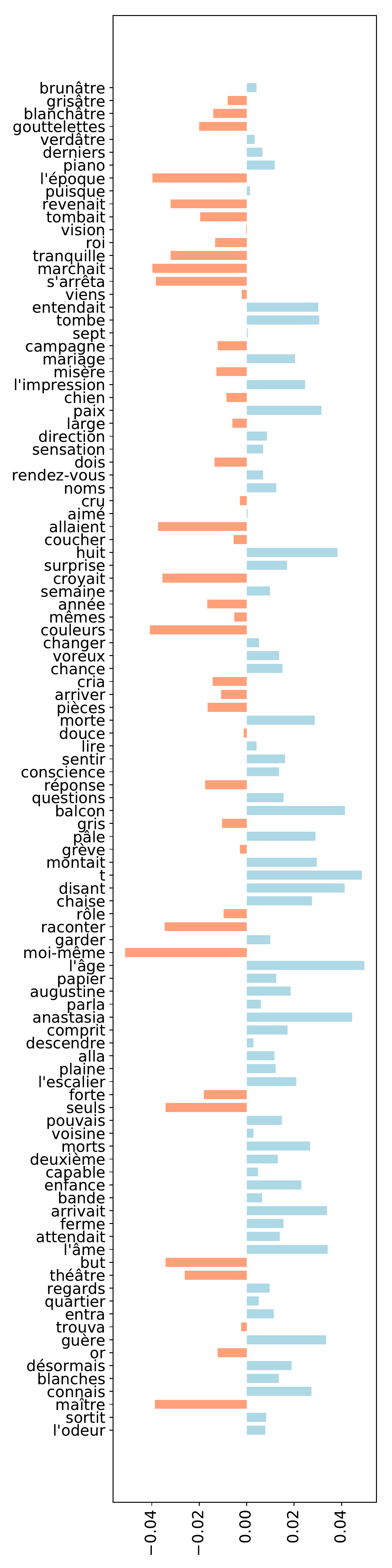}
     } 
\end{tabular}
\caption{\textbf{Ridge weights for groups A and B classification.} Words driving the classification outcome using Ridge trained on different ranges from the most common words ranking. Negative weights (orange bars) drive the decision towards Russian-French and positive weights (blue bars) towards French-French classic authors.\label{fig:ridge_weights}}
\end{figure}

\begin{figure}
  \advance\leftskip-1cm
\begin{tabular}{ccccc}
\subfloat[range $500$ - $600$]{%
       \includegraphics[scale=0.2]{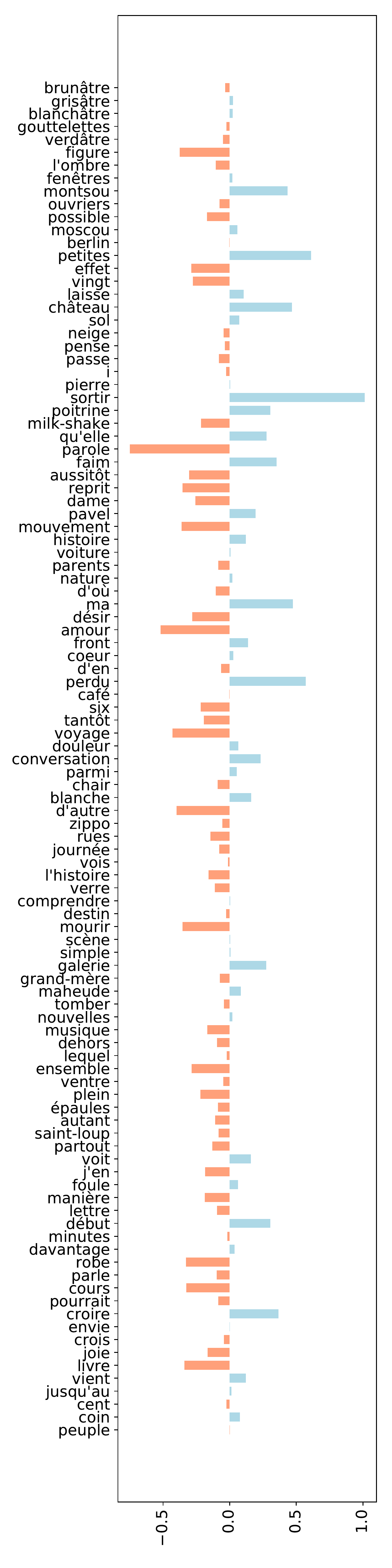}
     } &
\subfloat[range $600$ - $700$]{%
       \includegraphics[scale=0.2]{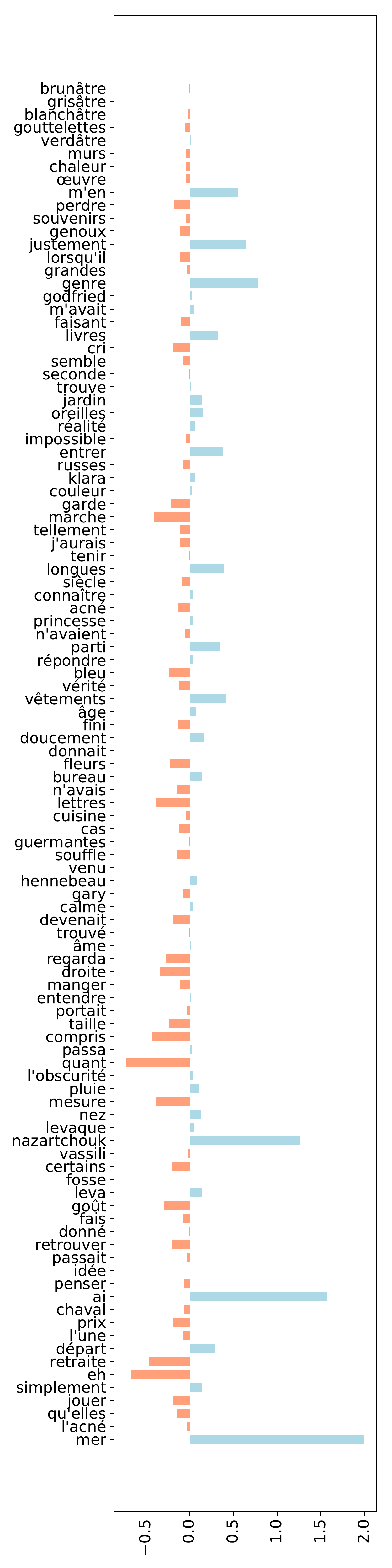}
     }  &
\subfloat[range $700$ - $800$]{%
       \includegraphics[scale=0.2]{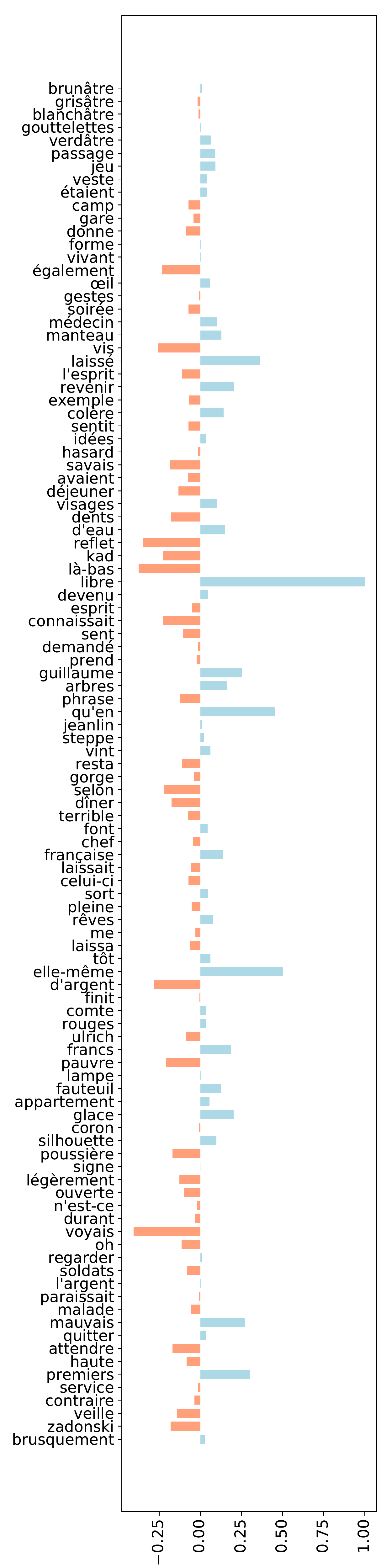}
     }  &
\subfloat[range $800$ - $900$]{%
       \includegraphics[scale=0.2]{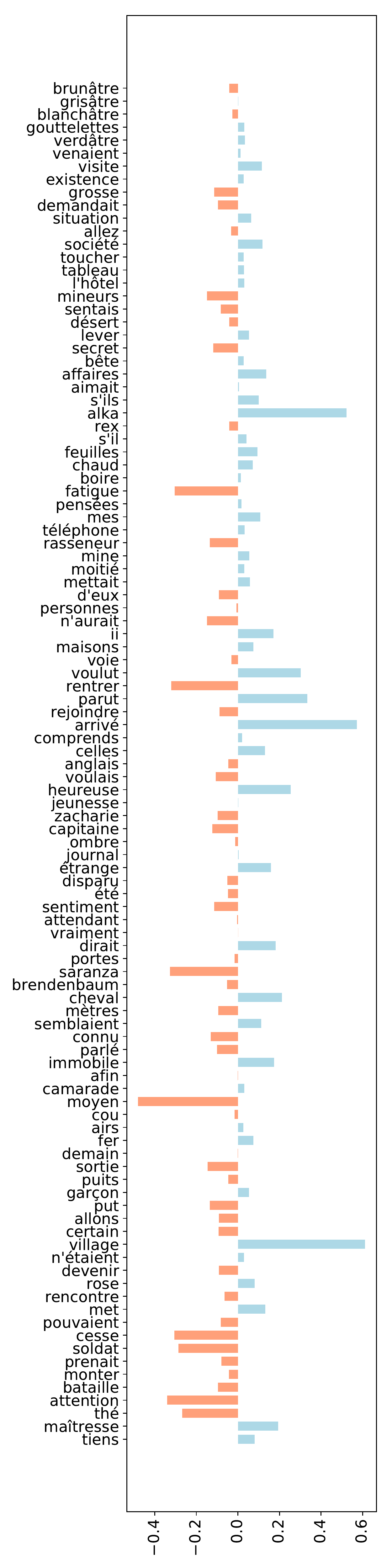}
     }  &
\subfloat[range $900$ - $1000$]{%
       \includegraphics[scale=0.2]{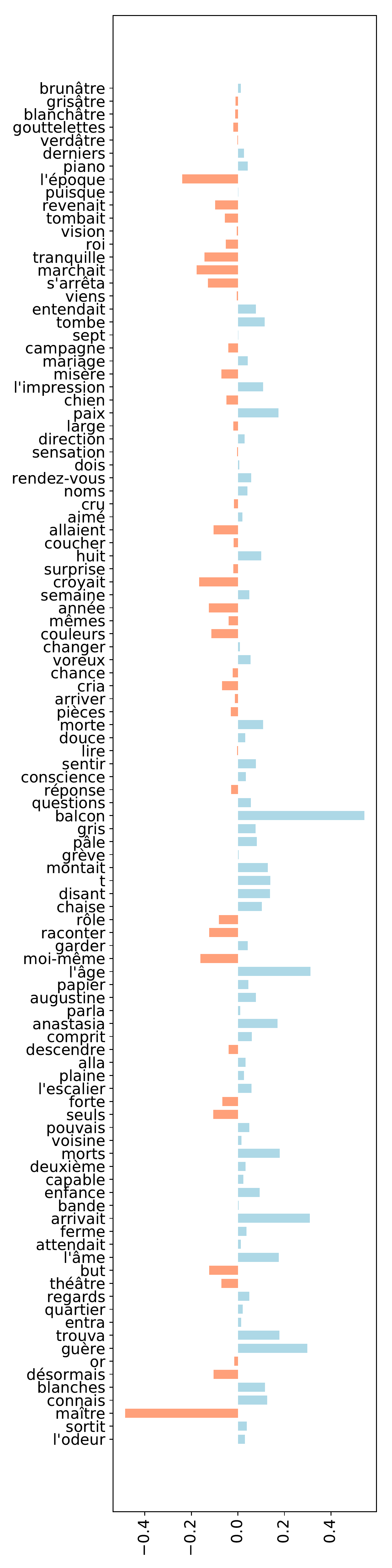}
     } 
\end{tabular}
\caption{\textbf{SVM weights for Groups A and B classification.} Words driving the classification outcome using SVM trained on different ranges from the most common words ranking. Negative weights (orange bars) drive the decision towards Russian-French and positive weights (blue bars) towards French-French classic authors.
\label{fig:svm_weights}}
\end{figure}

According to K. Baleevskikh, Andreï Makine uses many adverbs with ``ment'' ending, nouns with suffixes, diminutives, all of which are common in Russian, but unusual in French (\cite{baleevskikh}: 164). In order to verify this claim, we check the weights of the following words: ``brunâtre'', ``grisâtre'', ``blanchâtre'', ``verdâtre'', ``gouttelette[s]''. We note them while reading Makine's five novels (``brunâtre'', ``grisâtre'', ``blanchâtre'', ``verdâtre''), {\itshape Éducation nocturne} (``gouttelettes'', ``brunâtre'', ``grisâtre'', ``blanchâtre'', ``verdâtre'') and {\itshape Acné festival} (``gouttelettes''). They appear on the top of Figures~\ref{fig:ridge_weights},~\ref{fig:svm_weights}, there are 100 words in total. 

According to HathiTrust library (\url{https://www.hathitrust.org}), ``blanchâtre" has been used since the first half of the seventeenth century, mostly in scientific literature, especially in texts on medicine, zoology, botany, chemistry, sometimes physics. Since the twentieth century, it has been used more in botany and geology than in other disciplines. Google Books N-gram Viewer (\url{https://books.google.com/ngrams}) yields sporadic uses as early as the sixteenth century. In Gallica digital library (\url{https://gallica.bnf.fr}), French literature, with works by Hugo, Dumas, Zola, Balzac, and others, comes in second place after medicine in terms of usage.

Russian-French authors use ``blanchâtre'' in contexts associated with light, illumination or the color white (9 occurrences), for instance:

\begin{enumerate}
  \item ``..., ces visages lui apparaissent dans la lumière \textbf{blanchâtre} de la mort..." (\cite{jurgenson}: 182).
  \item ``...un énorme champignon de fumée dont le lourd chapeau \textbf{blanchâtre} s’enroulait sur une tige" (\cite{makine1}: 65).
  \item ``..., le champignon \textbf{blanchâtre}, majestueux et arrogant" (\cite{makine1}: 66).
  \item ``Les bâtons dans les mains, debout sur les éclats \textbf{blanchâtres}, nous attendîmes" (\cite{makine1}: 81).
  \item ``Nous interrogeâmes longuement son regard dans la lumière \textbf{blanchâtre} d’un réverbère nocturne" (\cite{makine2}: 124).
  \end{enumerate}

French-French authors use it only twice. Zola describes trees:

\begin{quote}
``Ces bois, du reste, devenaient amusants à voir, d’une pâleur jaunie de marbre, frangés de guipures \textbf{blanchâtres}, de végétations floconneuses..." (Zola, \url{https://gutenberg.org}).
\end{quote}

Balzac uses this adjective in a very uncommon way, to describe rain:

\begin{quote}
``…puis tous prirent une expression malicieuse en regardant le badaud qu'ils aspergèrent d’une pluie fine et \textbf{blanchâtre}..." (Balzac, \url{https://gutenberg.org}).
\end{quote}

The noun ``gouttelettes" is used more often in the plural than in the singular in the French Literary Corpus (\url{https://www.classiques-garnier.com/numerique-bases/}). The same can be seen in Google Books N-gram Viewer, which shows that the plural form has been used since the beginning of the sixteenth century. ``Gouttelette[s]" can be found in scientific literature in medicine, biology, zoology, physics, and chemistry, as shown in HathiTrust. The two most common uses of ``gouttelette[s]" in Gallica library are medicine and French literature. In both the French Literature Corpus and Frantext (\url{https://www.frantext.fr}) it is used quite often in relation to blood. 

In all examples from the Russian-French group (9 occurrences), except for the sentences below, ``gouttelettes'' refers to drops of water or sweat:

\begin{enumerate}
  \item ``La rougeur afflue à la peau d’Edwige en une myriade de \textbf{gouttelettes} brûlantes" (\cite{jurgenson}: 217).
  \item ``... ce petit-four surmonté d’une \textbf{gouttelette} de crème, dans un  plateau que lui tendait le serveur (\cite{makine5}: 217).
\end{enumerate}

 In both Russian and French, ``gouttelette" and ``a little" can be synonyms, but it is rarely the case in French. In the first sentence, the combination of ``gouttelettes" meaning ``a little" and the adjacent ``myriade" forms an oxymoron. In the second sentence, it also most likely means ``a little".

French-French writers use ``gouttelette" only once:

\begin{quote}
``…quand les Guermantes me furent devenus indifférents et que la \textbf{gouttelette} de leur originalité ne fut plus vaporisée par mon imagination,..." (Proust, \url{https://gutenberg.org}).
\end{quote}

Thus, ``blanchâtre" and ``gouttelette" are not cases of interference. They are common in Russian and in French. They are used by the authors of both groups, although more by group A. Ridge and SVM show similar weights for both groups, the bars are small. These words do not influence the classification. 

As for the adverbs ending in ``ment", the adverbs ``brusquement", ``doucement" and ``simplement" tilted both Ridge and SVM classifications in favor of the French-French classics and not the Russian-French bilingual writers (Figures~\ref{fig:ridge_weights},~\ref{fig:svm_weights}). This refutes Baleevskikh's hypothesis that adverbs with ``ment'' ending are characteristic of Russian language. Their weights are, on the contrary, more important for the French-French group. 

\paragraph{Groups A and C}

In the classification between groups A and C we do not check the weights of words introduced in advance, as in previous classifications. 

We check a new hypothesis: whether Russian-French authors use the demonstratives ``ce", ``cette", ``ces" instead of definite articles ``le", ``la", ``les" by analogy with Russian, where the definite article is substituted by demonstratives ``this"/``these" (``этот" s., m., ``эта" s., f., ``это" s., n., ``эти" pl.). The forms ``cet" and ``l' " are excluded as least frequent. This idea is inspired by the article by \cite{koppeletal2009}. These researchers note the tendency of Russian-speakers who speak English to omit the article in English because of its absence in Russian. A parallel arises with French, which, like English, has articles. The omission or replacement of the article under the influence of Russian can be regarded as a case of grammatical interference.

To correctly process the data, we remove the step ``remove stop words" in the algorithm, since ``ce", ``cette", and ``ces" are included in the list of stop words in the NLTK natural language processing package.

\begin{figure}
  \advance\leftskip-1cm
\begin{tabular}{ccccc}
\subfloat[range $500$ - $600$]{%
       \includegraphics[scale=0.2]{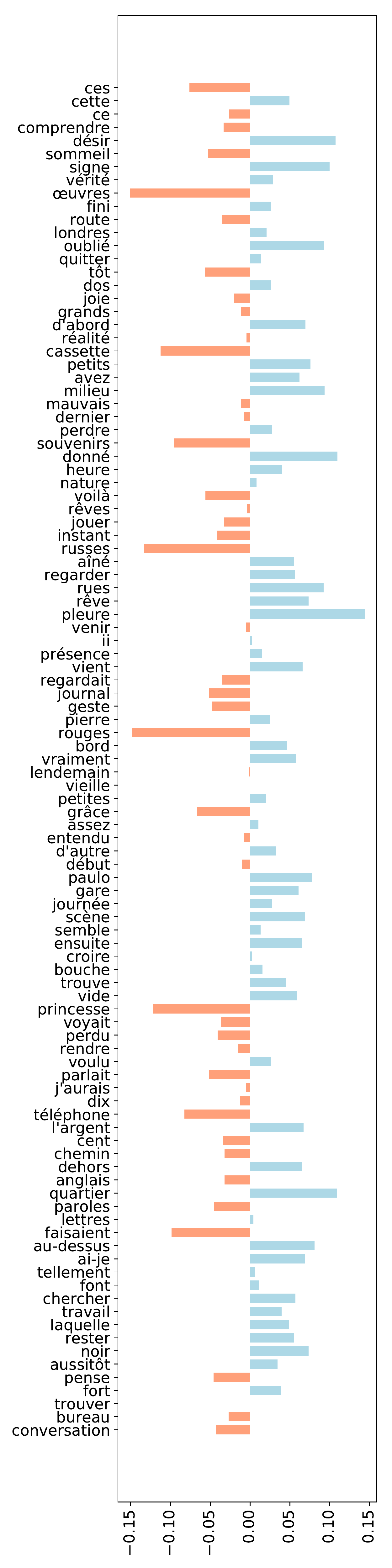}
     } &
\subfloat[range $600$ - $700$]{%
       \includegraphics[scale=0.2]{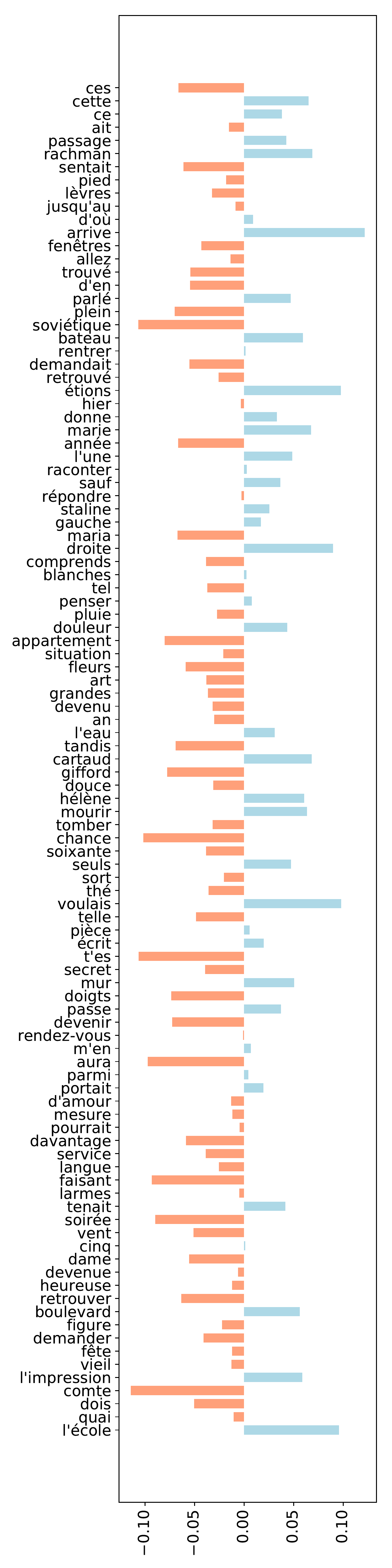}
     }  &
\subfloat[range $700$ - $800$]{%
       \includegraphics[scale=0.2]{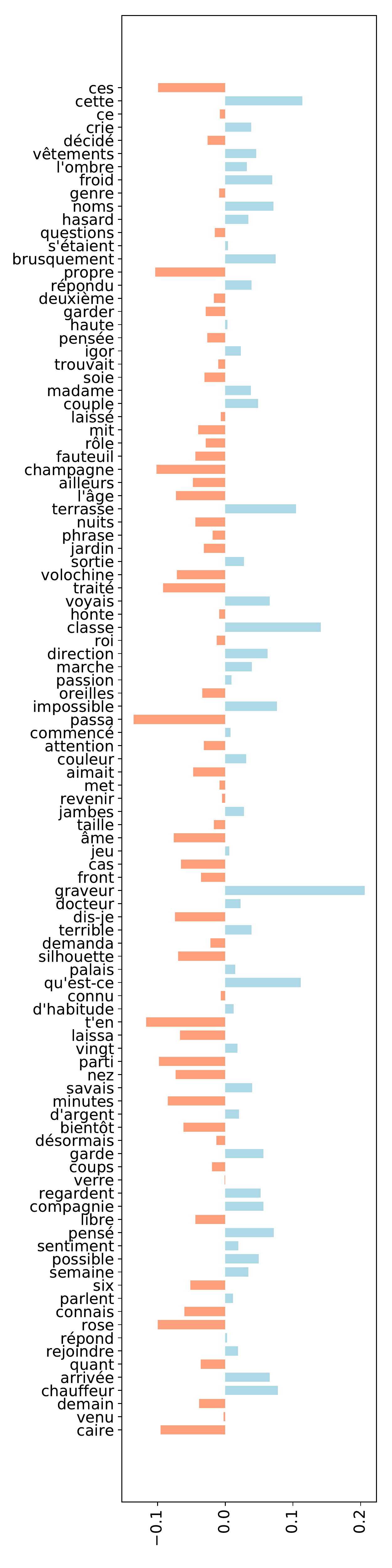}
     }  &
\subfloat[range $800$ - $900$]{%
       \includegraphics[scale=0.2]{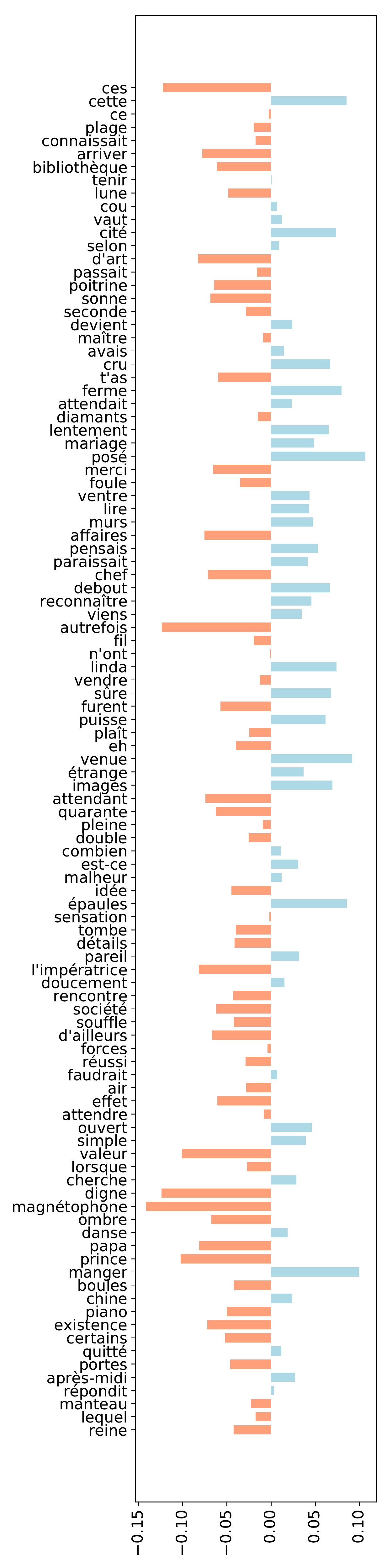}
     }  &
\subfloat[range $900$ - $1000$]{%
       \includegraphics[scale=0.2]{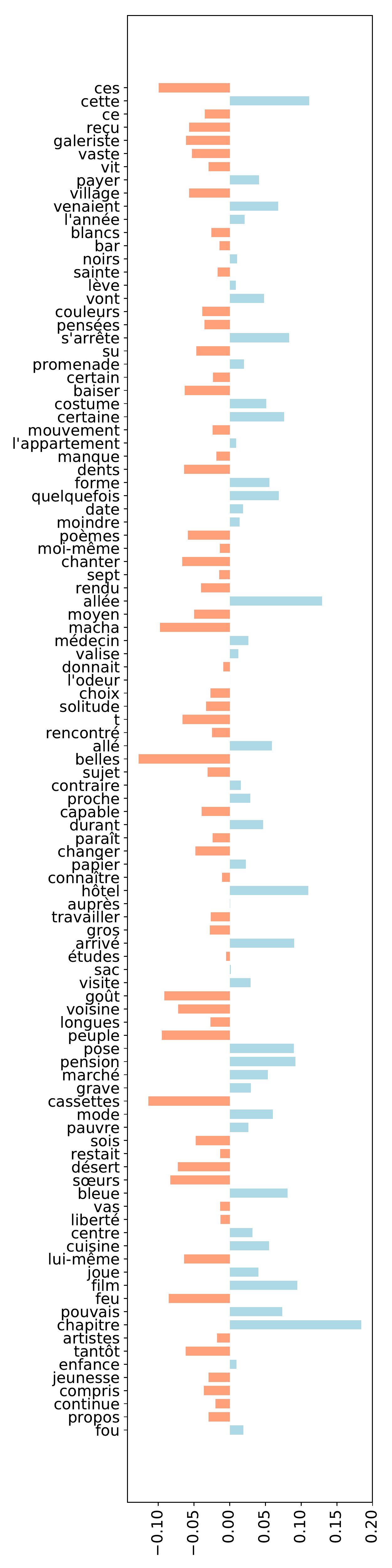}
     } 
\end{tabular}
\caption{\textbf{Ridge weights for Groups A and C classification.} Words driving the classification outcome using Ridge trained on different ranges from the most common words ranking. Negative weights (orange bars) drive the decision towards Russian-French and positive weights (blue bars) towards French-French contemporary authors. \label{fig:ridge_weights_modern}}
\end{figure}
\begin{figure}
  \advance\leftskip-1cm
\begin{tabular}{ccccc}
\subfloat[range $500$ - $600$]{%
       \includegraphics[scale=0.2]{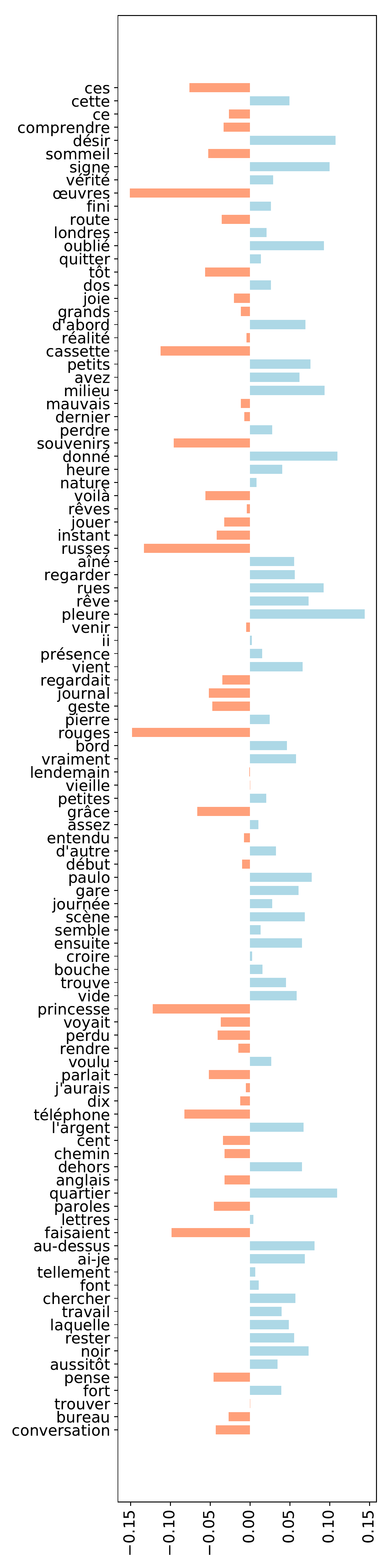}
     } &
\subfloat[range $600$ - $700$]{%
       \includegraphics[scale=0.2]{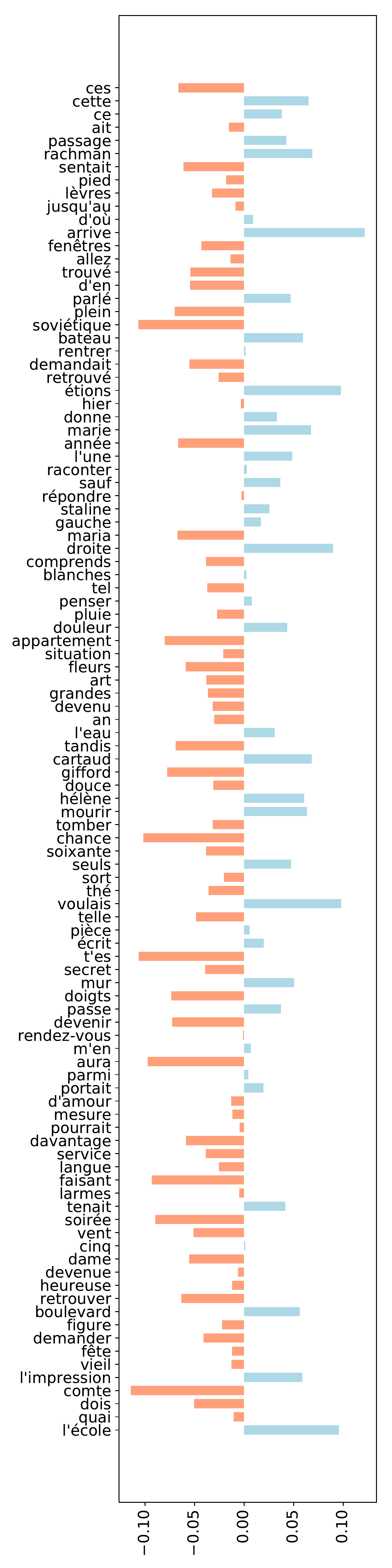}
     }  &
\subfloat[range $700$ - $800$]{%
       \includegraphics[scale=0.2]{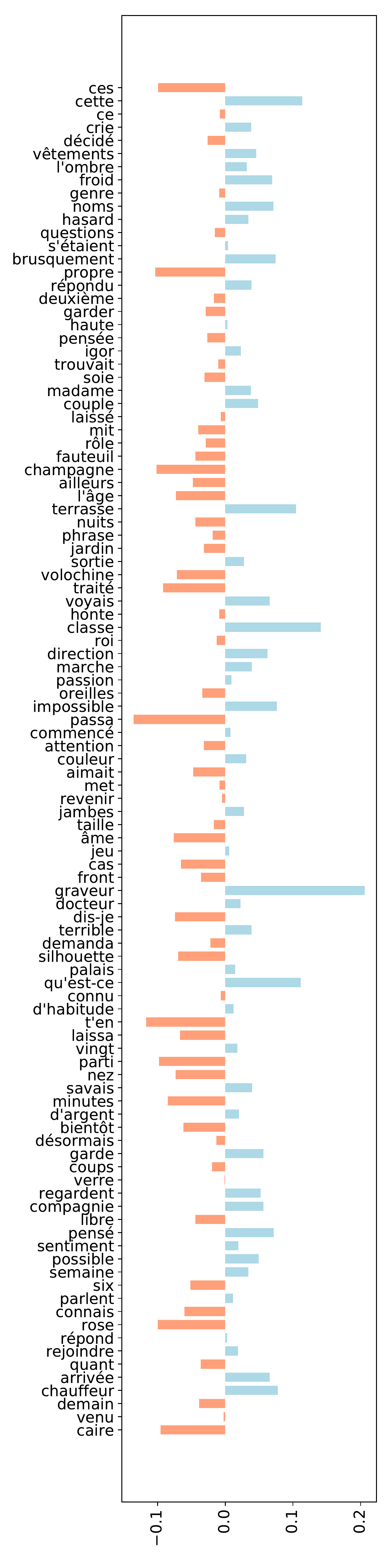}
     }  &
\subfloat[range $800$ - $900$]{%
       \includegraphics[scale=0.2]{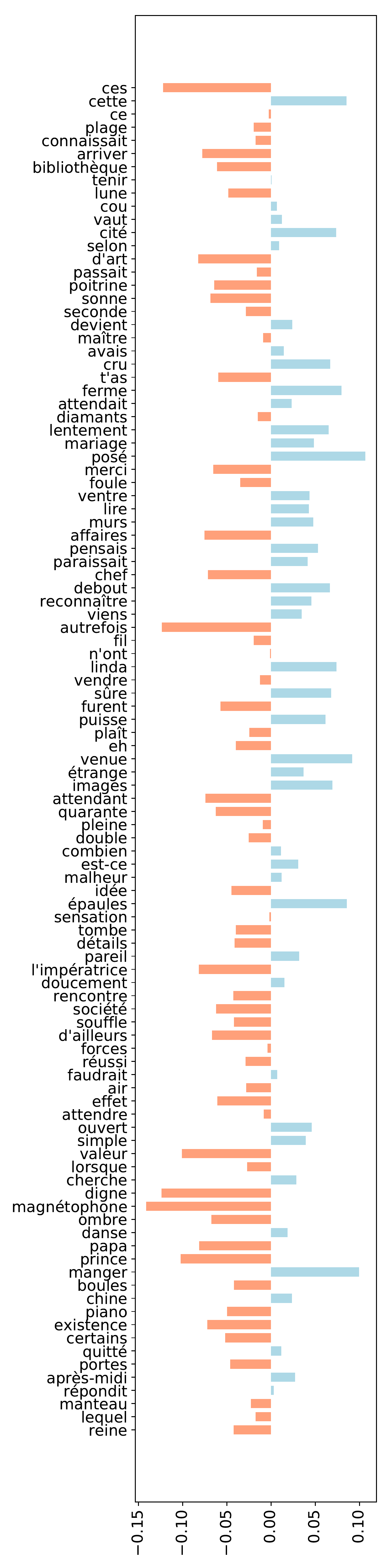}
     }  &
\subfloat[range $900$ - $1000$]{%
       \includegraphics[scale=0.2]{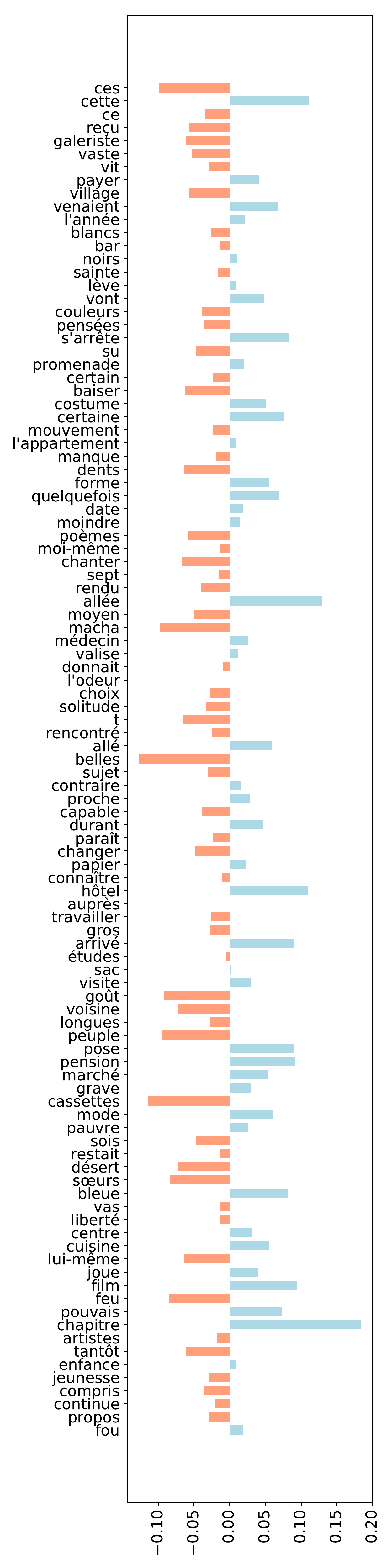}
     } 
\end{tabular}
\caption{\textbf{SVM weights for Groups A and C classification.} Words driving the classification outcome using Ridge trained on different ranges from the most common words ranking. Negative weights (orange bars) drive the decision towards Russian-French and positive weights (blue bars) towards French-French contemporary authors. \label{fig:svm_weights_modern}}
\end{figure}

The graphs (Figures~\ref{fig:ridge_weights_modern},~\ref{fig:svm_weights_modern}) show that the demonstrative ``ces" has a weight for the classification in favor of the Russian-French group. If we count the use of the above mentioned demonstratives and articles in both groups, their frequencies are close in both groups, but the intervals of imprecision (values after the sign ± Table \ref{celecettela}) do not overlap. 

\begin{table}[h!]
\begin{tabular}{|l|c|c|c|c|c|c|}
\hline
\rowcolor[HTML]{EFEFEF} 
\multicolumn{1}{|c|}{\cellcolor[HTML]{EFEFEF}}                       & ce                                                                                       & cette                                                             & ces                                                                & \multicolumn{1}{l|}{\cellcolor[HTML]{EFEFEF}le}                  & \multicolumn{1}{l|}{\cellcolor[HTML]{EFEFEF}la}                  & \multicolumn{1}{l|}{\cellcolor[HTML]{EFEFEF}les}                 \\ \hline
Russian-French                                                       & {\color[HTML]{000000} \begin{tabular}[c]{@{}c@{}}0.45 \\ ± 0.02\%\end{tabular}}          & \textbf{\begin{tabular}[c]{@{}c@{}}0.37 \\ ± 0.05\%\end{tabular}} & \textbf{\begin{tabular}[c]{@{}c@{}}0.12 \\ ± 0.005\%\end{tabular}} & \begin{tabular}[c]{@{}c@{}}1.8 \\ ± 0.08\%\end{tabular}          & \begin{tabular}[c]{@{}c@{}}2.3 \\ ± 0.1\%\end{tabular}           & \begin{tabular}[c]{@{}c@{}}1.4 \\ ± 0.07\%\end{tabular}          \\ \hline
\begin{tabular}[c]{@{}l@{}}French-French \\ (contempt.)\end{tabular} & {\color[HTML]{000000} \textbf{\begin{tabular}[c]{@{}c@{}}0.5 \\ ± 0.008\%\end{tabular}}} & \begin{tabular}[c]{@{}c@{}}0.25 \\ ± 0.006\%\end{tabular}         & \begin{tabular}[c]{@{}c@{}}0.11 \\ ± 0.003\%\end{tabular}          & \textbf{\begin{tabular}[c]{@{}c@{}}2.5 \\ ± 0.05\%\end{tabular}} & \textbf{\begin{tabular}[c]{@{}c@{}}3.4 \\ ± 0.04\%\end{tabular}} & \textbf{\begin{tabular}[c]{@{}c@{}}1.6 \\ ± 0.03\%\end{tabular}} \\ \hline
\end{tabular}
\caption{\textbf{Frequencies of definite articles and demonstratives in groups A and C}. \label{celecettela}}
\end{table}

\begin{figure}
\centering
\begin{tabular}{c}
\subfloat[Google Books N-gram Viewer shows the frequency of definite articles ``le”, ``la”, ``les” and the frequency of demonstratives ``ce", ``cette", ``ces" in all French texts from the seventeenth to the twenty-first century. \label{fig:time1}]{%
       \includegraphics[scale=0.25]{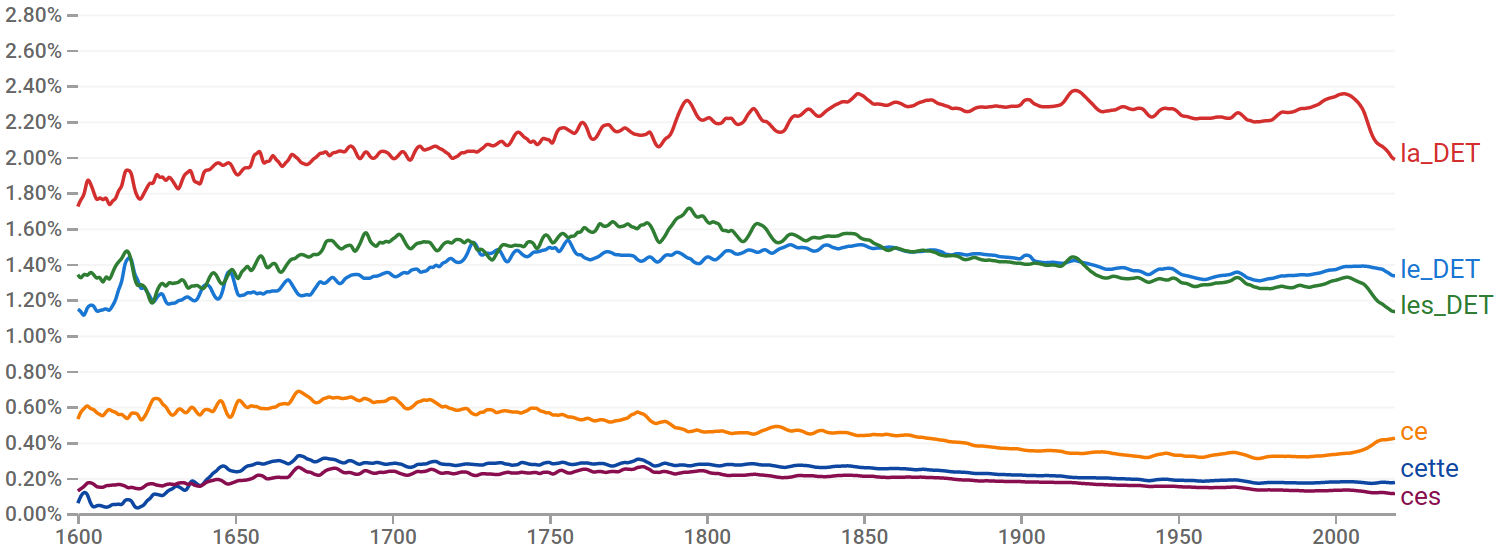}
     }  \\
\subfloat[Google Books N-gram Viewer shows the frequency of definite articles ``le”, ``la”, ``les” and the frequency of demonstratives ``ce", ``cette", ``ces" in all French texts from 1990 to 2000.\label{fig:time2}]{%
       \includegraphics[scale=0.25]{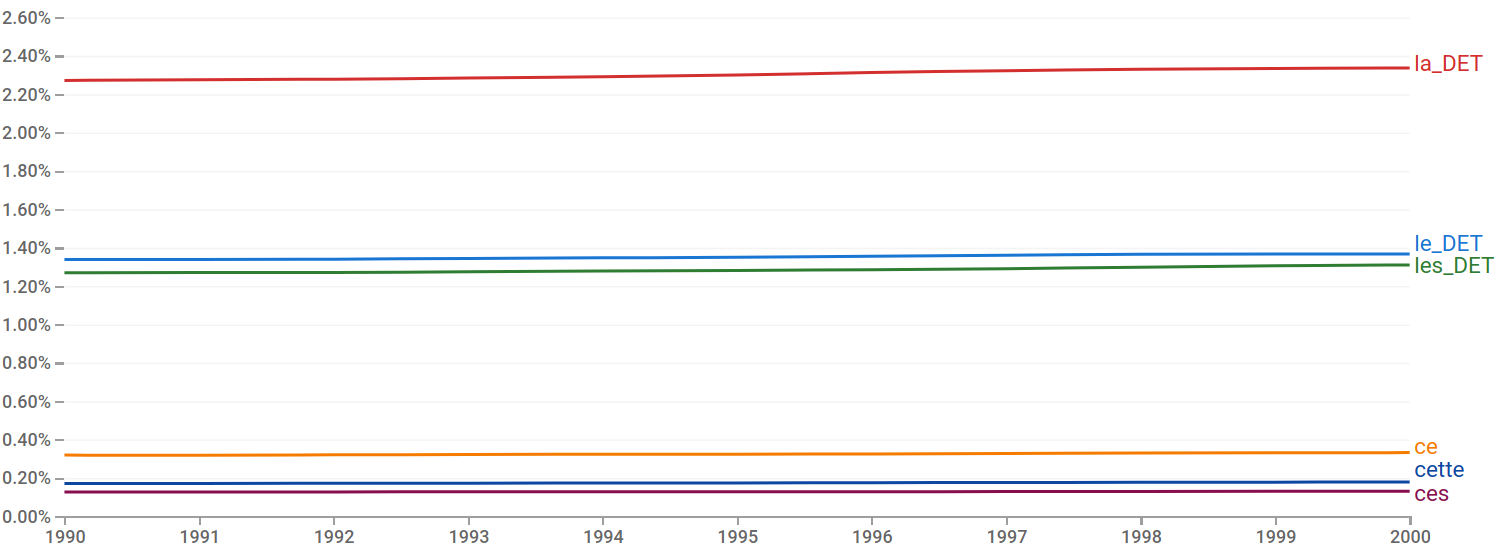}
     } 
\end{tabular}
\caption{\label{fig:google_n_grams}}
\end{figure}

Despite the fact that the intervals are close, it can be noted that ``cette" and ``ces" are used more frequently in the texts by Russian-French authors. ``Le", ``la", ``les" are used more frequently in the texts by French-French contemporary authors. The difference from the Russian-French group here is more significant. Google Books N-gram Viewer shows their frequency in all French texts from the seventeenth to the twenty-first century (Figure~\ref{fig:time1}) and all French texts from 1990 to 2000 (Figure~\ref{fig:time2}).

\looseness=-1
Thus, in Russian-French texts we notice the appearance of demonstratives instead of definite articles. The counts in Table \ref{celecettela} prove the predominance of demonstratives, except for the demonstrative ``ce". Ridge and SVM show that the demonstrative ``ces" tilts the classifications in favor of the Russian-French group. Using ``ces'' is specific to it, not to the group C, therefore it is a case of interference.

The Ridge and SVM classifications have disadvantages: 1) it takes a long time to search for each word that influenced the decision on the authorship of the text, because many segments can be created and many ranges can be considered; 2) we still have to evaluate the results using our knowledge of languages and Russian and French language corpora, which means Ridge and SVM are not self-sufficient; 3) since creating an algorithm with 100\% accuracy is impossible, classifications occasionally produce errors, which hinder us in finding interference; 4) sporadic interference is hard to detect.

\begin{figure}
       \advance\leftskip-1cm
       \begin{tabular}{cccc}
       \subfloat[Train for word range $300-1300$]{%
              \includegraphics[width=0.25\linewidth]{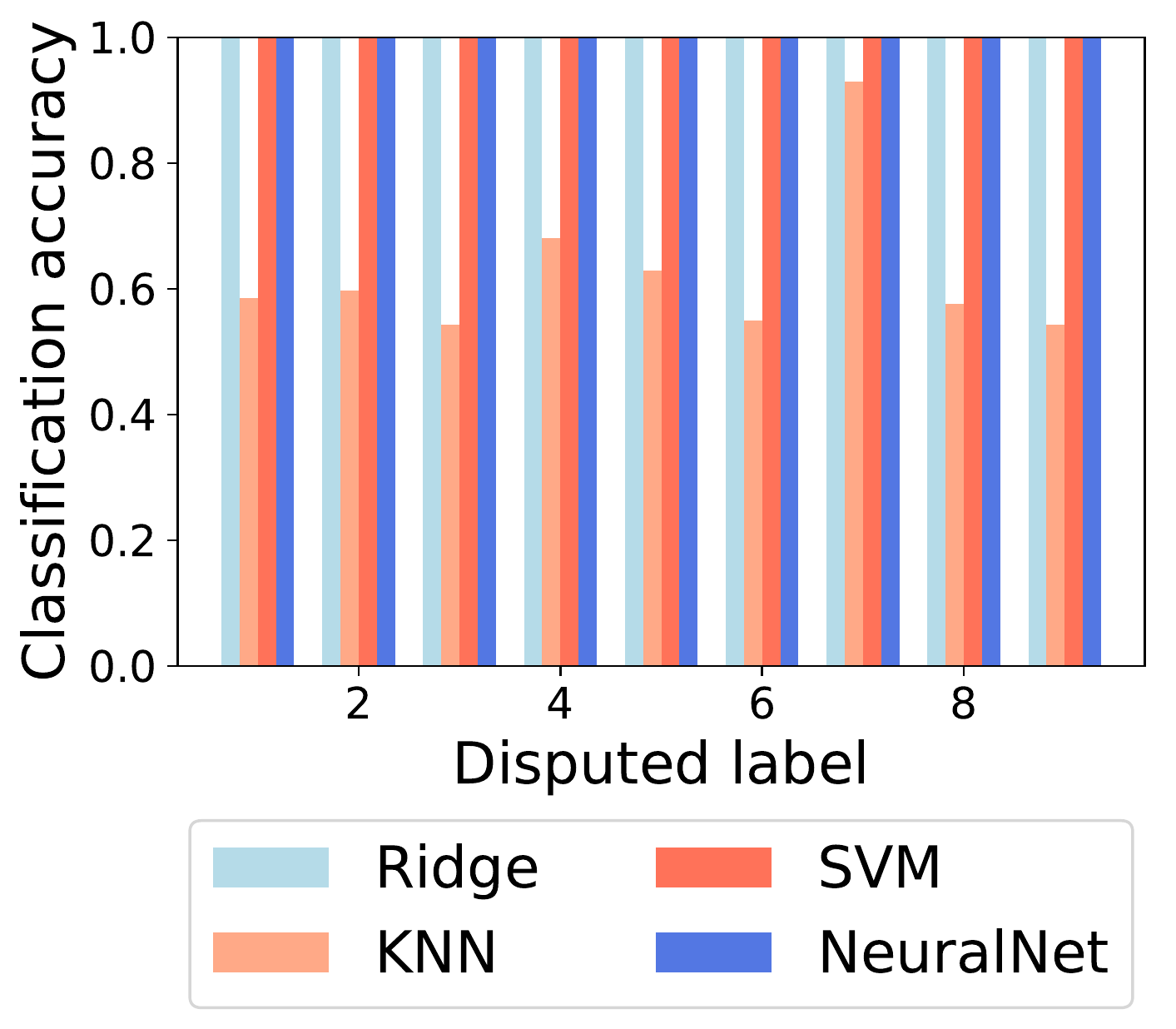}
            } &
       \subfloat[Test for word range $300-1300$]{%
              \includegraphics[width=0.25\linewidth]{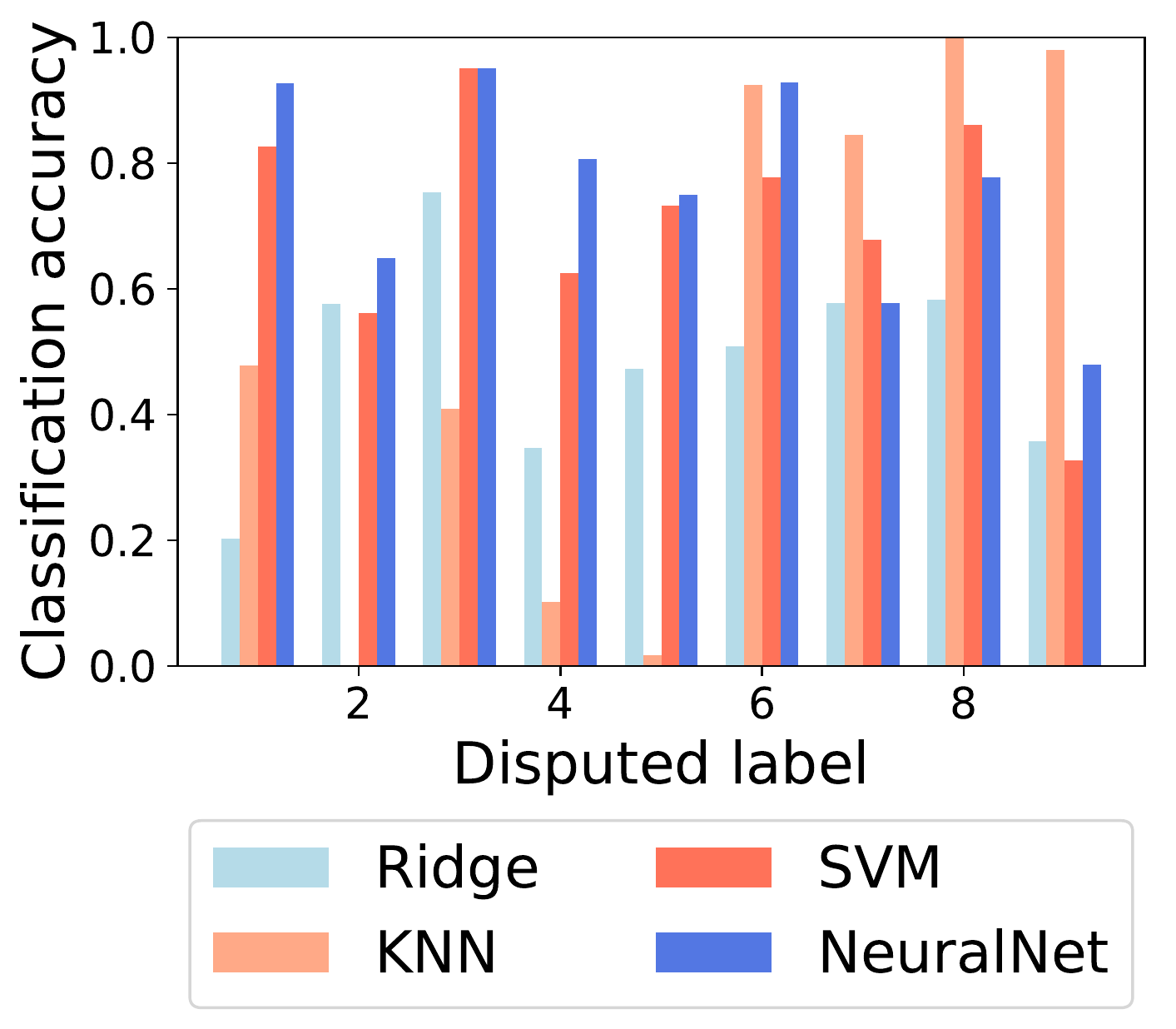}} &
       \subfloat[Train for word range $500-1000$]{%
              \includegraphics[width=0.25\linewidth]{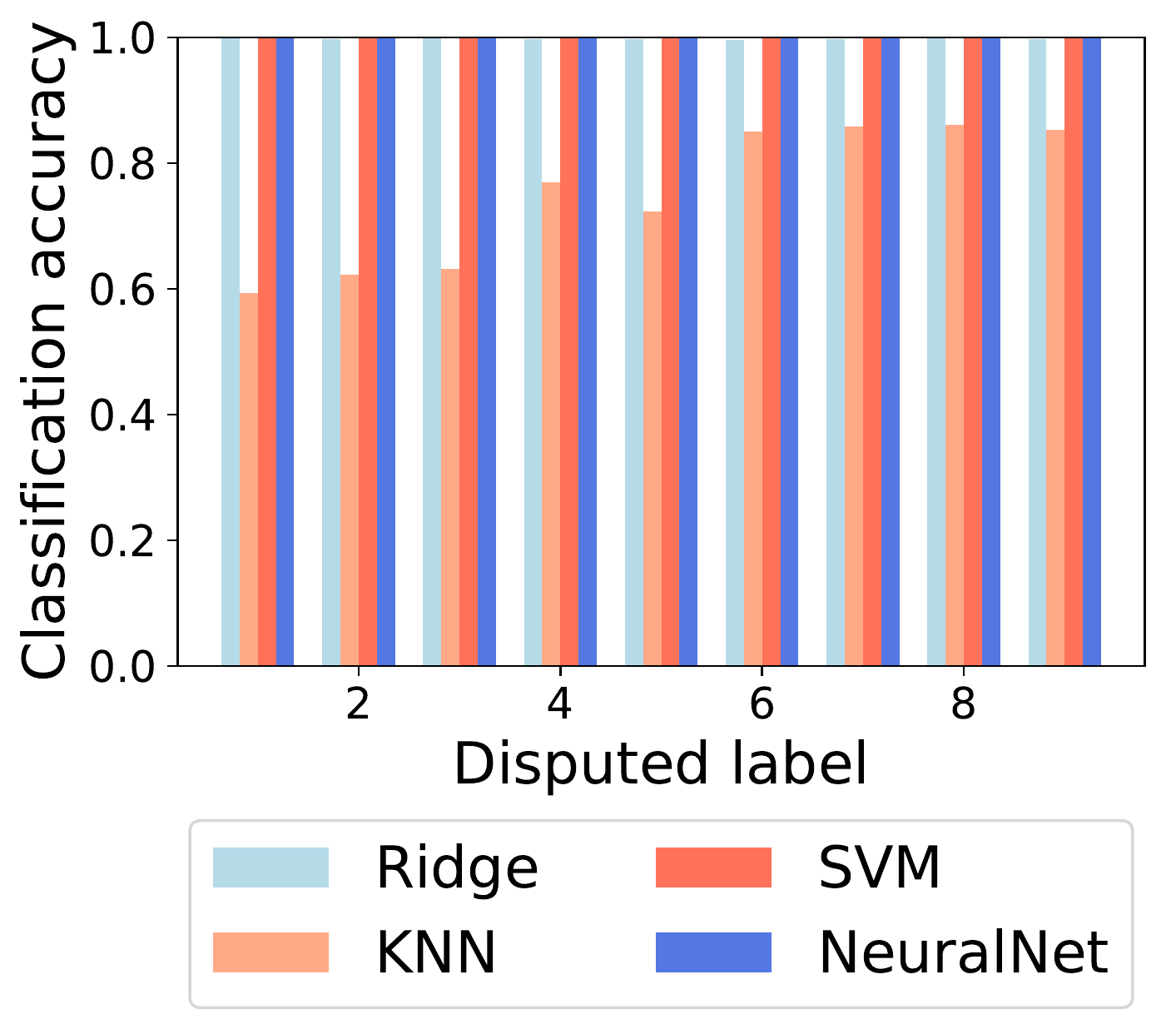}
            } &
       \subfloat[Test for word range $500-1000$]{%
              \includegraphics[width=0.25\linewidth]{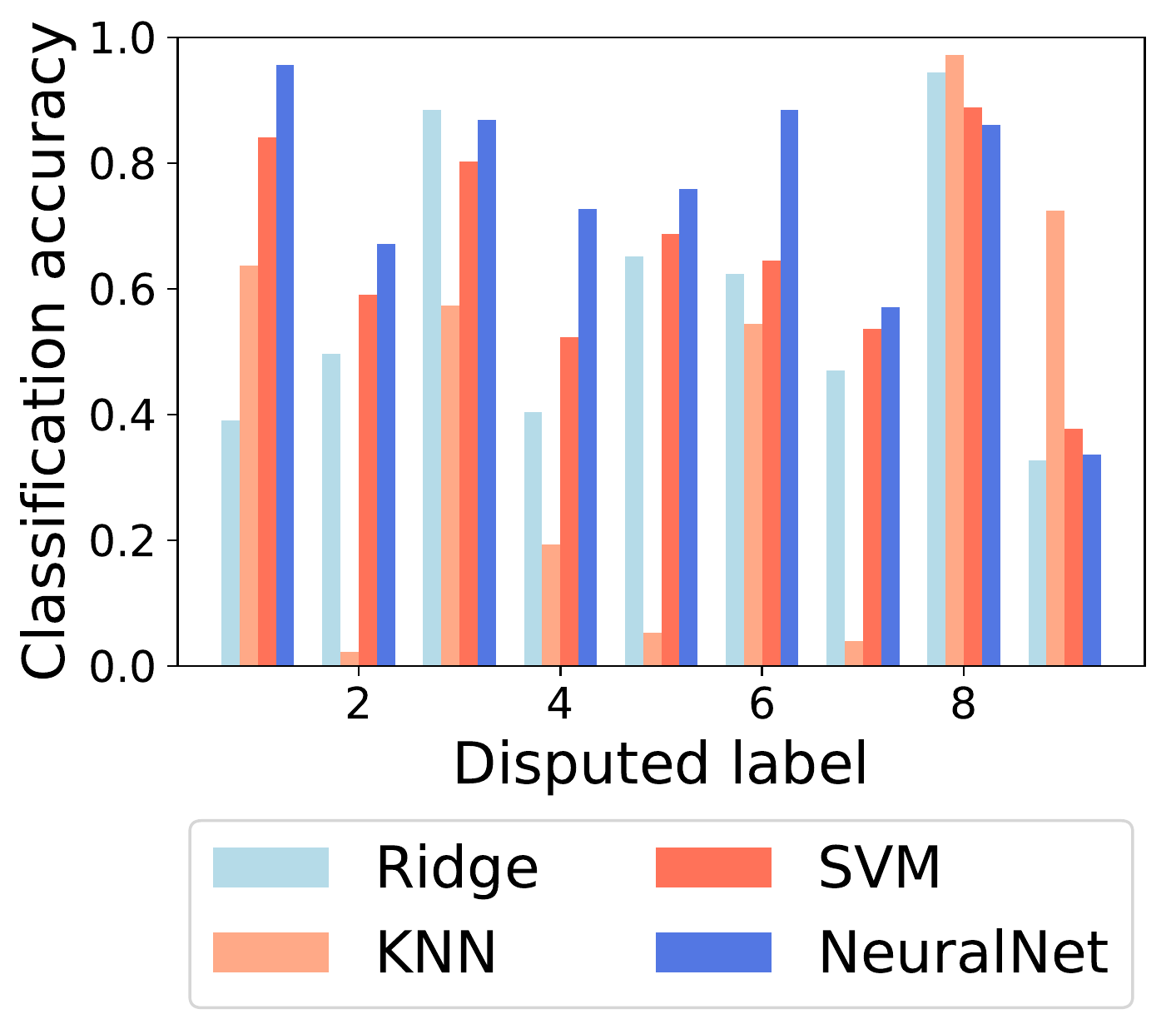}}\\
       \subfloat[Train for word range $300-800$]{%
              \includegraphics[width=0.25\linewidth]{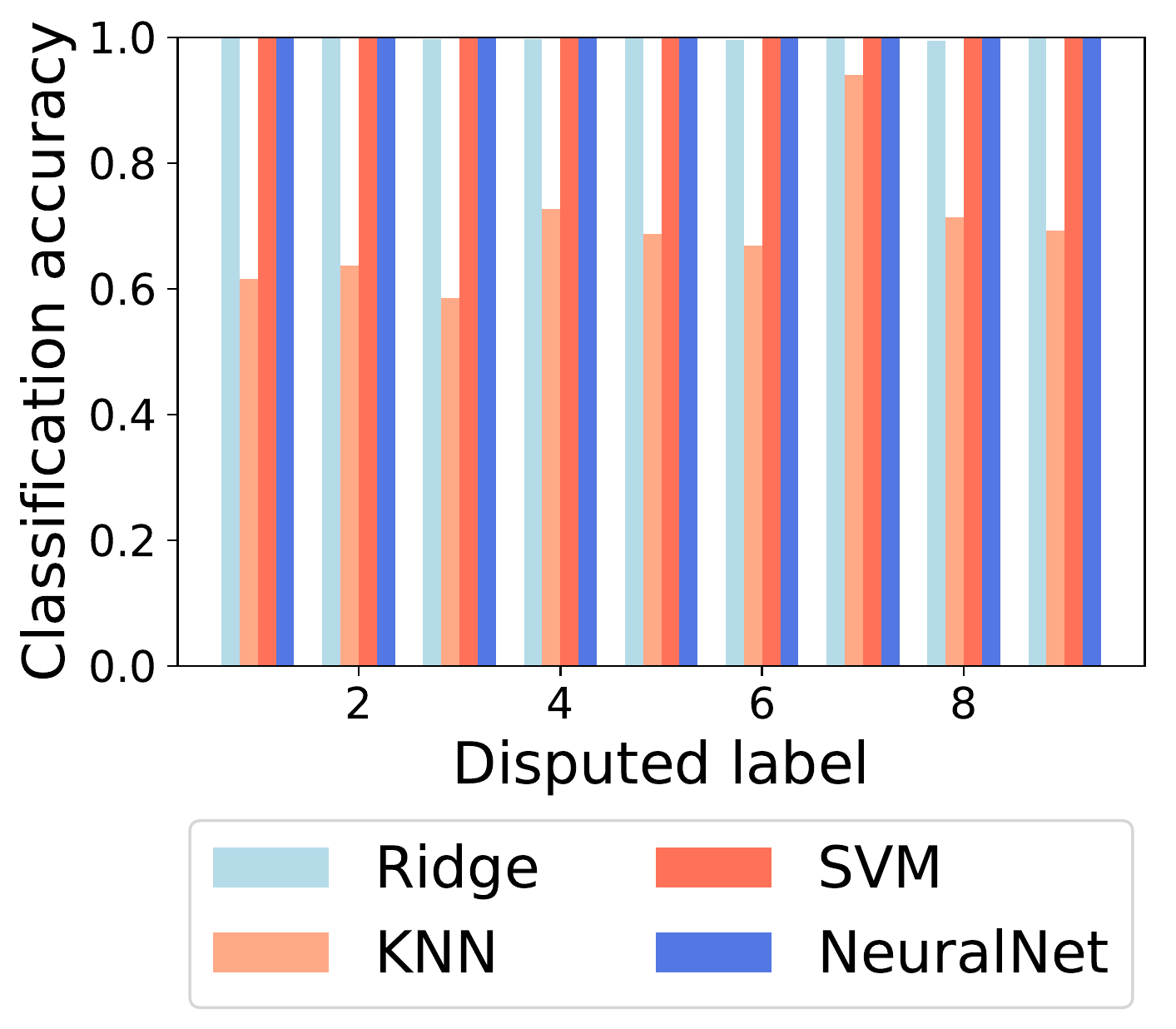}
            } &
       \subfloat[Test for word range $300-800$]{%
              \includegraphics[width=0.25\linewidth]{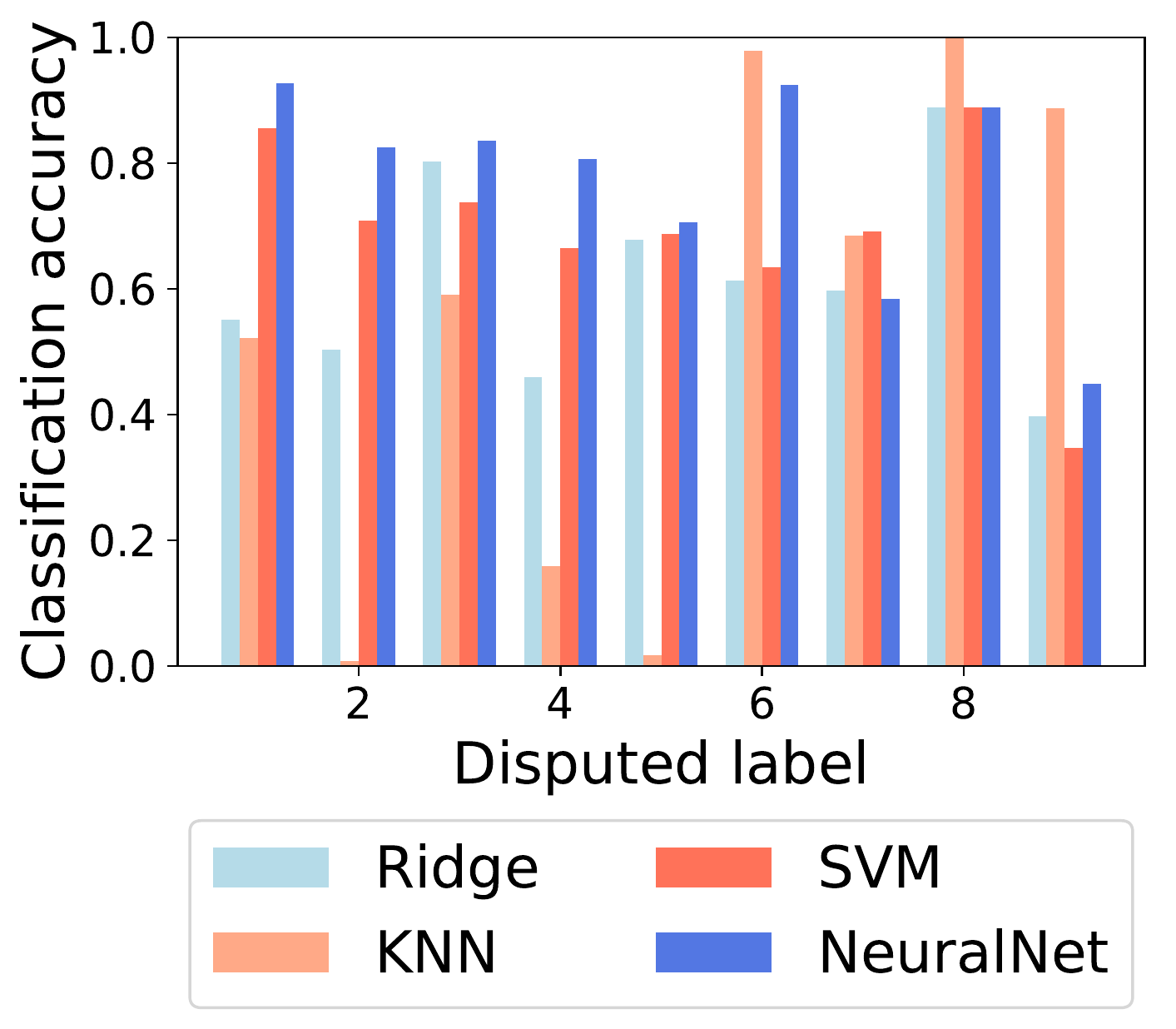}} &
       \subfloat[Train for word range $500-600$]{%
              \includegraphics[width=0.25\linewidth]{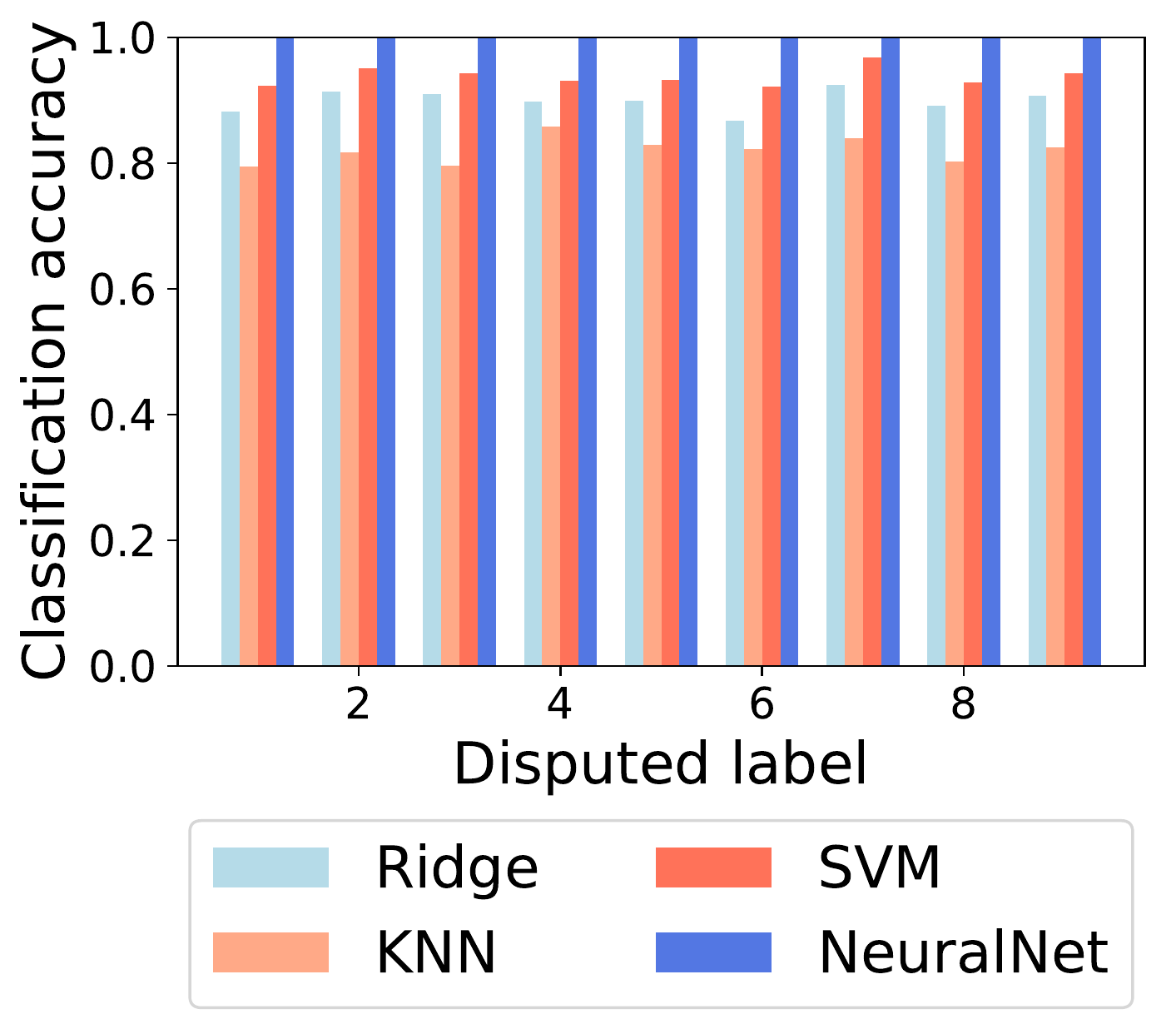}
            } &
       \subfloat[Test for word range $500-600$]{%
              \includegraphics[width=0.25\linewidth]{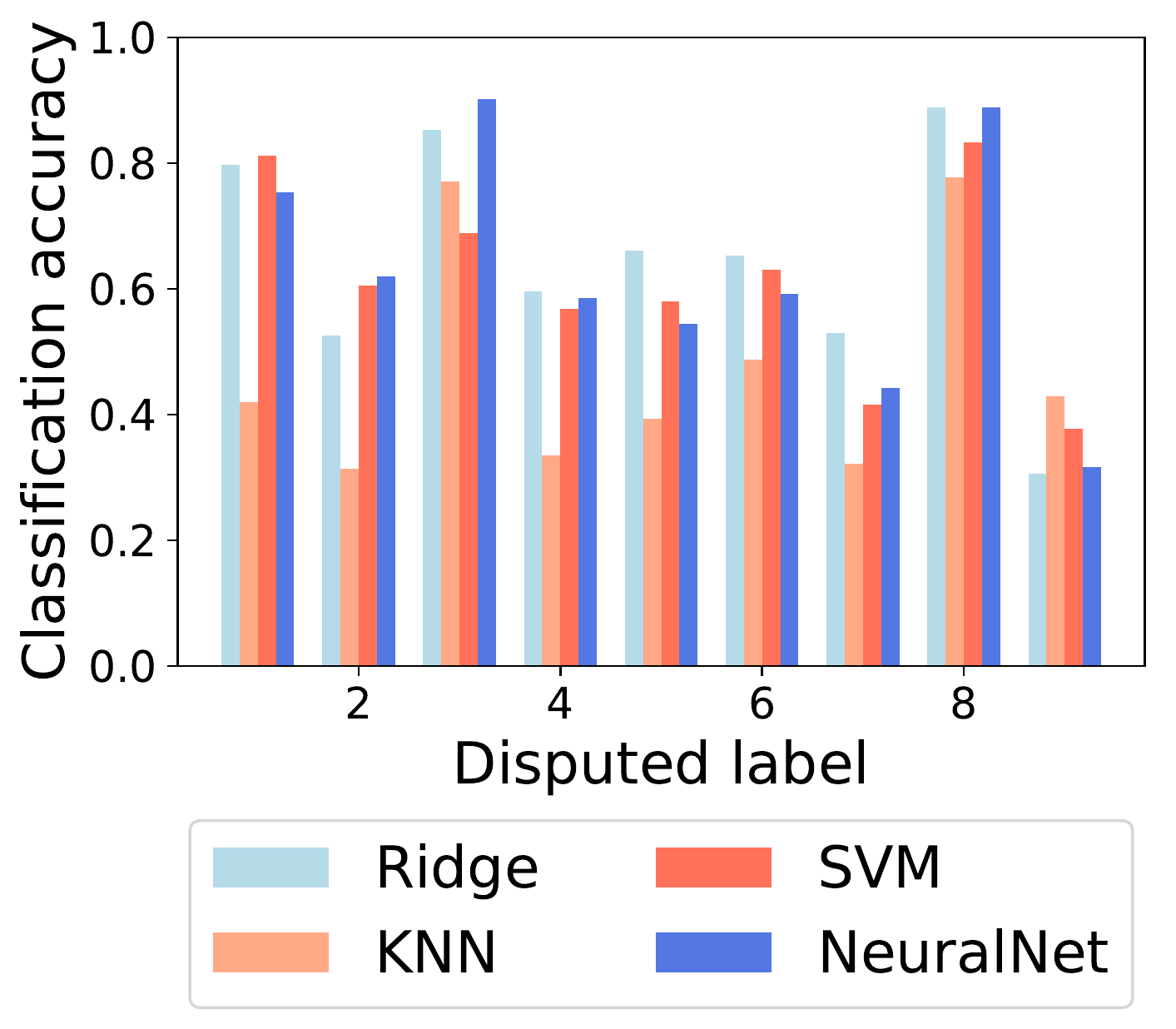}}\\
       \subfloat[Train for word range $600-700$]{%
              \includegraphics[width=0.25\linewidth]{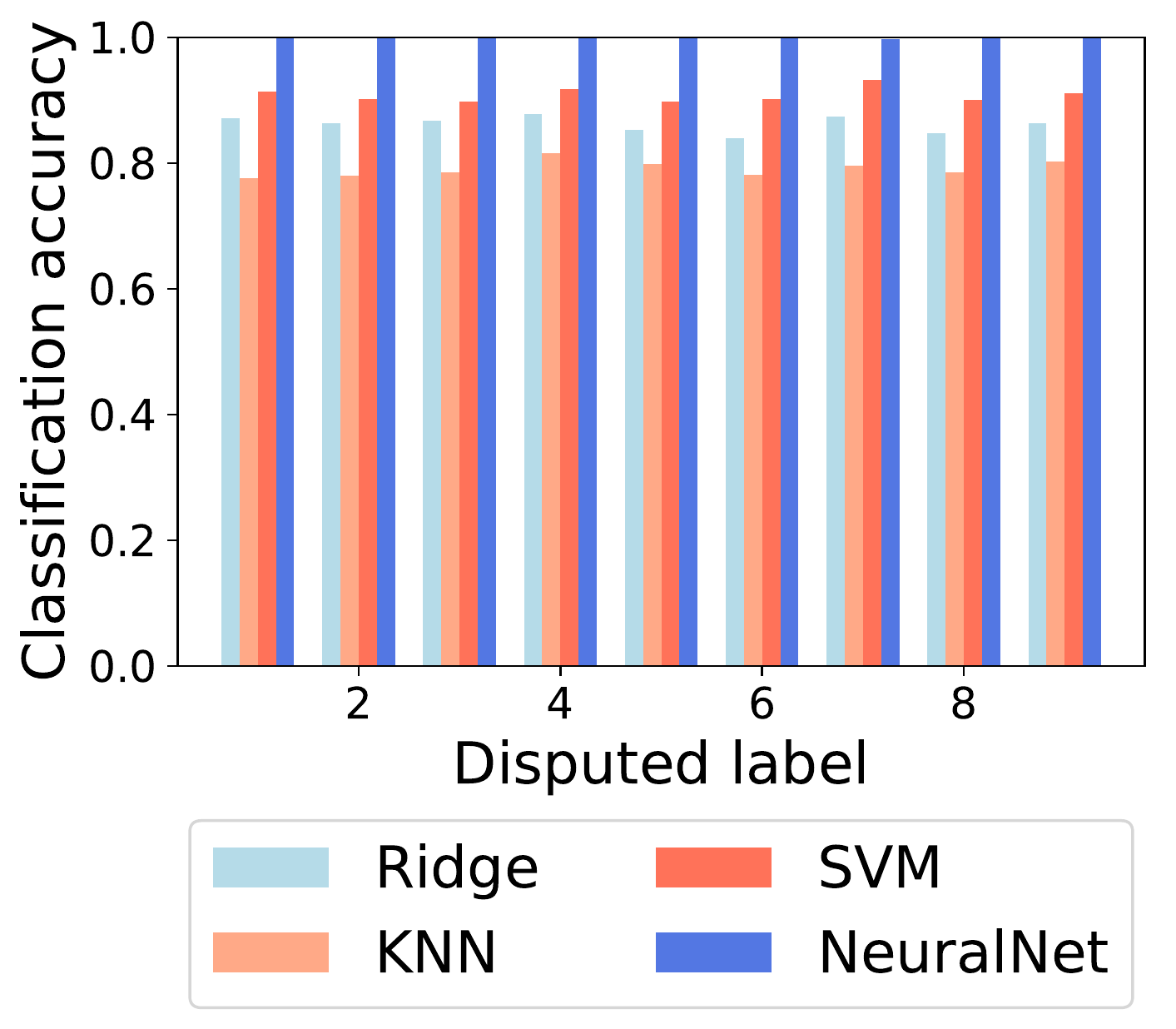}
            } &
       \subfloat[Test for word range $600-700$]{%
              \includegraphics[width=0.25\linewidth]{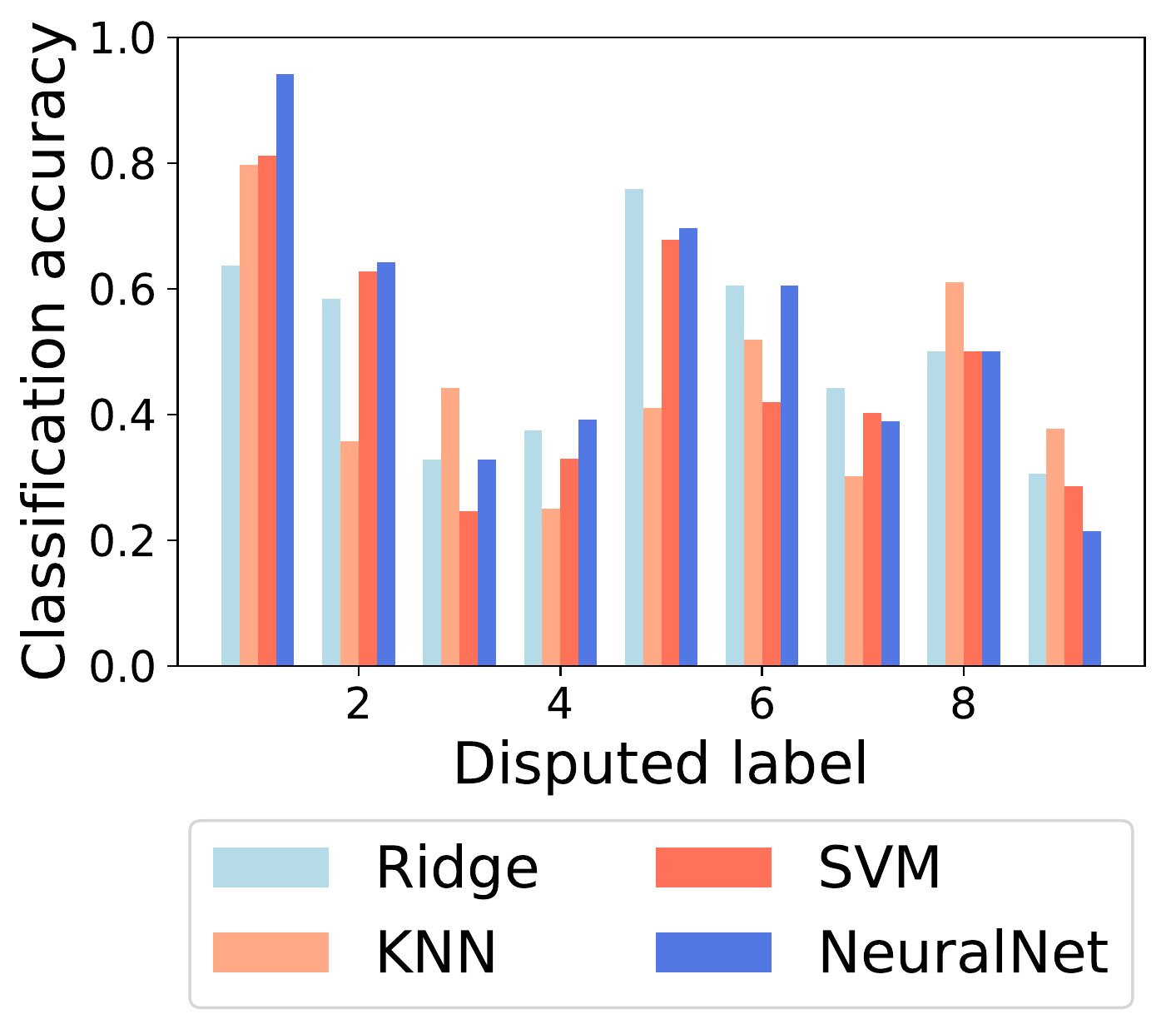}} &
       \subfloat[Train for word range $700-800$]{%
              \includegraphics[width=0.25\linewidth]{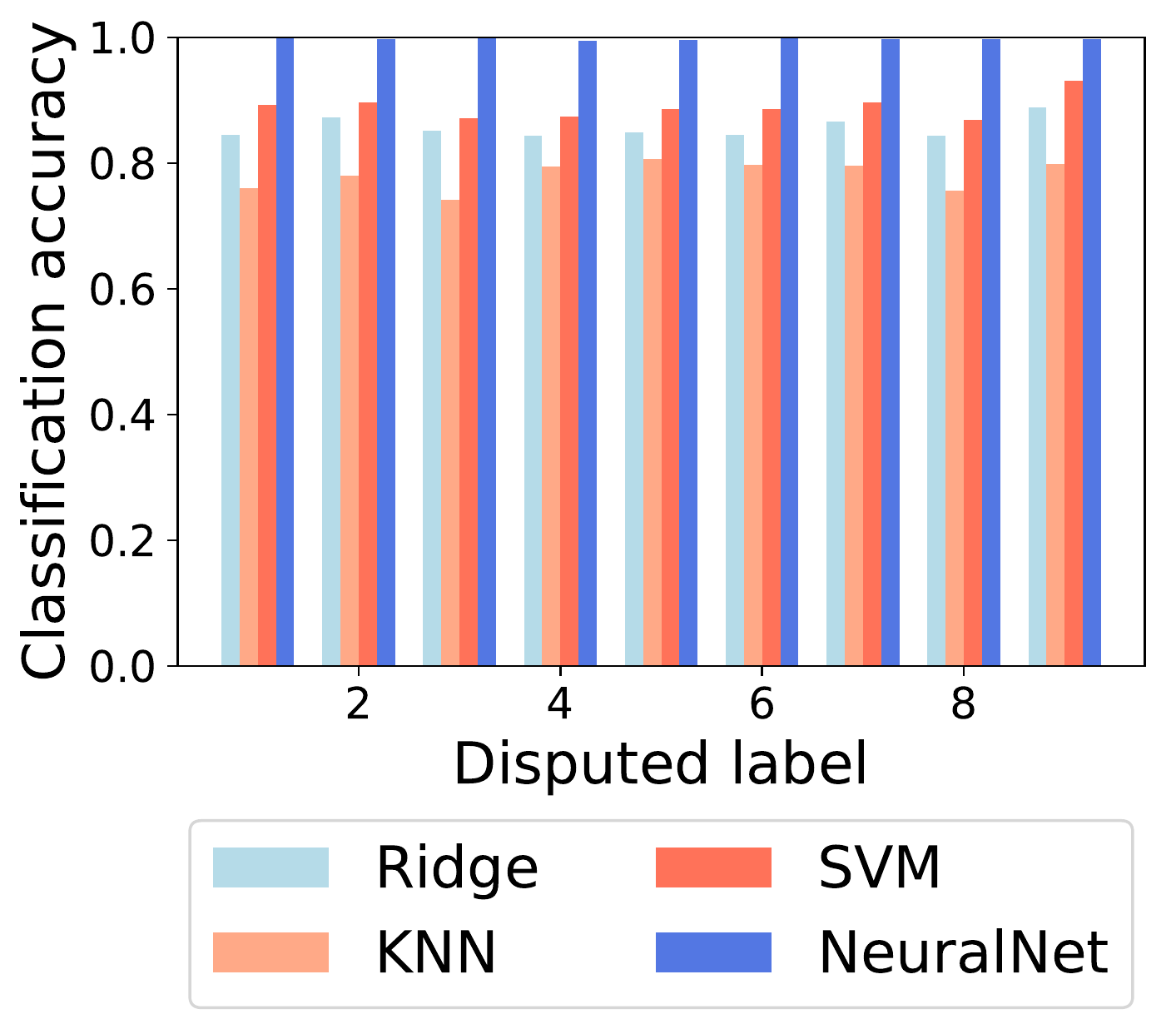}
            } &
       \subfloat[Test for word range $700-800$]{%
              \includegraphics[width=0.25\linewidth]{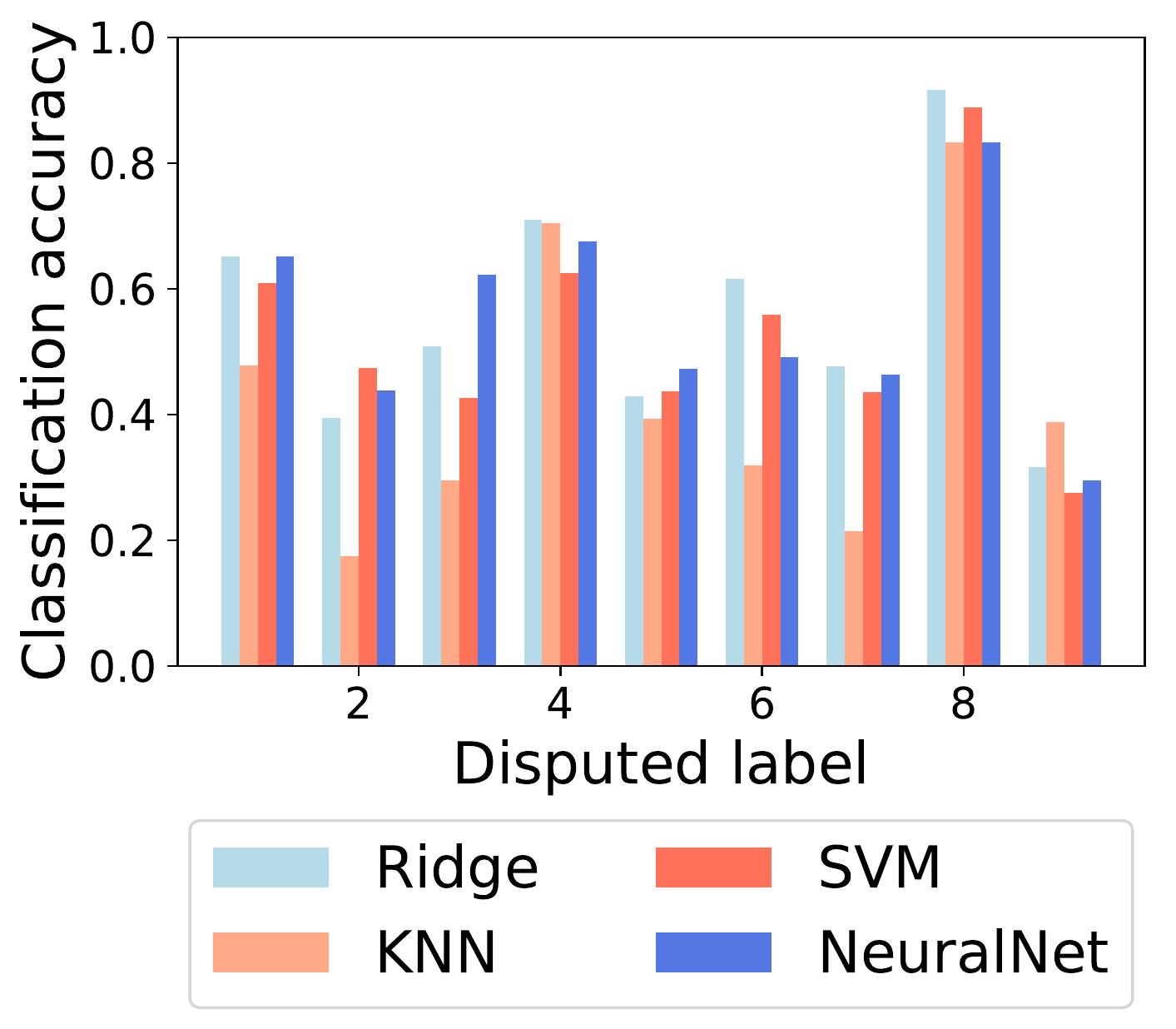}}\\
       \subfloat[Train for word range $800-900$]{%
              \includegraphics[width=0.25\linewidth]{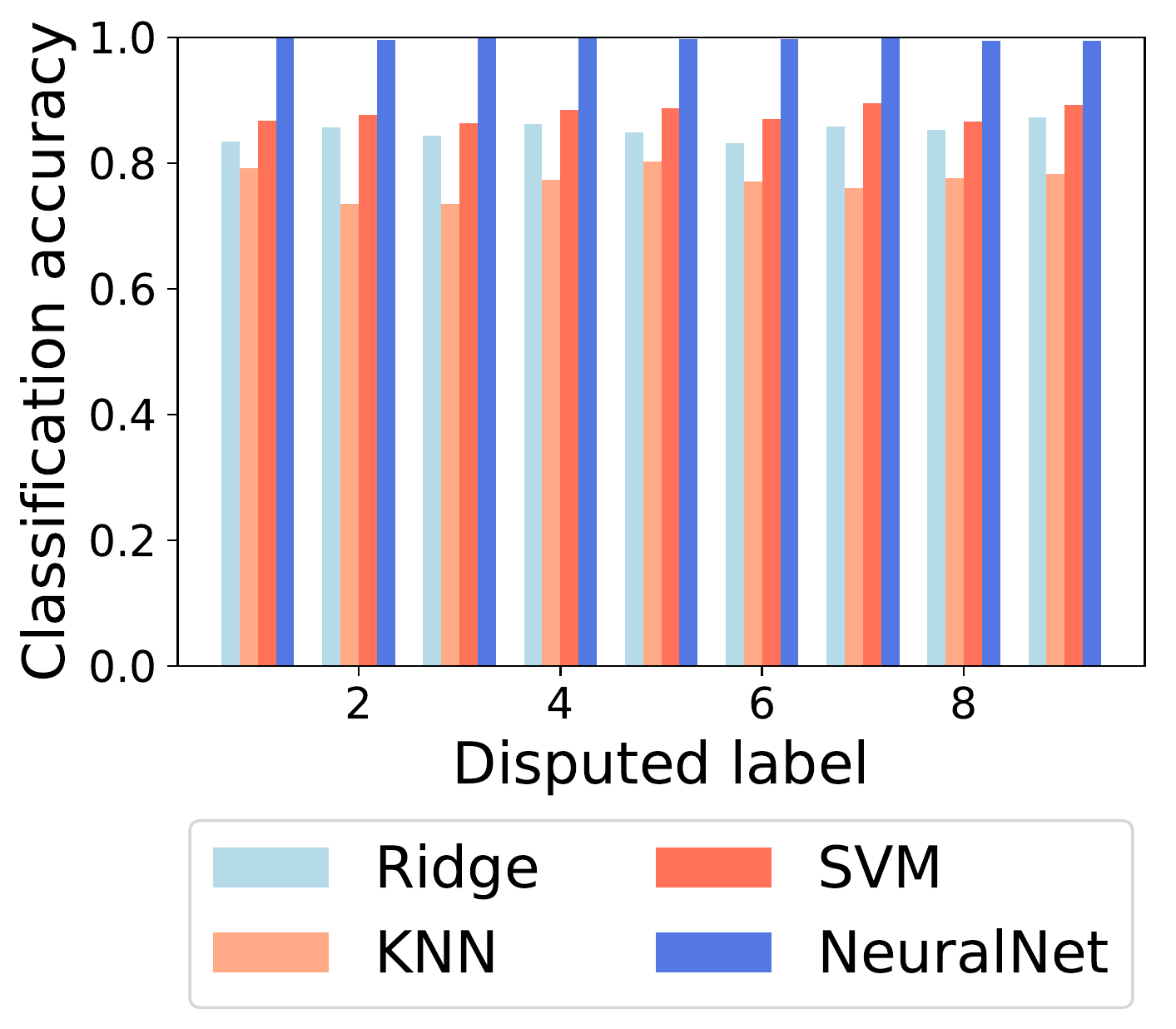}
            } &
       \subfloat[Test for word range $800-900$]{%
              \includegraphics[width=0.25\linewidth]{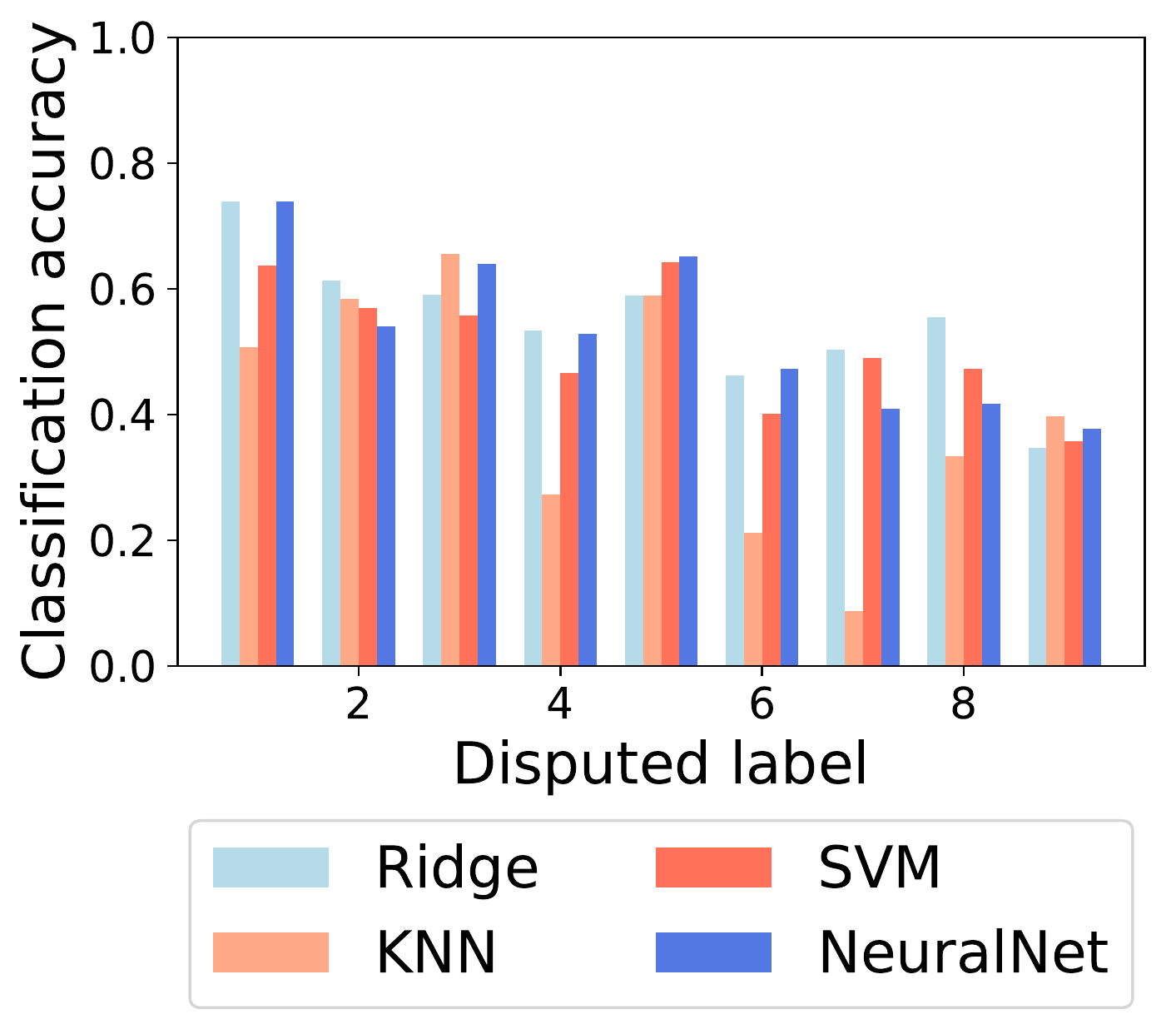}} &
       \subfloat[Train for word range $900-1000$]{%
              \includegraphics[width=0.25\linewidth]{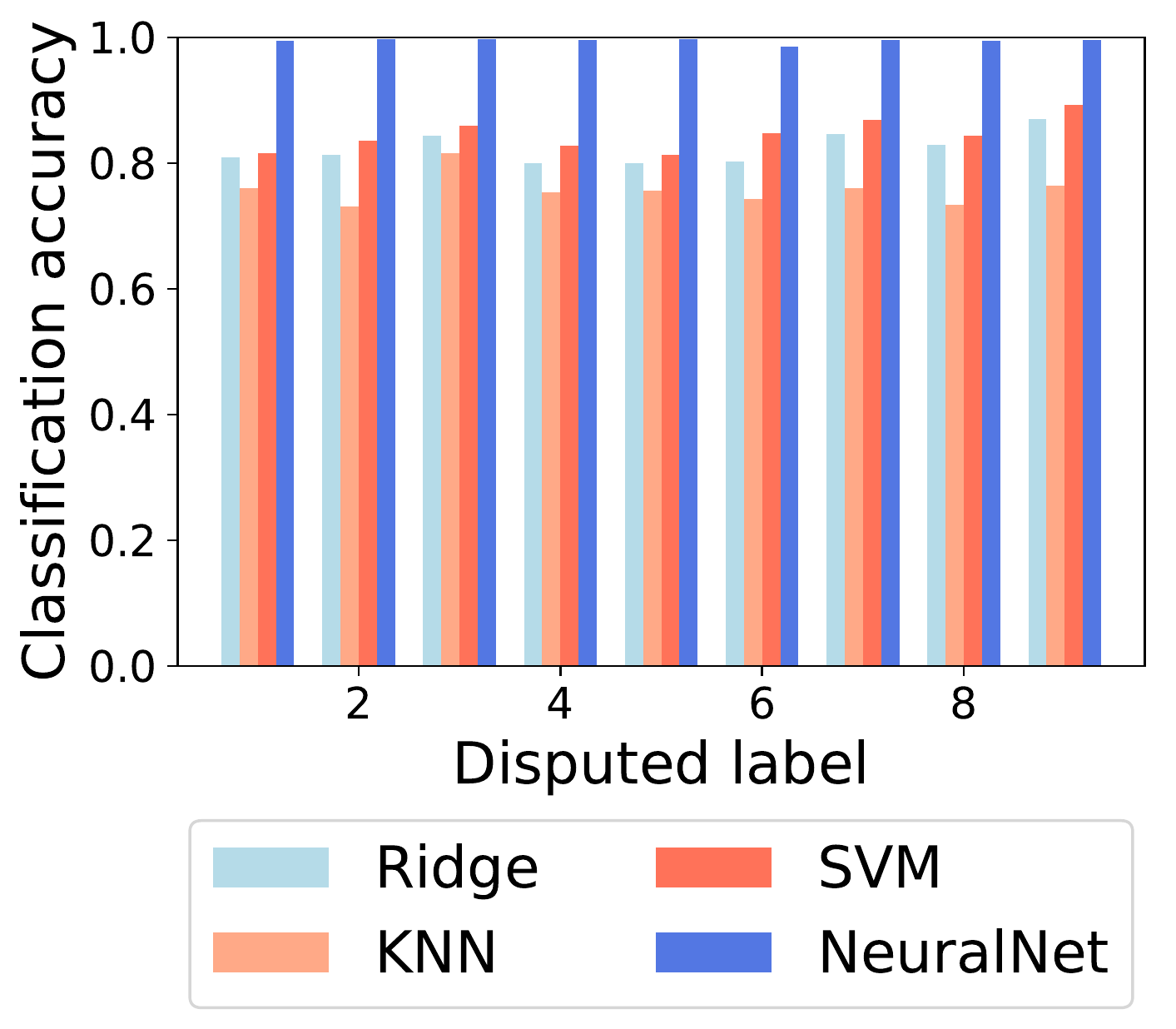}
            } &
       \subfloat[Test for word range $900-1000$]{ \includegraphics[width=0.25\linewidth]{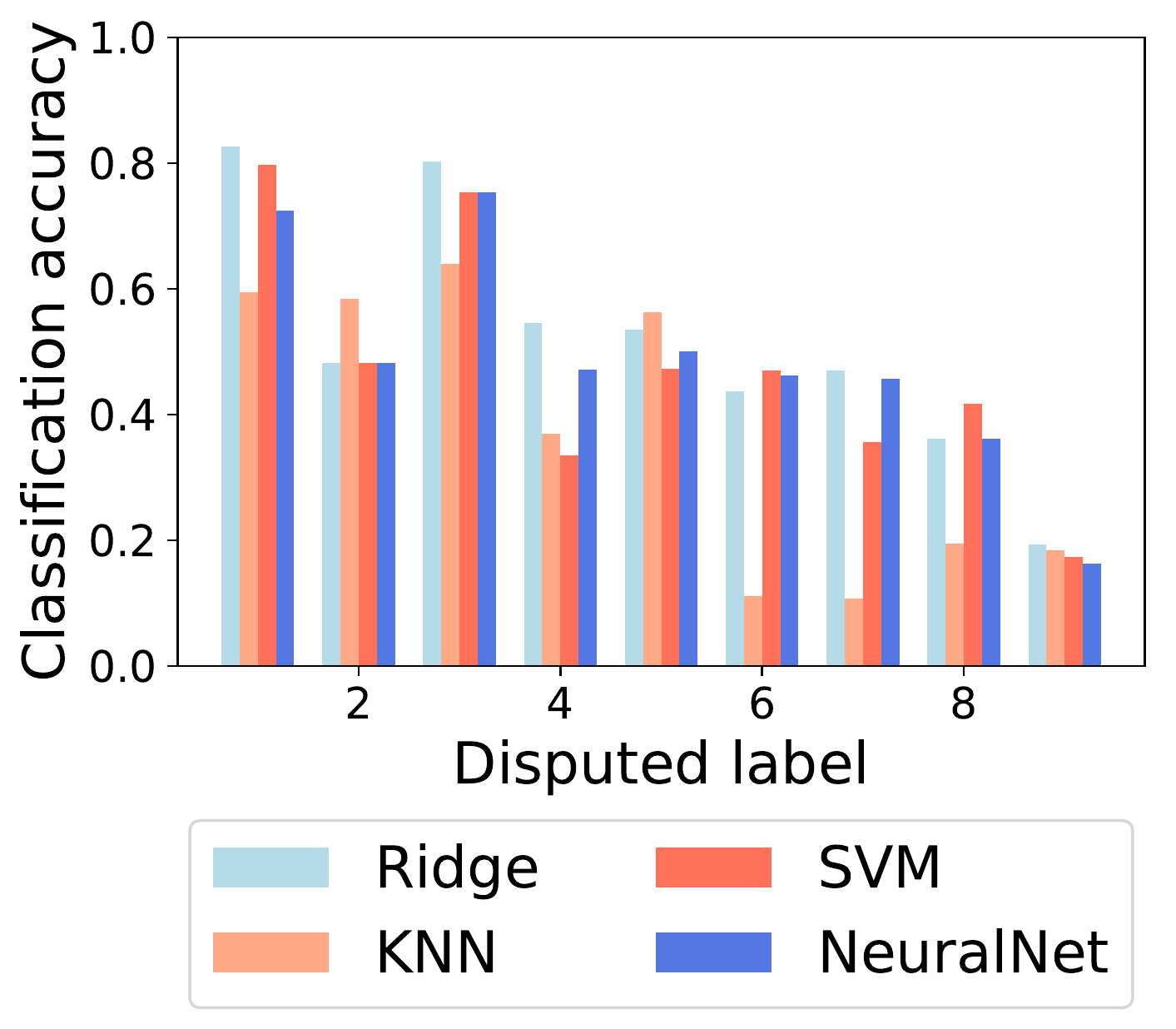}}\\
       \end{tabular}
       \caption{\textbf{Cross-validation for Groups A and B for different word intervals.} On the x-axis we report the text used as Disputed for that round. $1$ stands for {\itshape Lettres sonores} by Valéry Afanassiev, $2$ for  {\itshape Les Deux sœurs} by Vladimir Fédorovski, $3$ for {\itshape Éducation nocturne} by Luba Jurgenson, $4$ for  {\itshape Acné festival} by Iegor Gran, $5$ for {\itshape Testament français} by Andrei Makine, $6$ for {\itshape Germinal
       } by Émile Zola, $7$ for
       {\itshape Bouvard et Péuchet} by Gustave Flaubert, $8$ for {\itshape La maison du chat-qui-pelote } by Honoré de Balzac, $9$ for {\itshape Le Côté de Guermantes} by Marcel Proust. \label{fig:complete_cross_validation}}
       \end{figure}
       \begin{figure}
       \advance\leftskip-1cm
       \begin{tabular}{cccc}
       \subfloat[Train for word range $300-1300$]{%
              \includegraphics[width=0.25\linewidth]{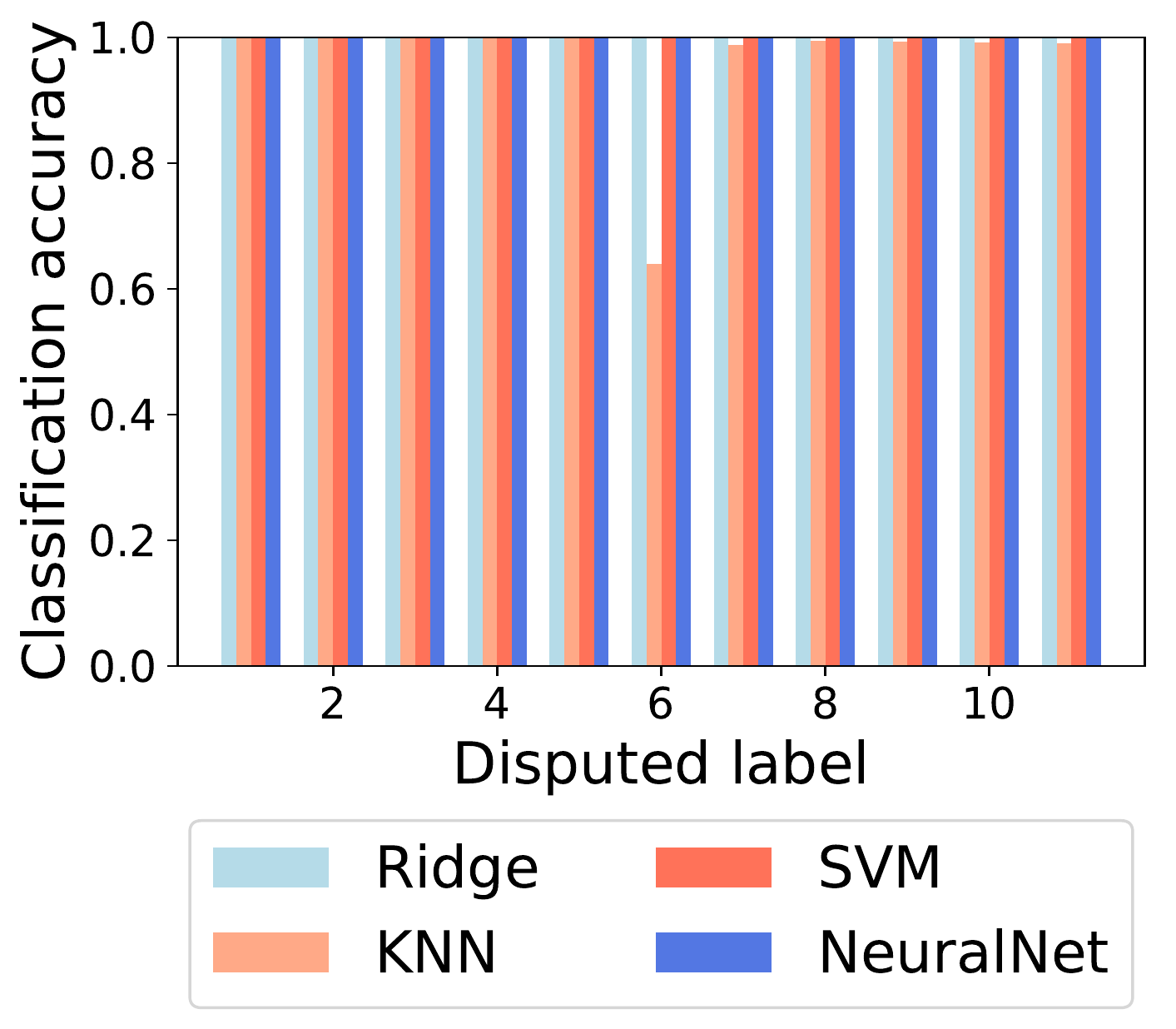}
            } &
       \subfloat[Test for word range $300-1300$]{%
              \includegraphics[width=0.25\linewidth]{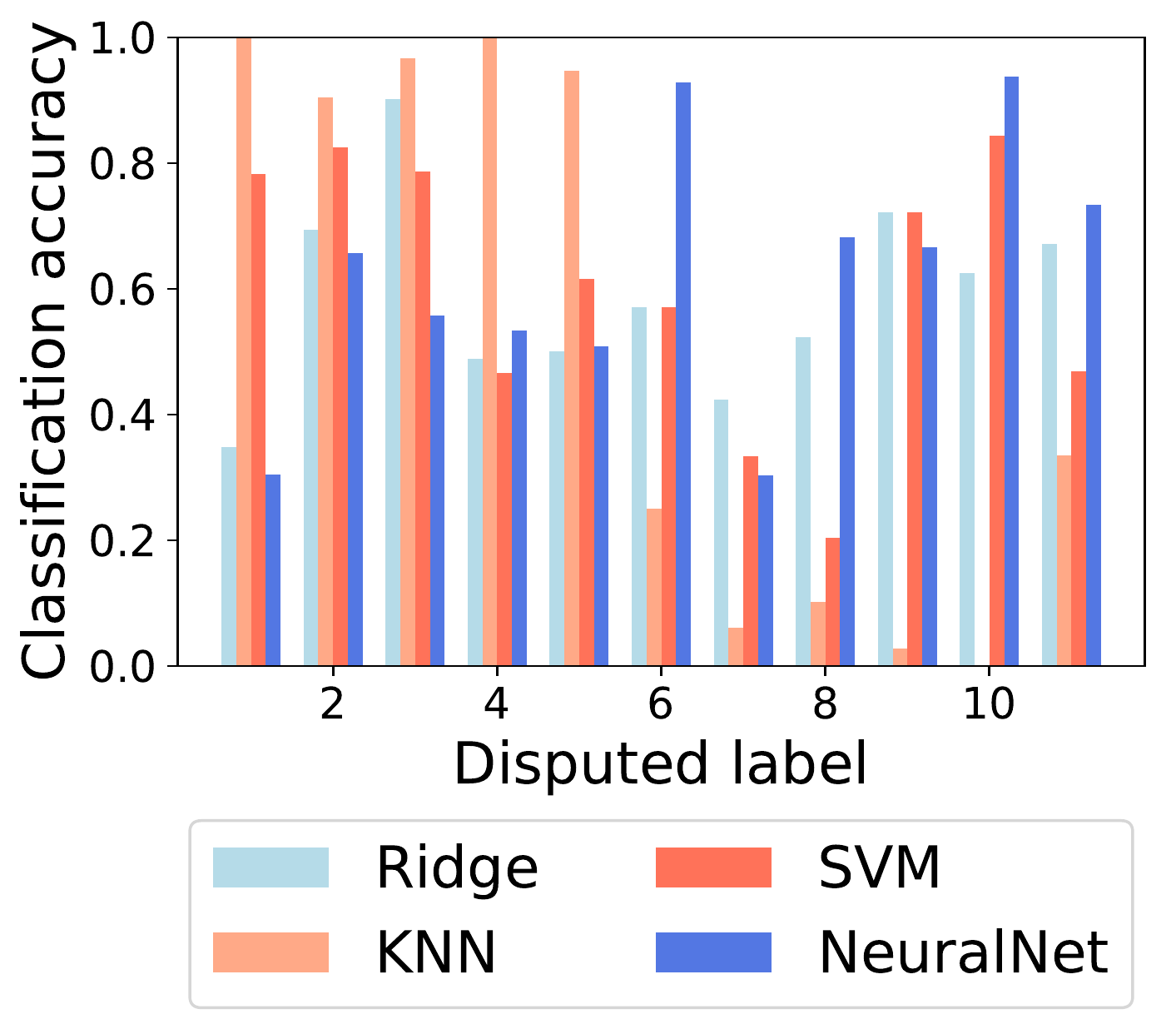}} &
       \subfloat[Train for word range $500-1000$]{%
              \includegraphics[width=0.25\linewidth]{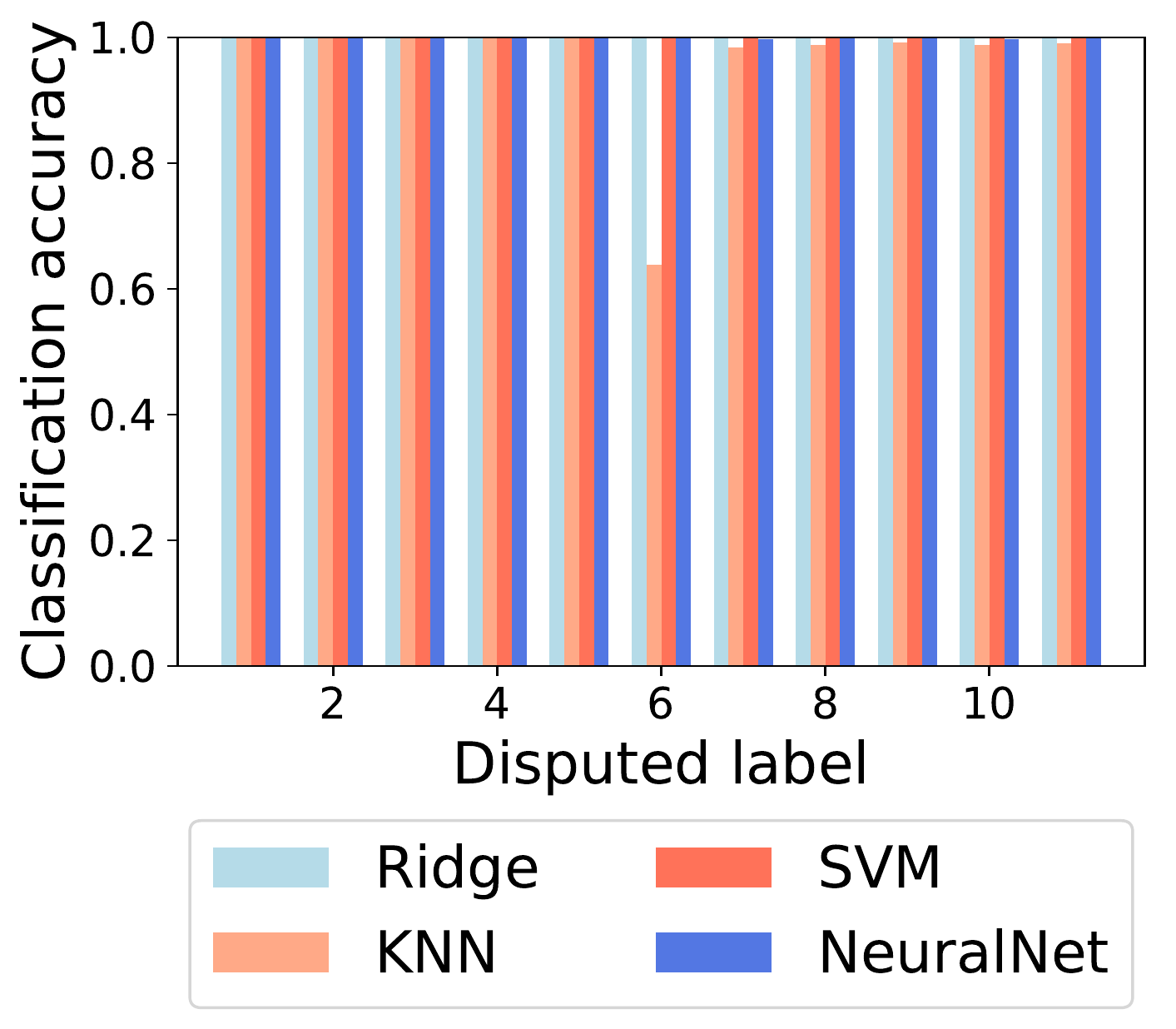}
            } &
       \subfloat[Test for word range $500-1000$]{%
              \includegraphics[width=0.25\linewidth]{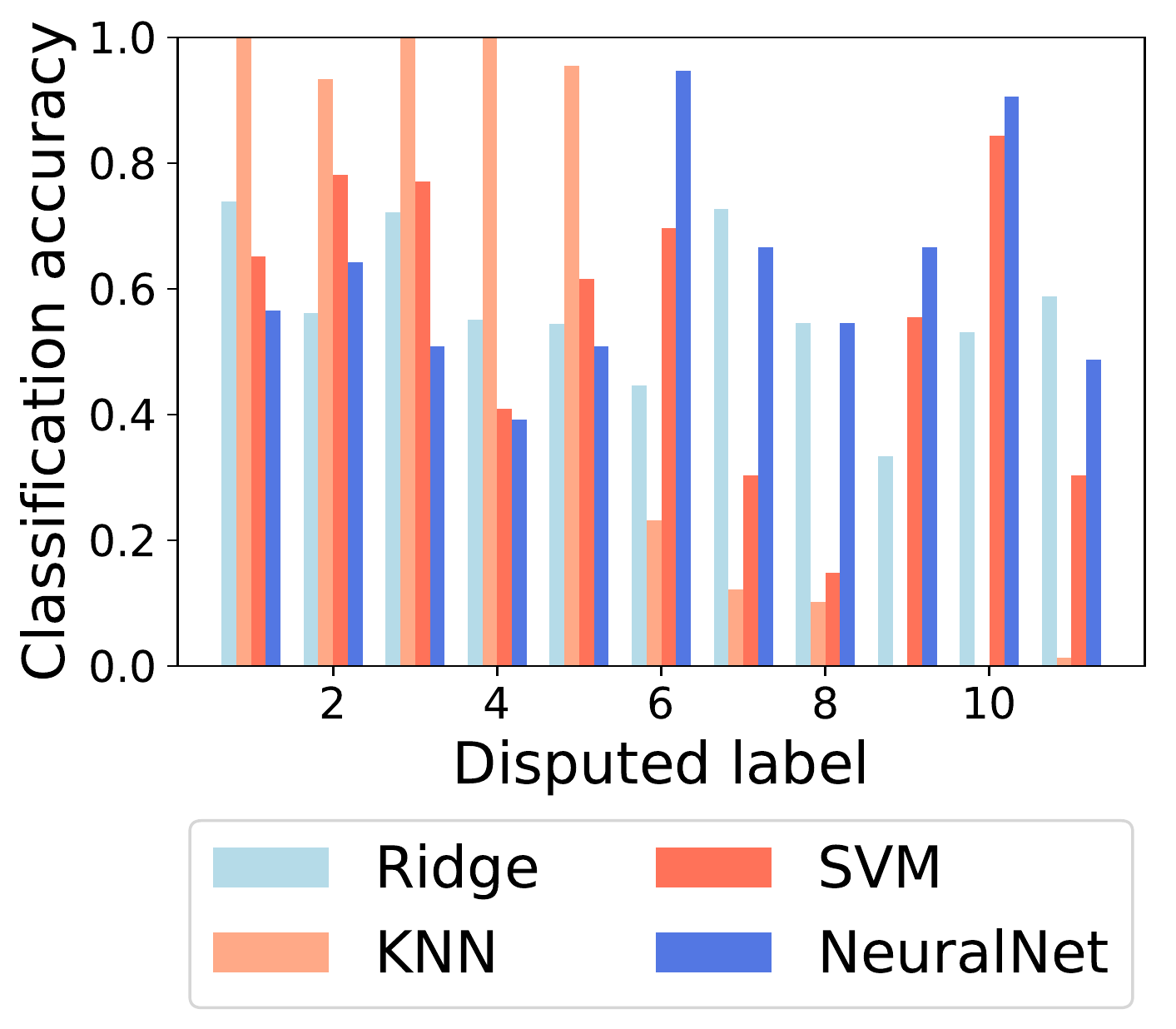}}\\
       \subfloat[Train for word range $500-600$]{%
              \includegraphics[width=0.25\linewidth]{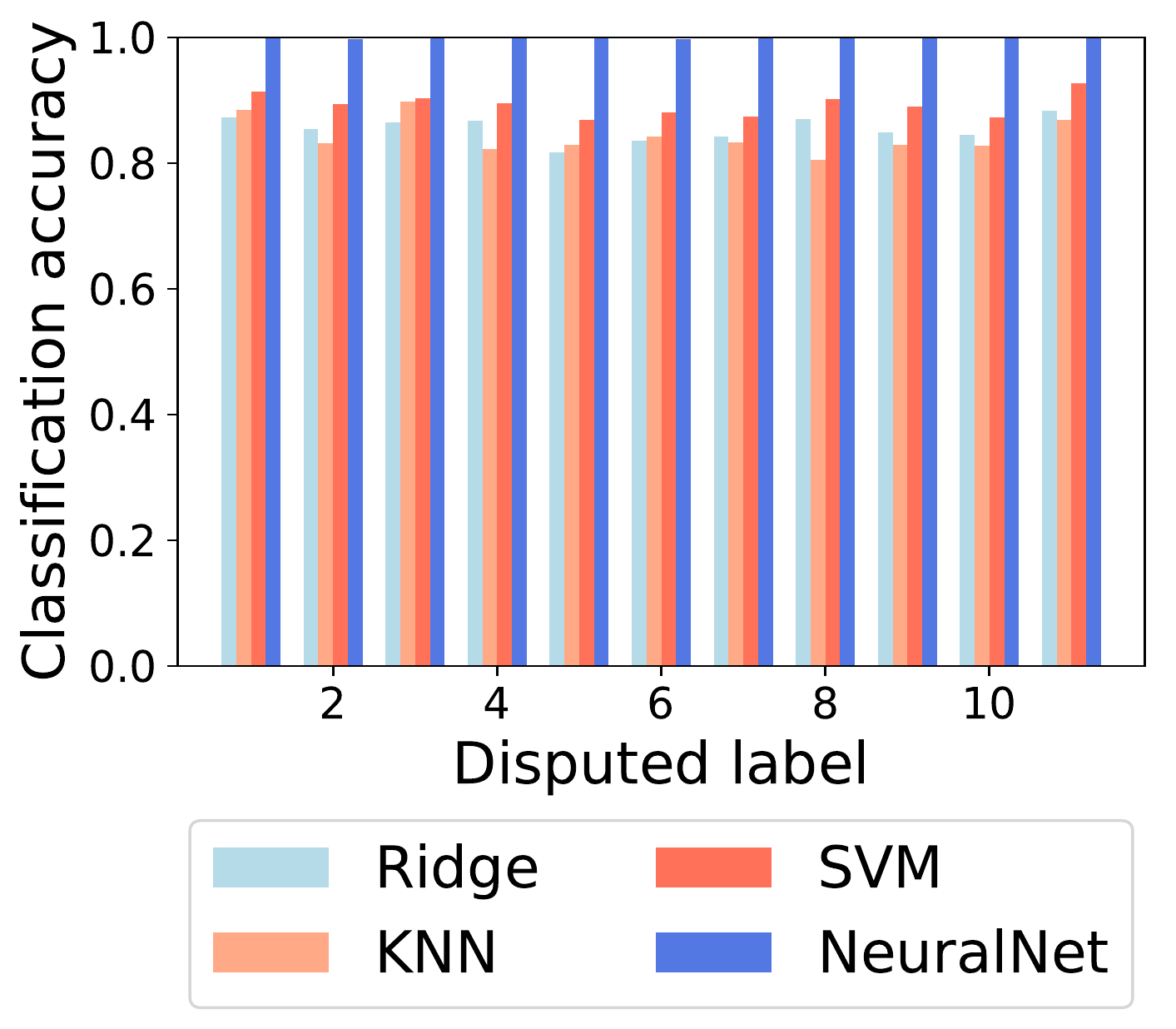}
            } &
       \subfloat[Test for word range $500-600$]{%
              \includegraphics[width=0.25\linewidth]{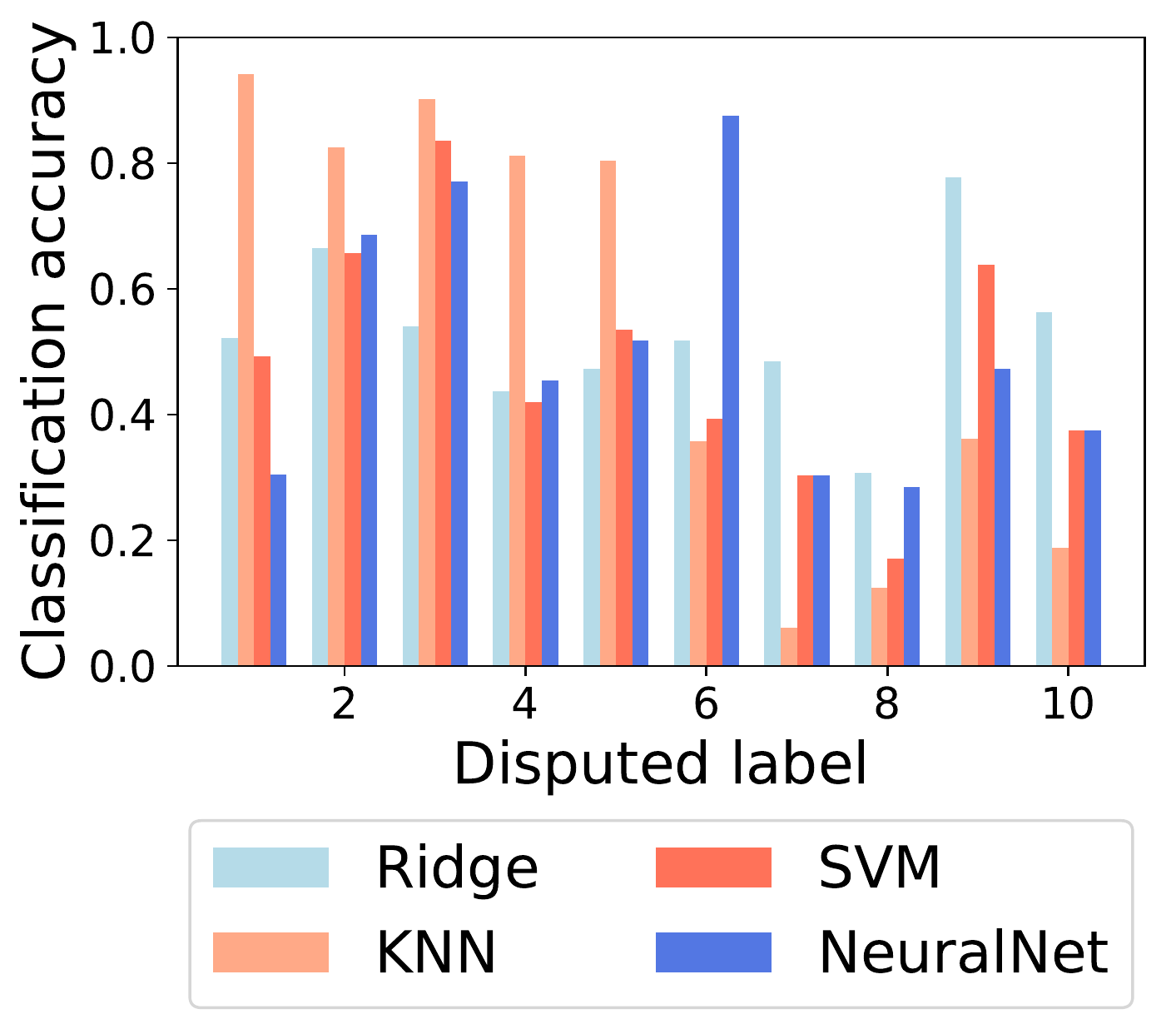}}&
       \subfloat[Train for word range $600-700$]{%
              \includegraphics[width=0.25\linewidth]{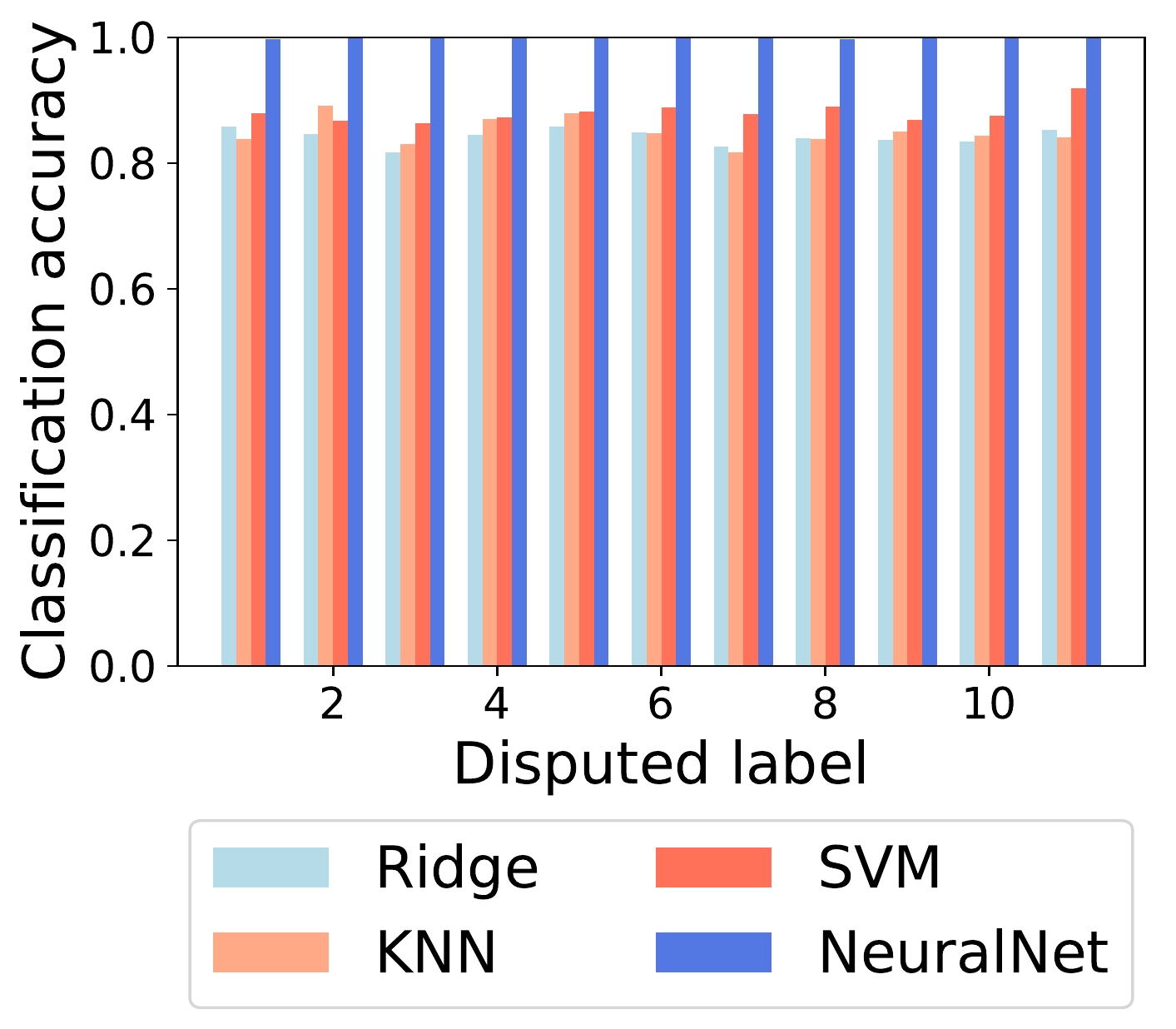}
            } &
       \subfloat[Test for word range $600-700$]{%
              \includegraphics[width=0.25\linewidth]{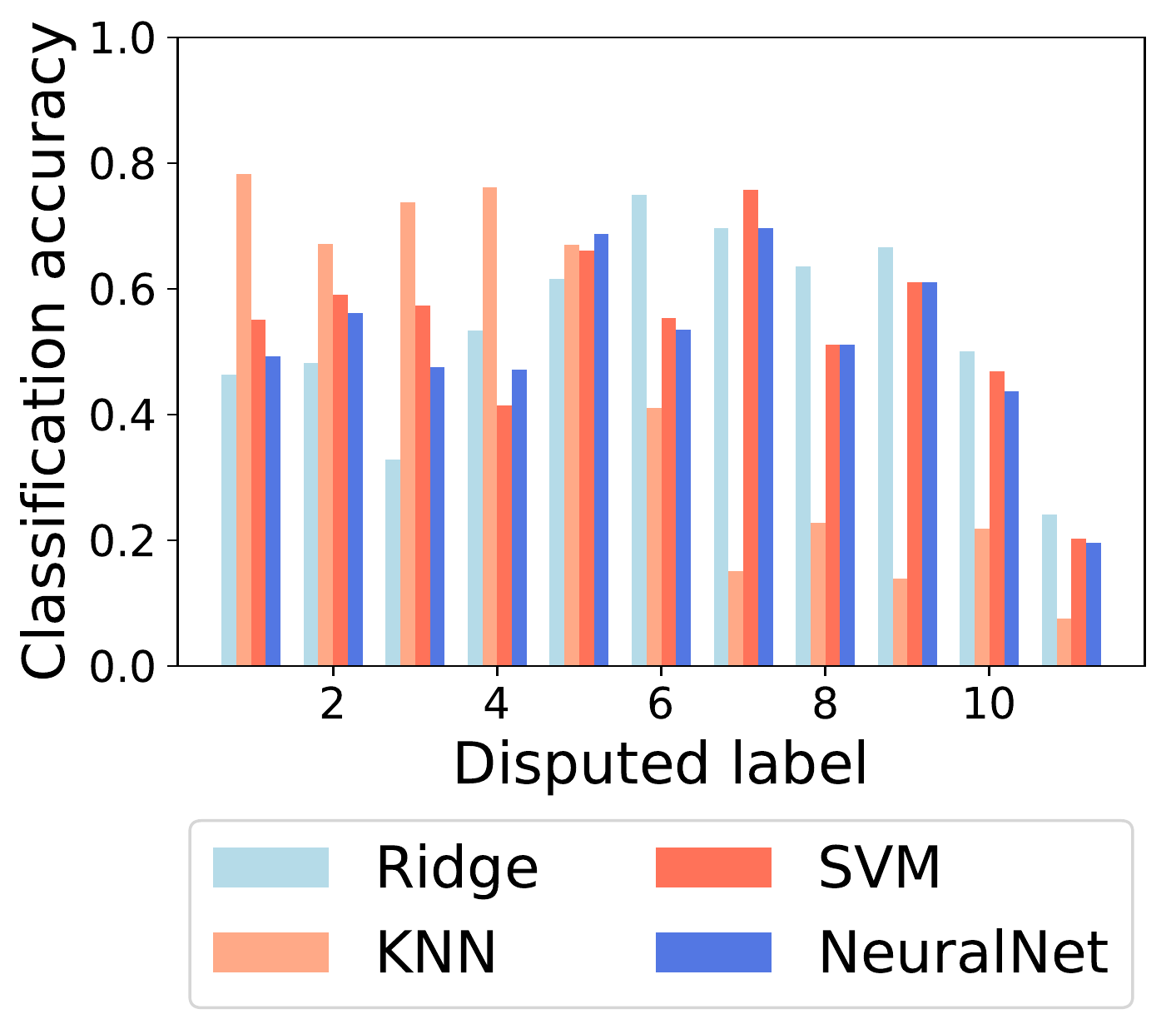}} \\
       \subfloat[Train for word range $700-800$]{%
              \includegraphics[width=0.25\linewidth]{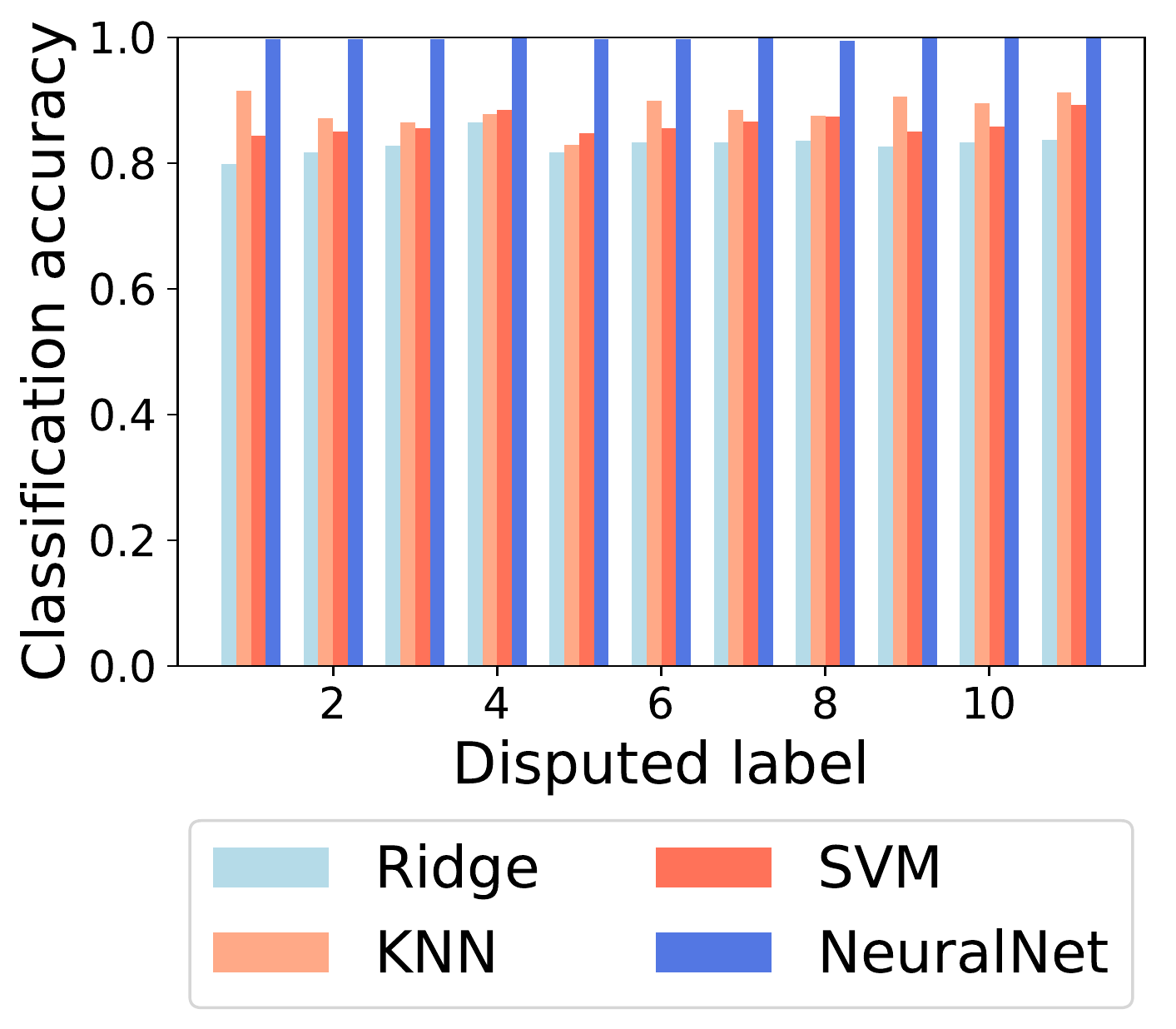}
            } &
       \subfloat[Test for word range $700-800$]{%
              \includegraphics[width=0.25\linewidth]{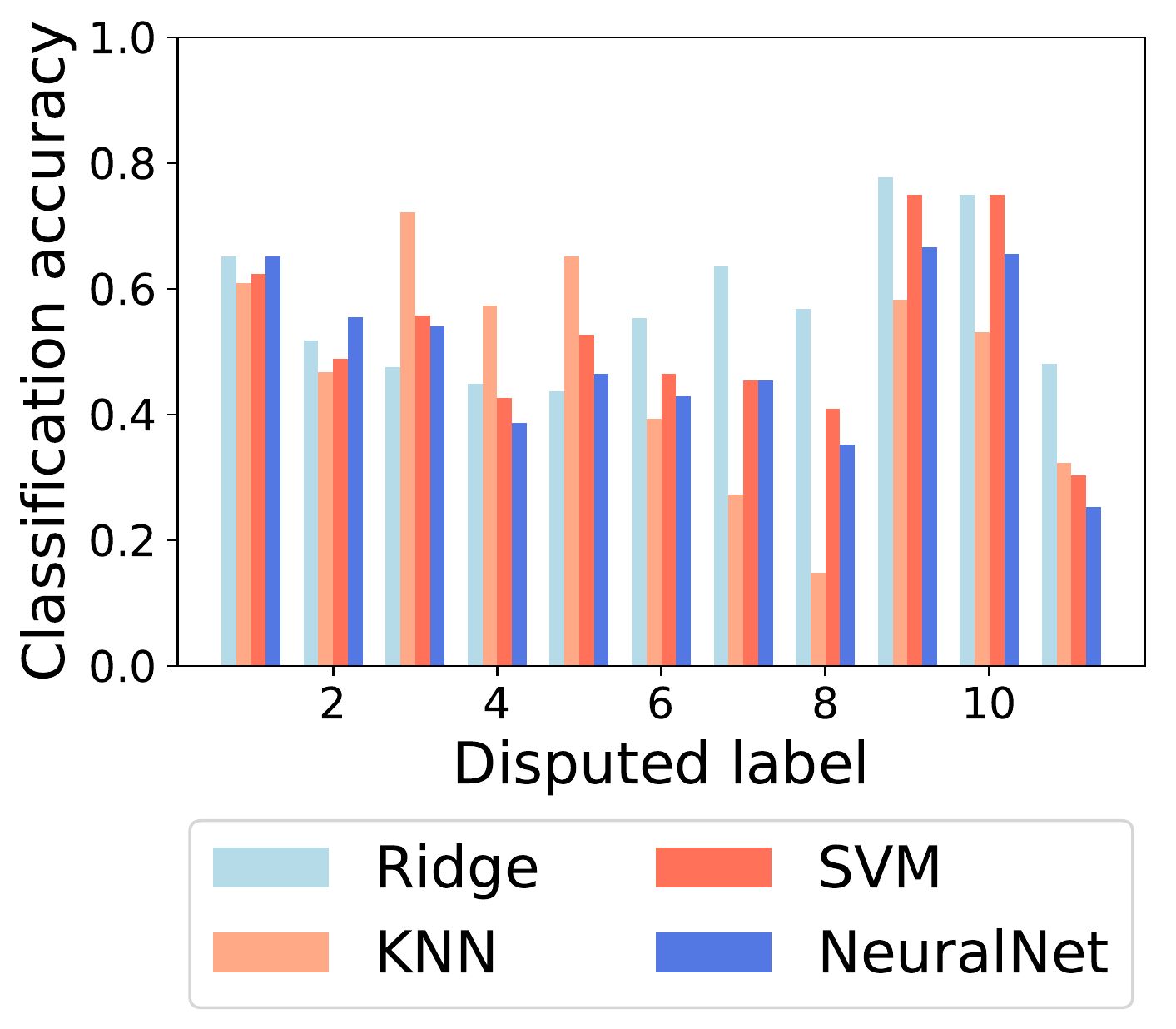}}&
       \subfloat[Train for word range $800-900$]{%
              \includegraphics[width=0.25\linewidth]{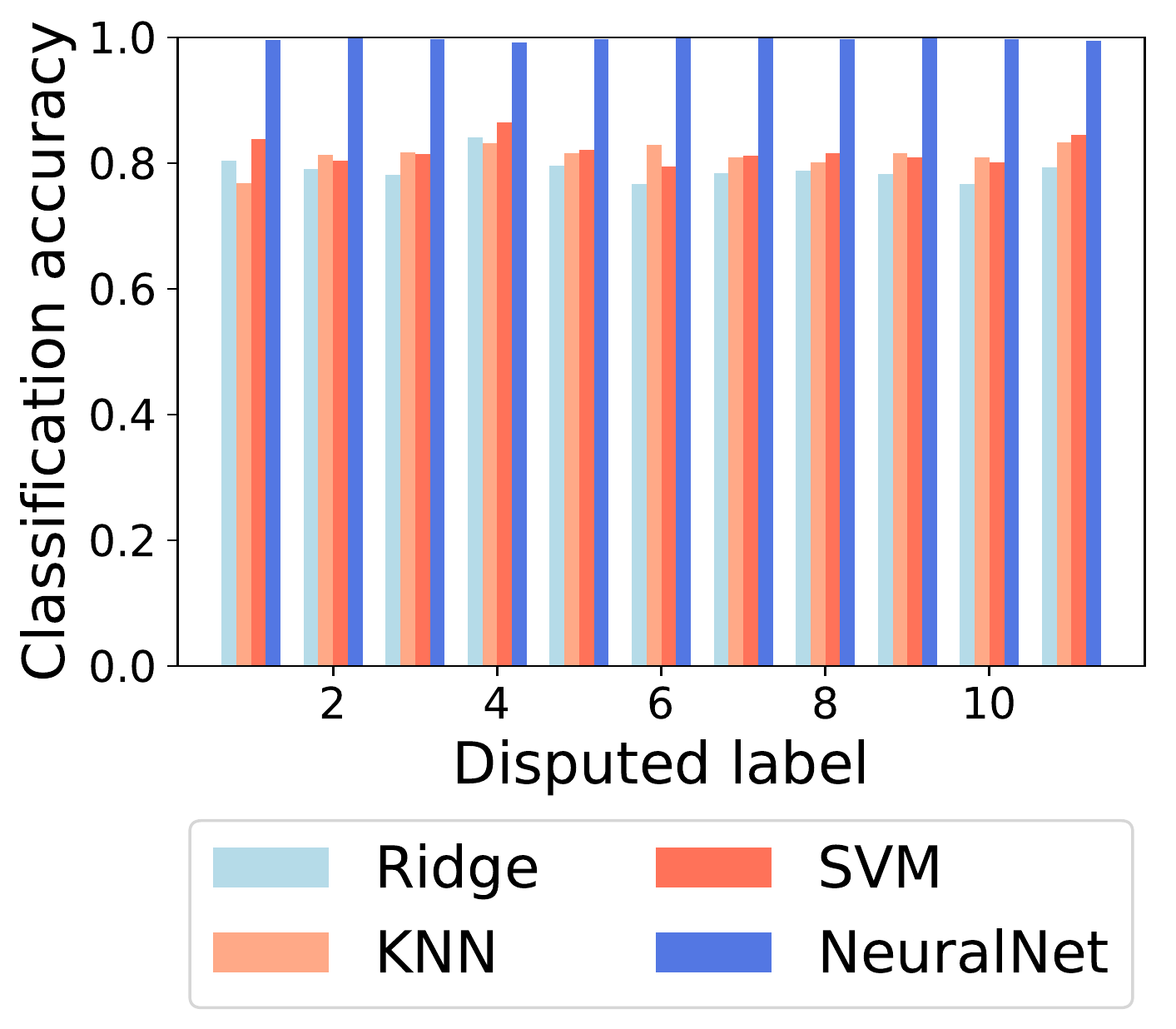}
            } &
       \subfloat[Test for word range $800-900$]{%
              \includegraphics[width=0.25\linewidth]{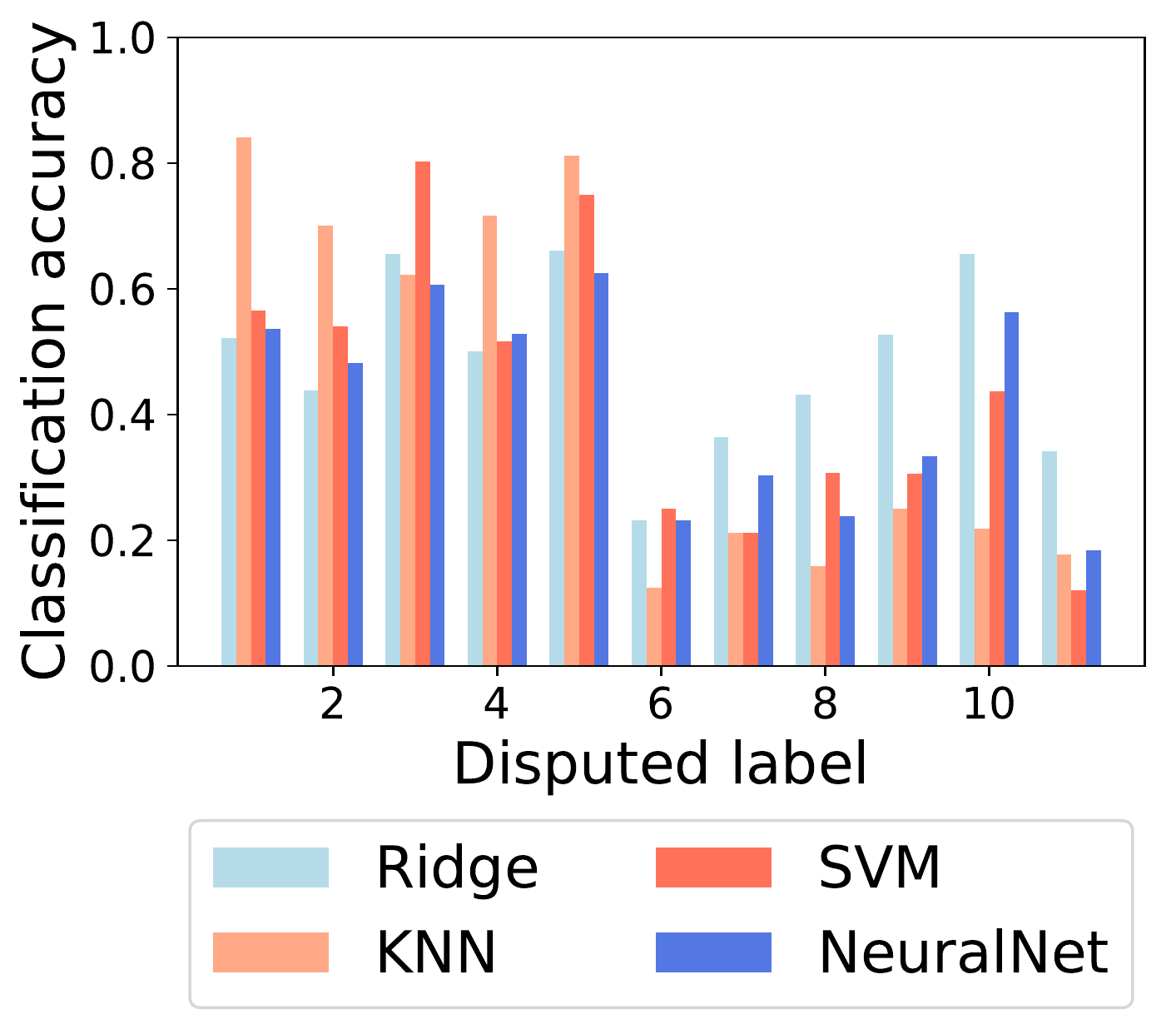}} \\
       \end{tabular}
       \caption{\textbf{Cross-validation for Groups A and C for different word intervals.} On the x-axis we report the text used as Disputed for that round. $1$ stands for {\itshape Lettres sonores} by Valéry Afanassiev, $2$ for  {\itshape Les Deux sœurs} by Vladimir Fédorovski, $3$ for {\itshape Éducation nocturne} by Luba Jurgenson, $4$ for  {\itshape Acné festival} by Iegor Gran, $5$ for {\itshape Testament français} by Andrei Makine, $6$ for {\itshape Du plus loin de l’oubli} by Patrick Modiano, $7$ for
       {\itshape Terrasse à Rome} by Pascal Quignard, $8$ for {\itshape L’Amant de la Chine du Nord} by Marguerite Duras, $9$ for {\itshape La Honte} by Annie Ernaux, $10$ for {\itshape L’Evénement} by Annie Ernaux, $11$ for {\itshape Les Particules élémentaires} by Michel Houellebecq. \label{fig:modern_cross_validation}}
       \end{figure}

\subsubsection{Cross-validation: Ridge, SVM, KNN, Neural Network}

In this section we evaluate the cross-validation accuracy for the Ridge and SVM methods that were discussed earlier, as well as for $K$-Nearest Neighbors (KNN) and the Neural Network not applied before this section. The goal is to find the method with the best accuracy for classifying authors into two groups.

First, we consider the classification accuracy of the texts of the Russian-French group, the texts of French-French classics, and Disputed (Figure~\ref{fig:complete_cross_validation}).

The highest average cross-validation accuracy was achieved with the Neural Network, 77\% on the range from 300 to 800 of the most frequent words. SVM also performs well scoring 69\% cross-validation accuracy on the same segment. Ridge cross-validation accuracy is 61\%. Furthermore, we noticed that non-parametric models such as KNN perform poorly, with KNN cross-validation accuracy slightly above 50\%. We noticed that Ridge cross-validation accuracy is higher for a small range (e.g., 100 words). It reaches 65\% for a range of 100 words (500 to 600), while the accuracy of SVM and Neural Network is 61\% and 63\%.

Second, we perform the classification of the texts of the Russian-French authors, French-French contemporary writers and Disputed (Figure \ref{fig:modern_cross_validation}). The best average accuracy of cross-validation (also 77\%) was achieved by the Neural Network, as in the case of the French-French classic authors, but on a different range: 300-1300 words. Ridge and SVM show similar results, reaching 58\% and 60\%, while KNN is again the worst model with about 50\% accuracy.

Often the accuracy during model training is higher, so we can conclude that it seems difficult to avoid overfitting on the training dataset. This can be seen, for example, when comparing graphs i and j (Figure~\ref{fig:complete_cross_validation}) or g and h (Figure~\ref{fig:modern_cross_validation}). To avoid overfitting, it is necessary to have more data, at least about twenty books in each group, and to use a supercomputer, whose computational speed exceeds that of a standard computer and to which we do not have access.

\subsubsection{Hidden layers of Neural Network}

In this section we use Neural Network to classify authors in two groups without searching for interference. As proved in the previous section, classification with Neural Network method has the highest accuracy.

\begin{figure}
\centering
\begin{tabular}{ccc}
\subfloat[Disputed 1]{%
       \includegraphics[scale=0.28]{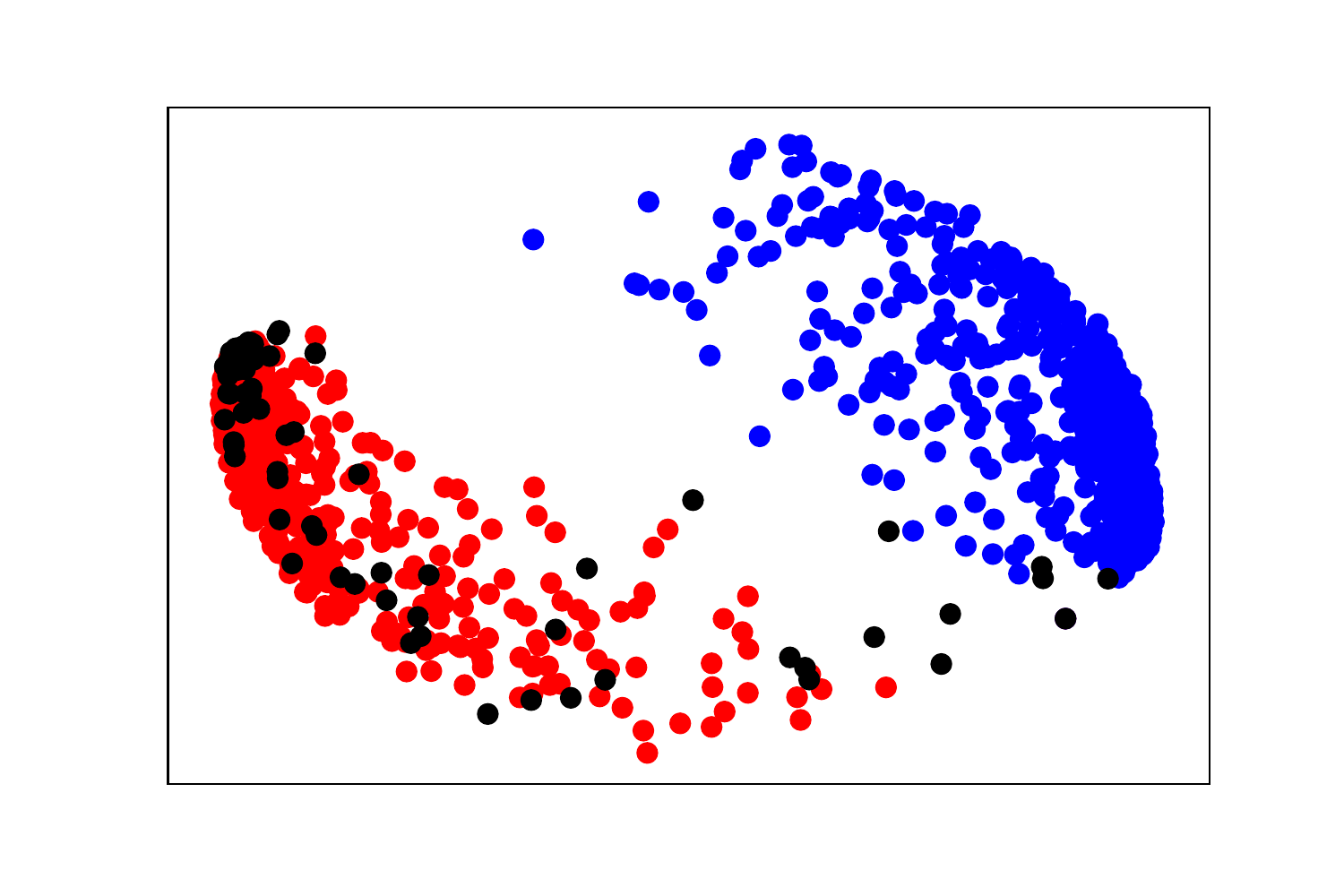}
     } &
\subfloat[Disputed 2]{%
       \includegraphics[scale=0.28]{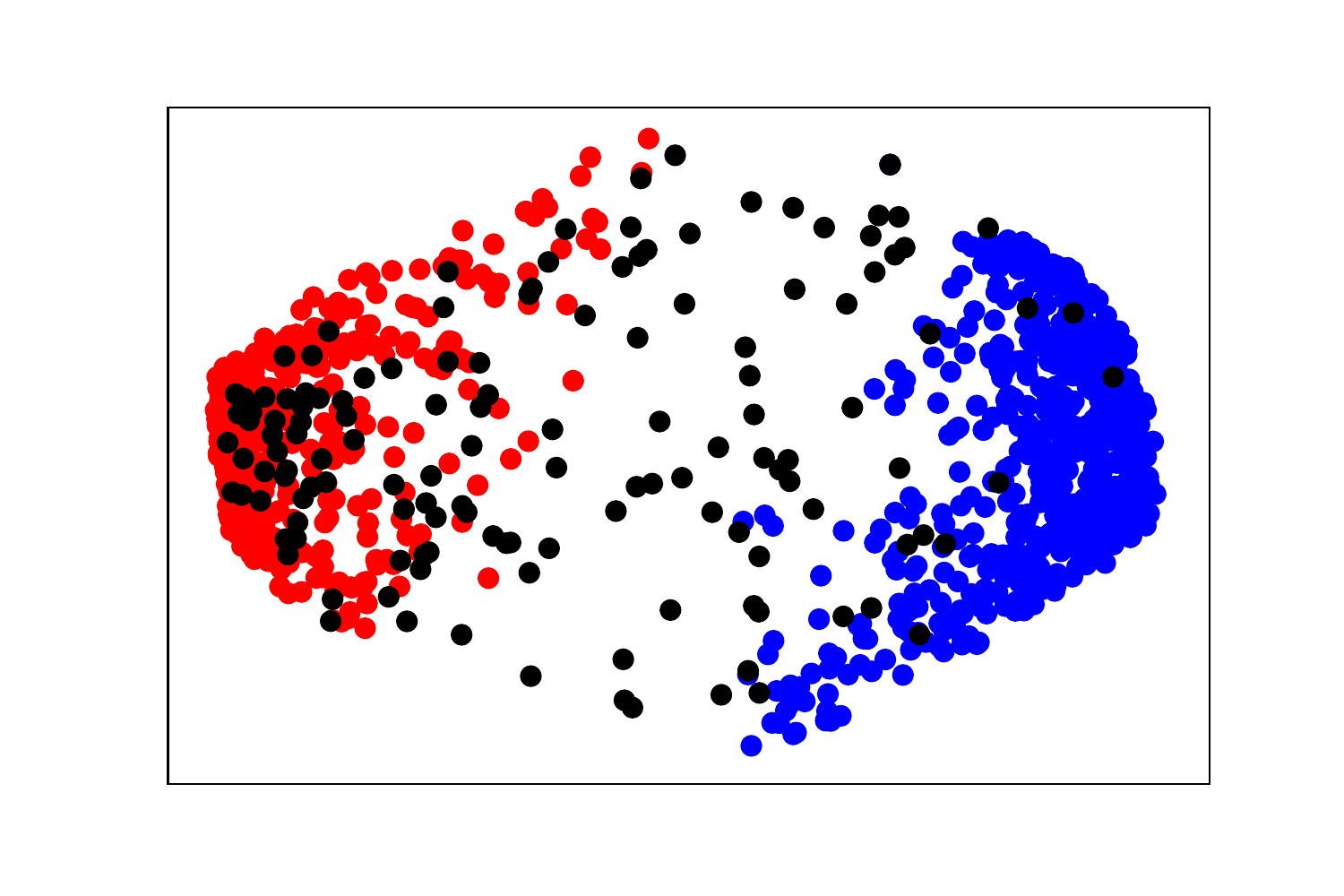}} &
\subfloat[Disputed 3]{%
       \includegraphics[scale=0.28]{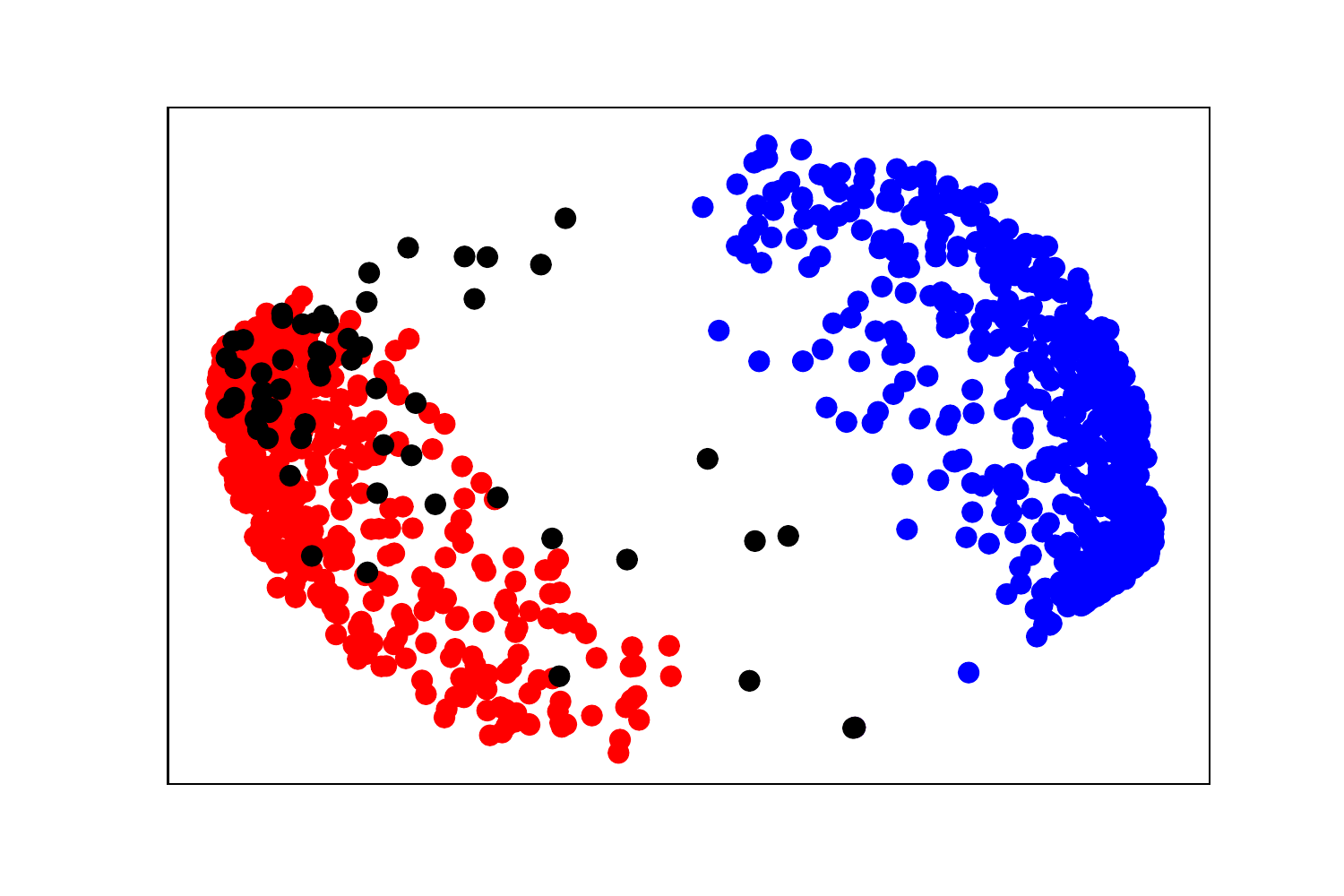}
     } \\
\subfloat[Disputed 4]{%
       \includegraphics[scale=0.28]{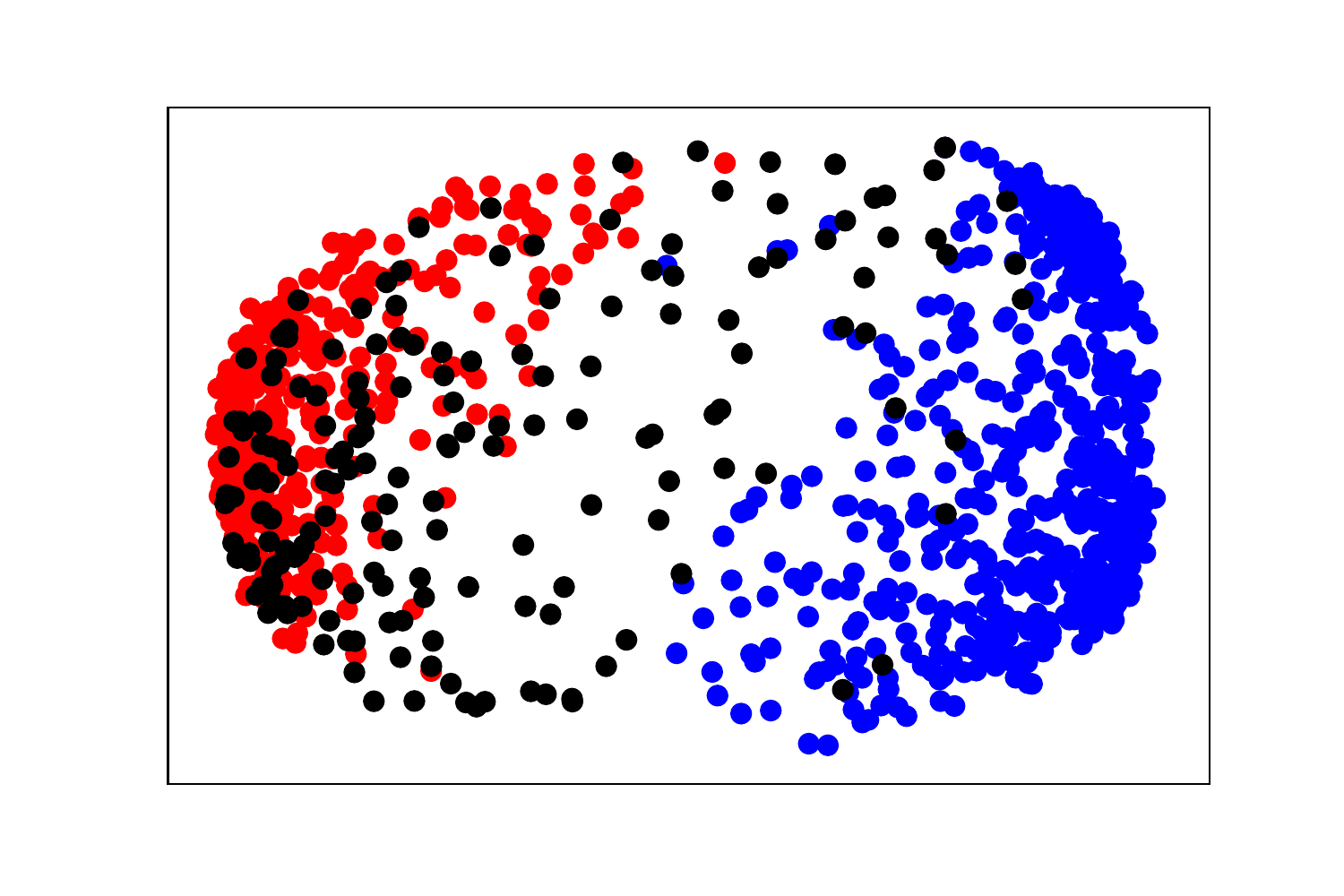}}&
\subfloat[Disputed 5]{%
       \includegraphics[scale=0.28]{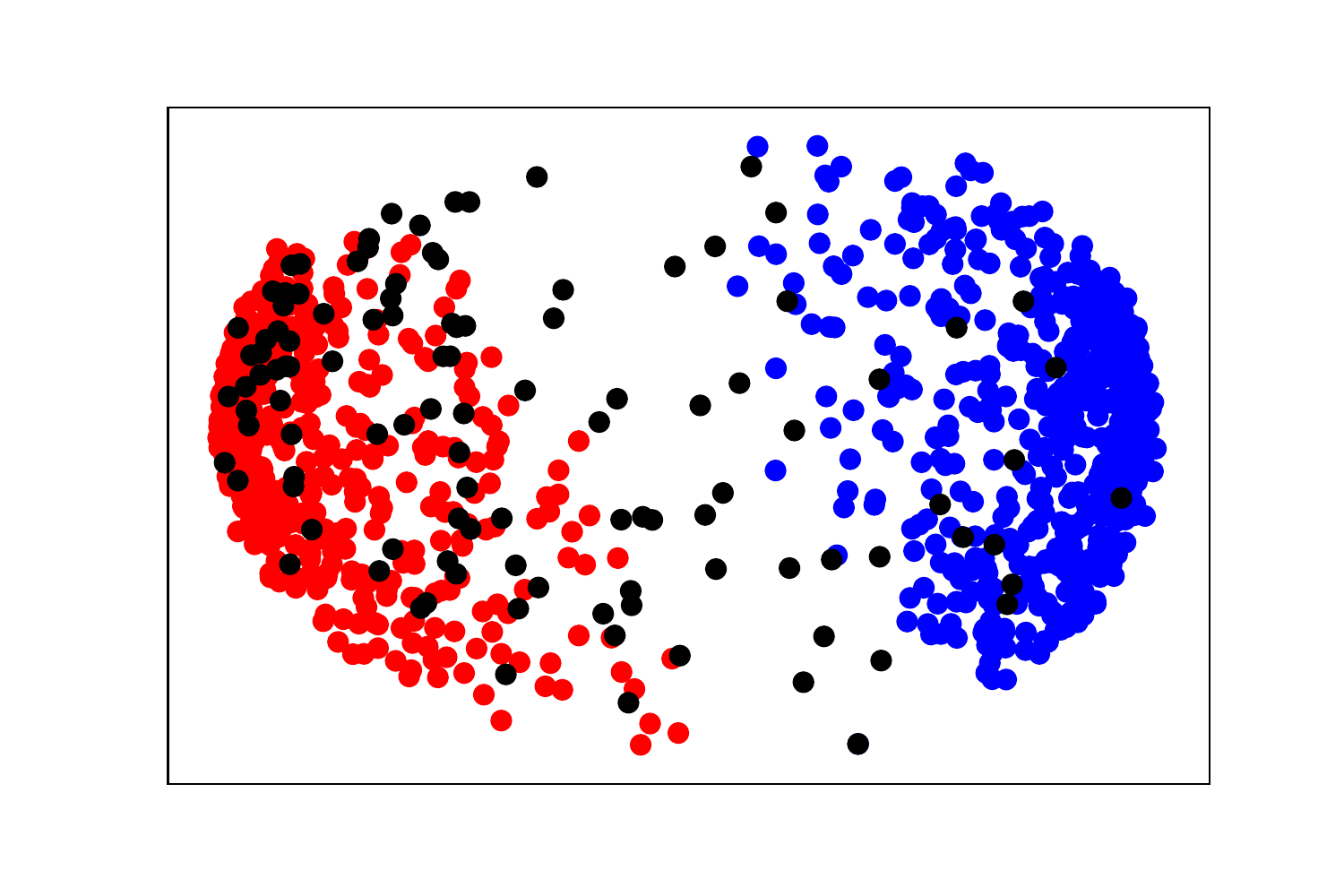}
     } &
\subfloat[Disputed 6]{%
       \includegraphics[scale=0.28]{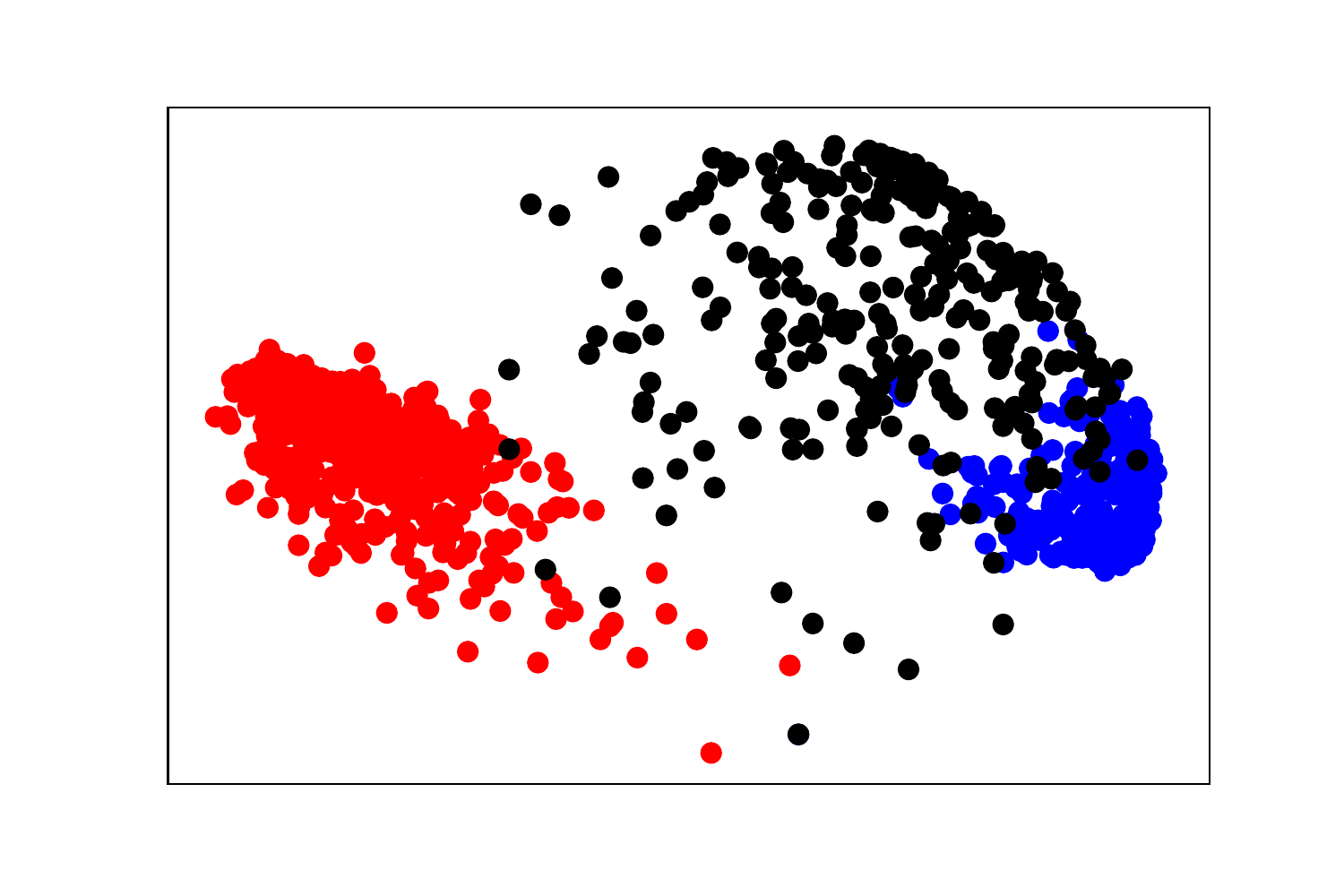}} \\
\subfloat[Disputed 7]{%
       \includegraphics[scale=0.28]{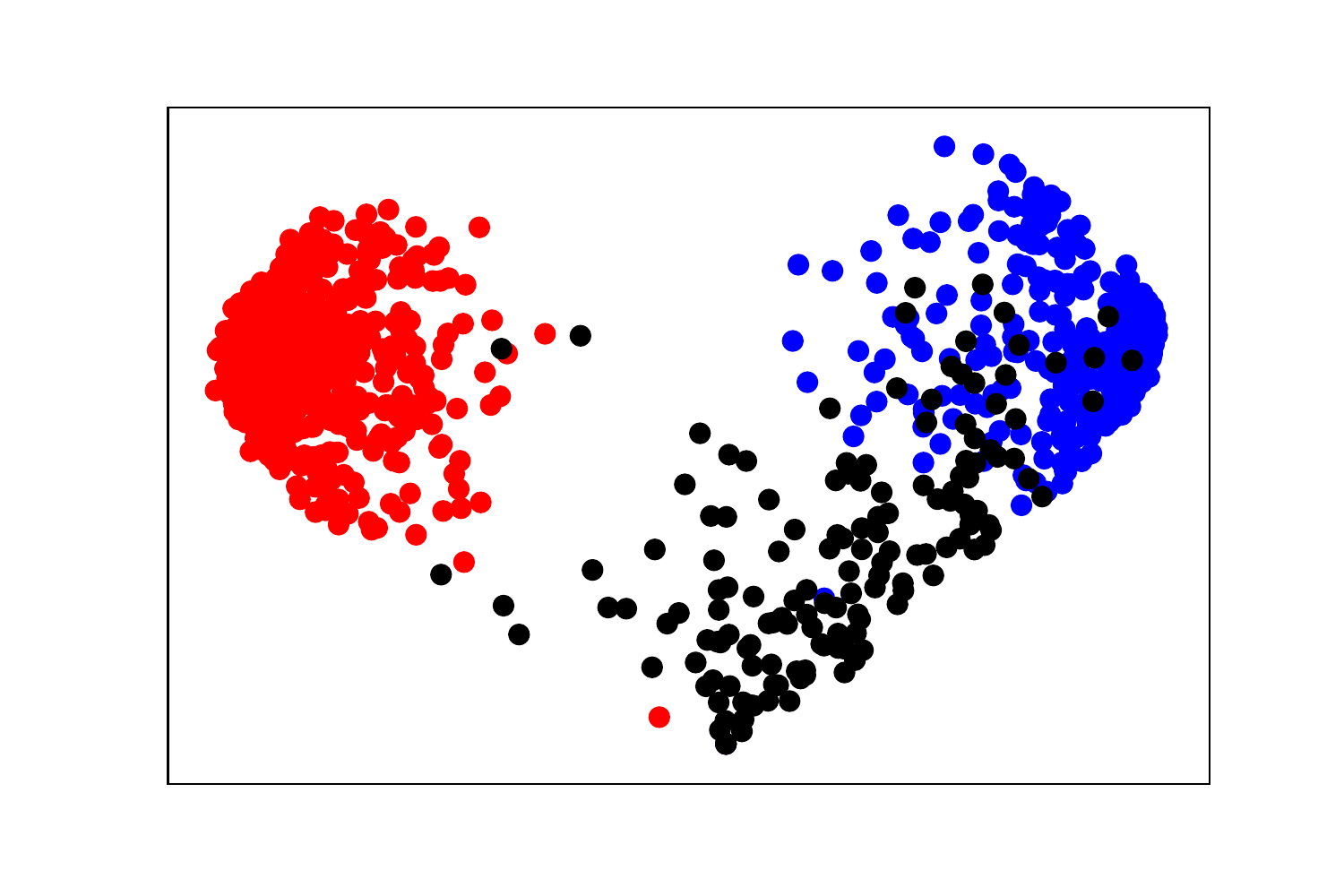}
     } &
\subfloat[Disputed 8]{%
       \includegraphics[scale=0.28]{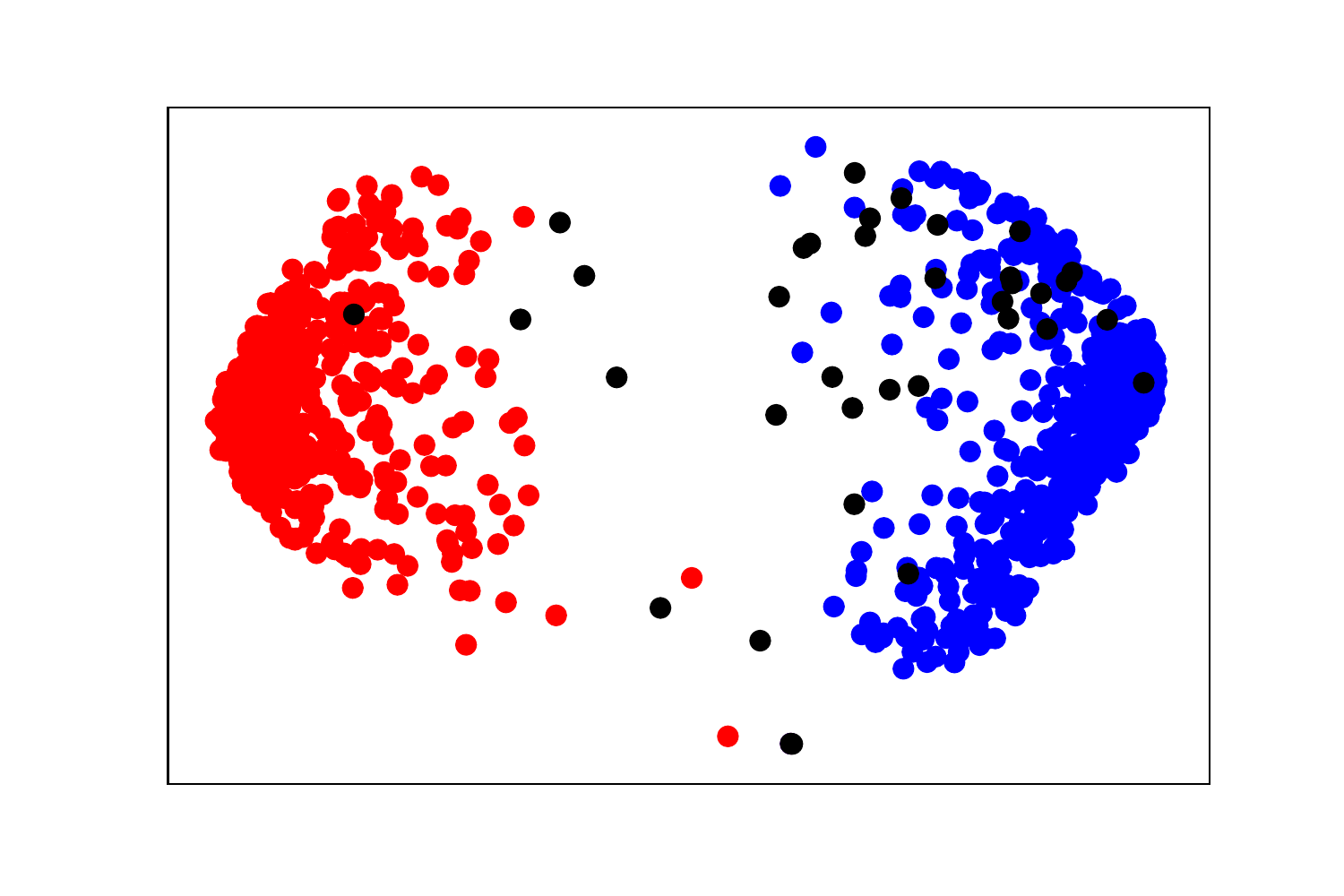}
     } &
\subfloat[Disputed 9]{%
       \includegraphics[scale=0.28]{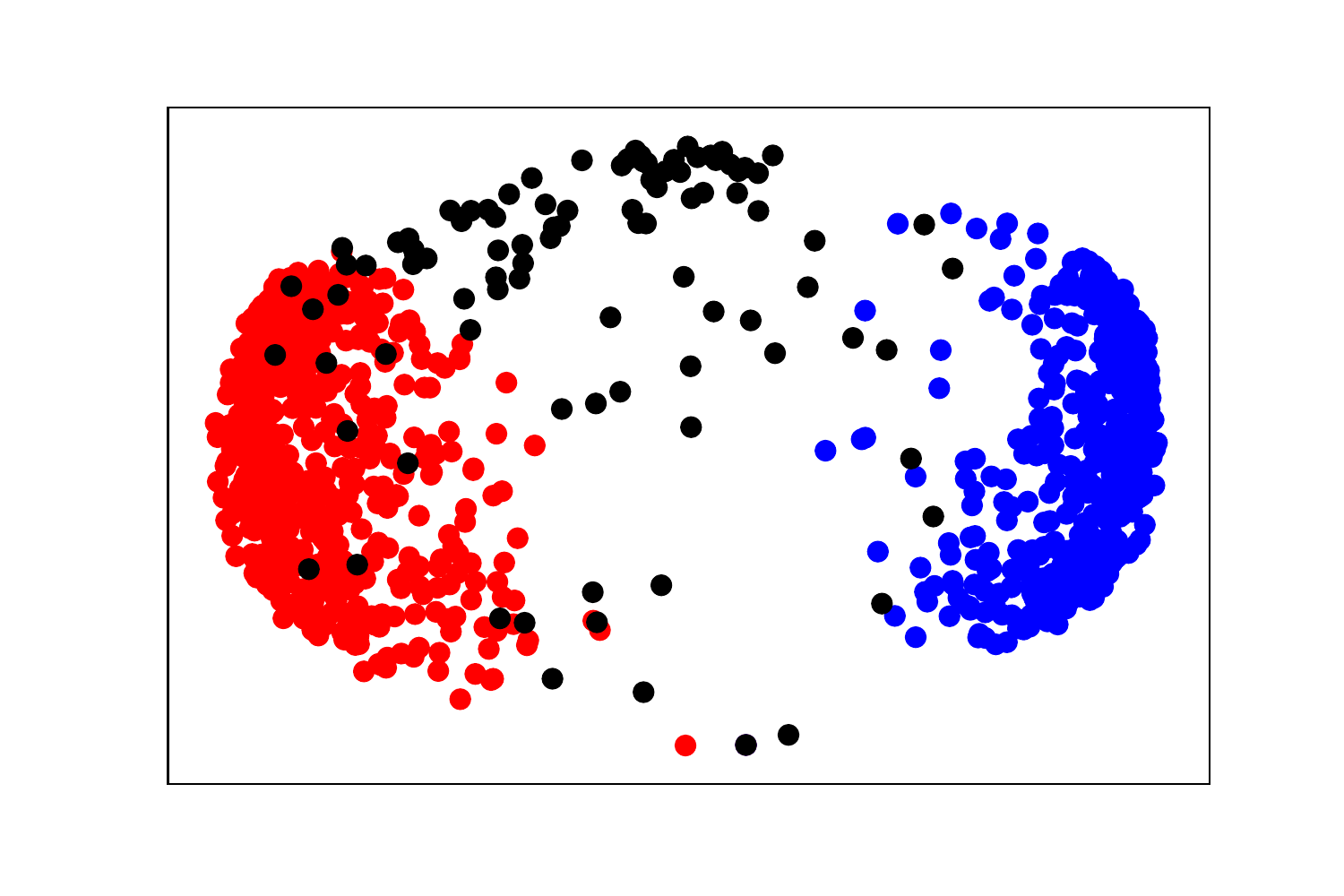}} \\
\end{tabular}
\caption{\textbf{First layer inspection for range 300-1300.} The Disputed id follows the mapping given in Figure \ref{fig:complete_cross_validation}. The plot is generated passing the text segments in the training set as input of the Neural Network and projecting the output of the first layer in a bi-dimensional space using PCA. We can see how the first layer learns to split the text segments in different groups: the red points represent the Russian-French group, while the blue dots represent the classic French-French group. The points in black are the text segments extracted from the Disputed text at that cross-validation round. We can see that the black points generally overlap with the points of the corresponding group. \label{fig:first_layer_300_1300}}
\end{figure}

\begin{figure}
\centering
\begin{tabular}{cccc}
\subfloat[Disputed 1]{%
       \includegraphics[scale=0.2]{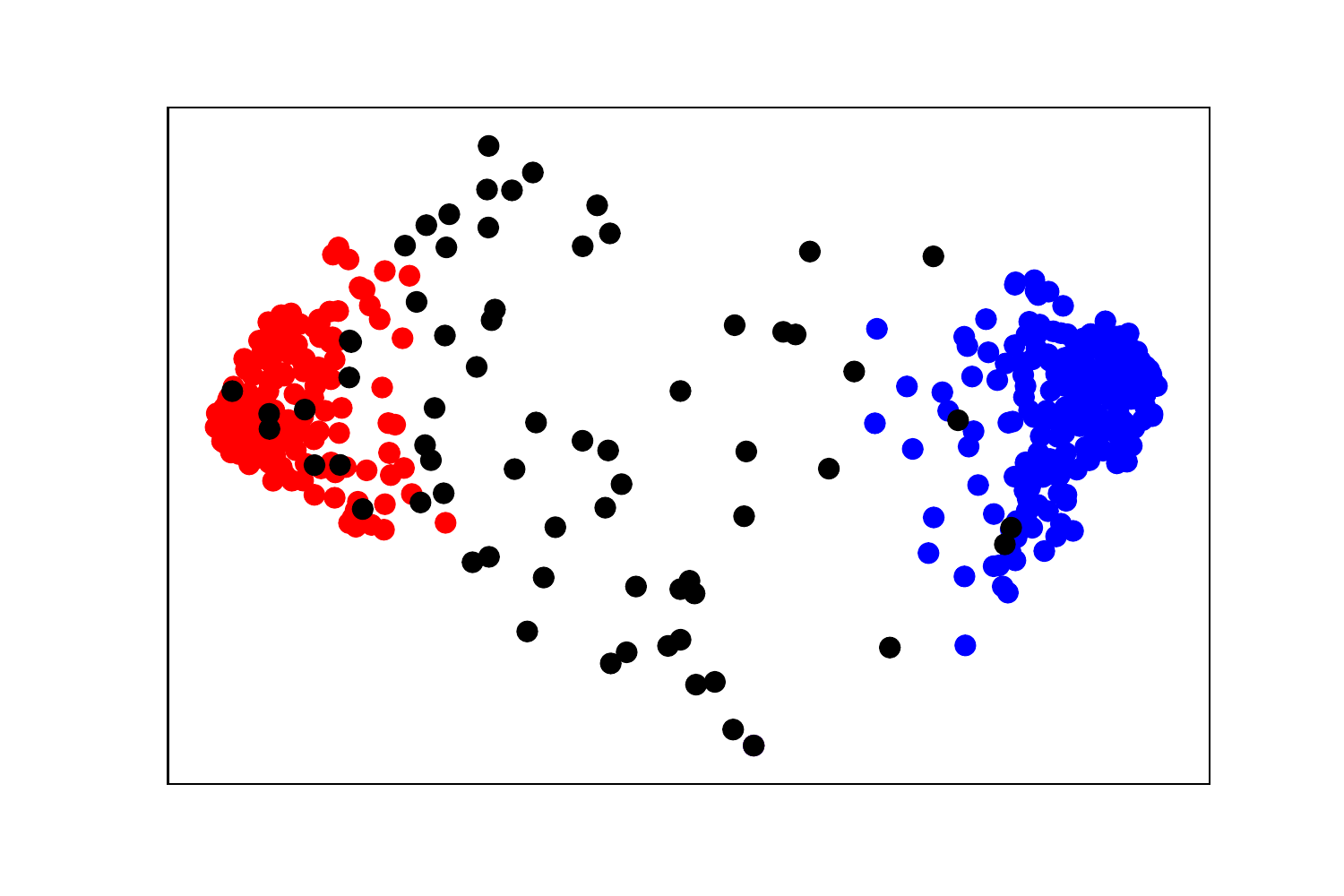}
     } &
\subfloat[Disputed 2]{%
       \includegraphics[scale=0.2]{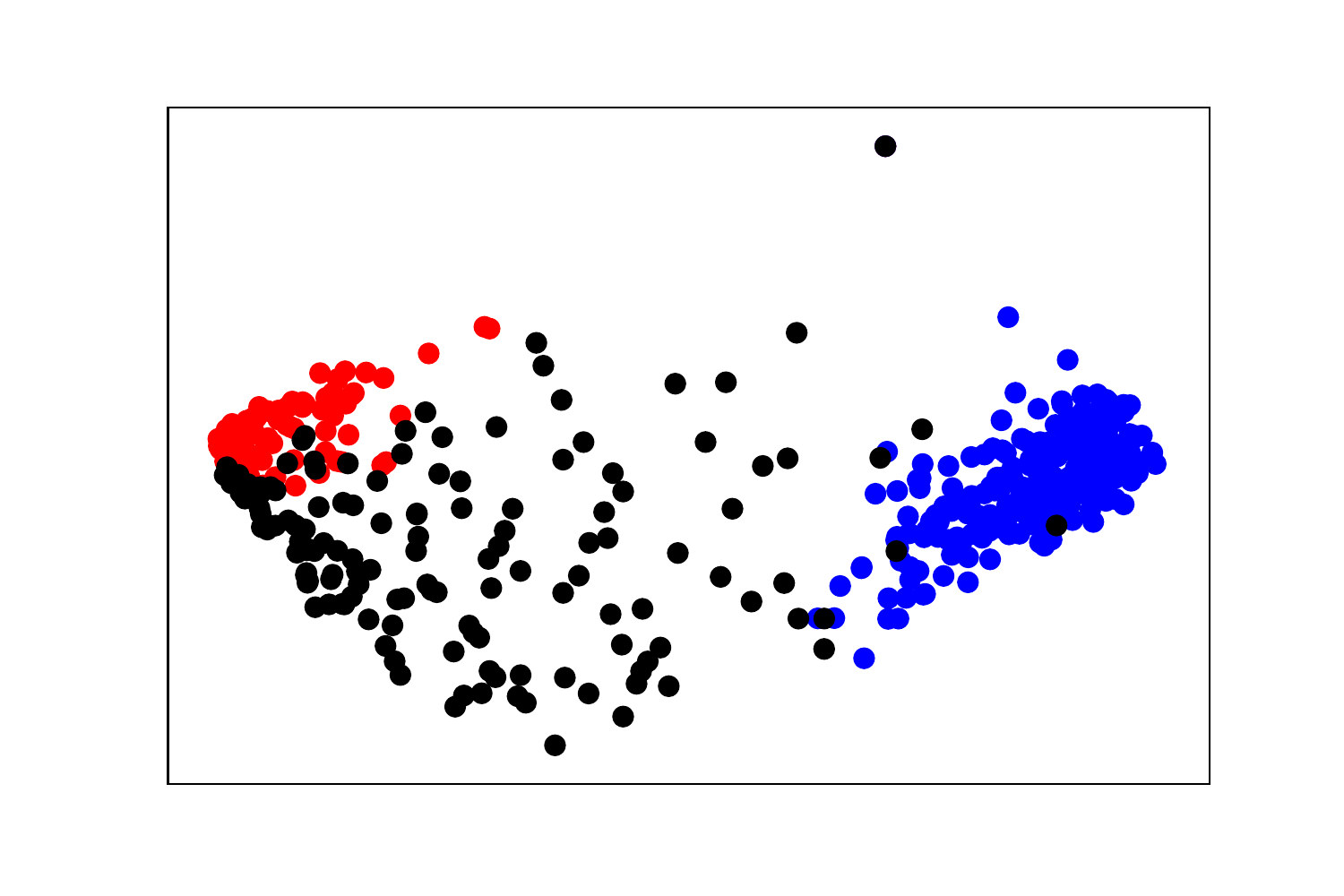}} &
\subfloat[Disputed 3]{%
       \includegraphics[scale=0.2]{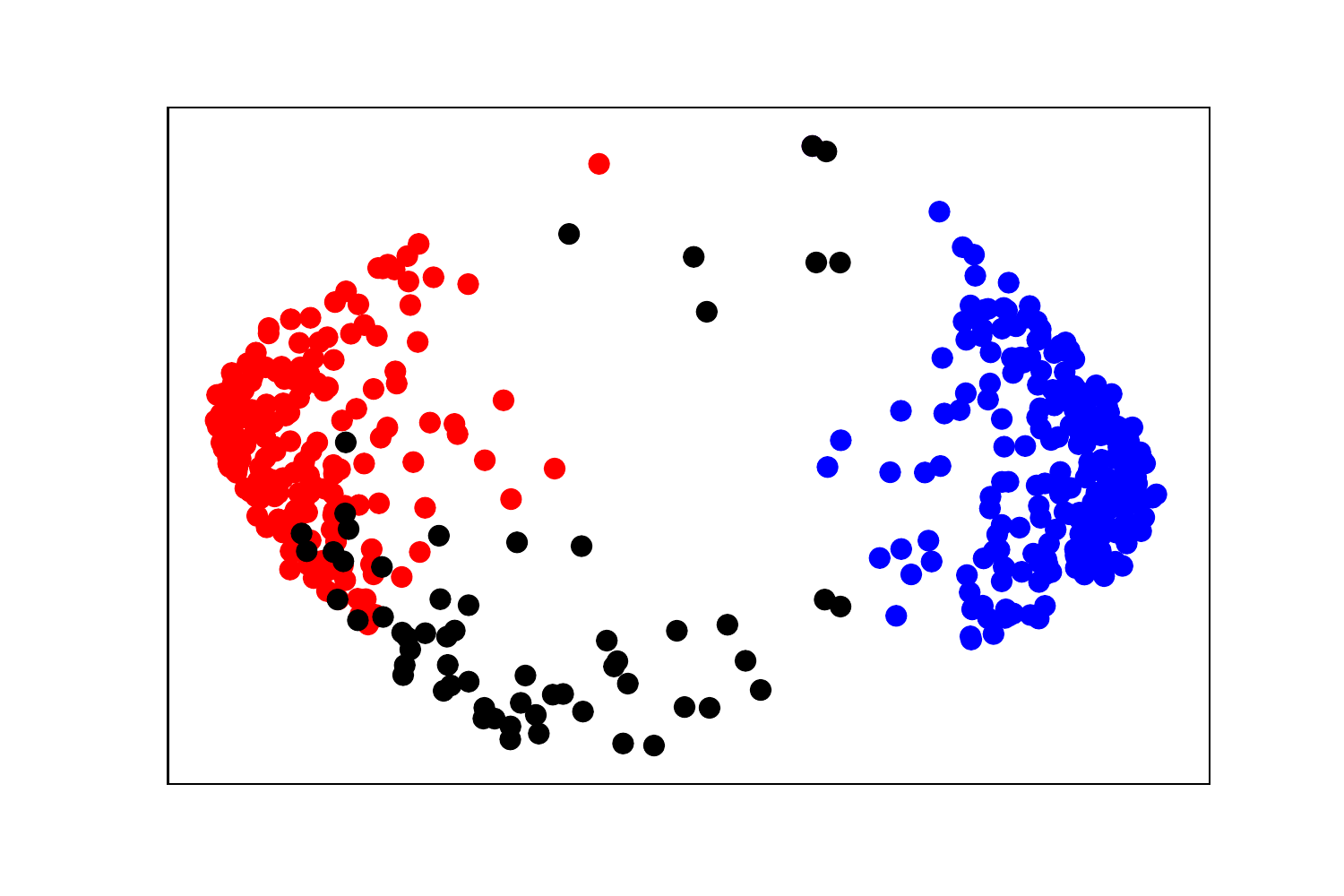}
     } &
\subfloat[Disputed 4]{%
       \includegraphics[scale=0.2]{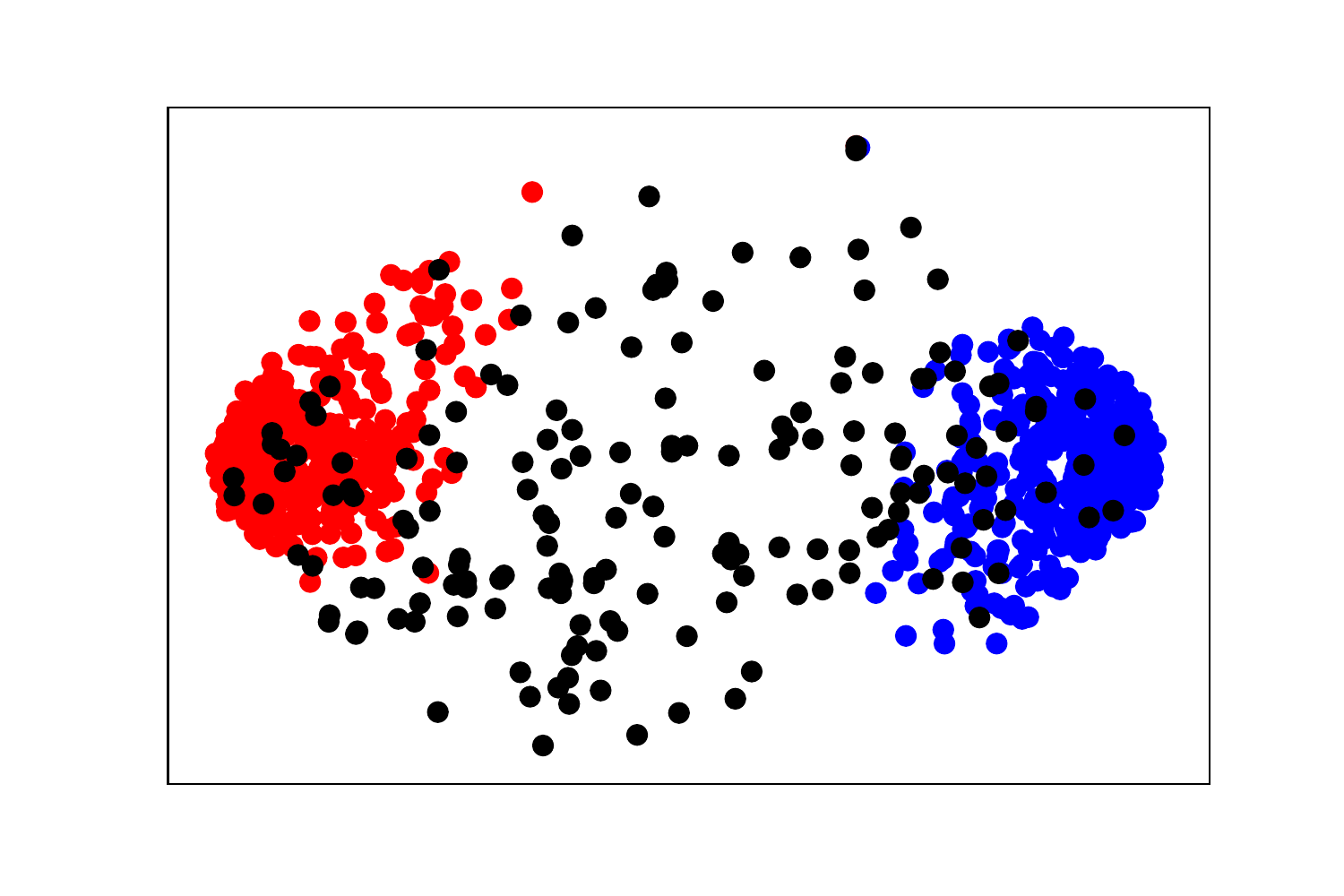}
     }\\
     \subfloat[Disputed 5]{%
       \includegraphics[scale=0.2]{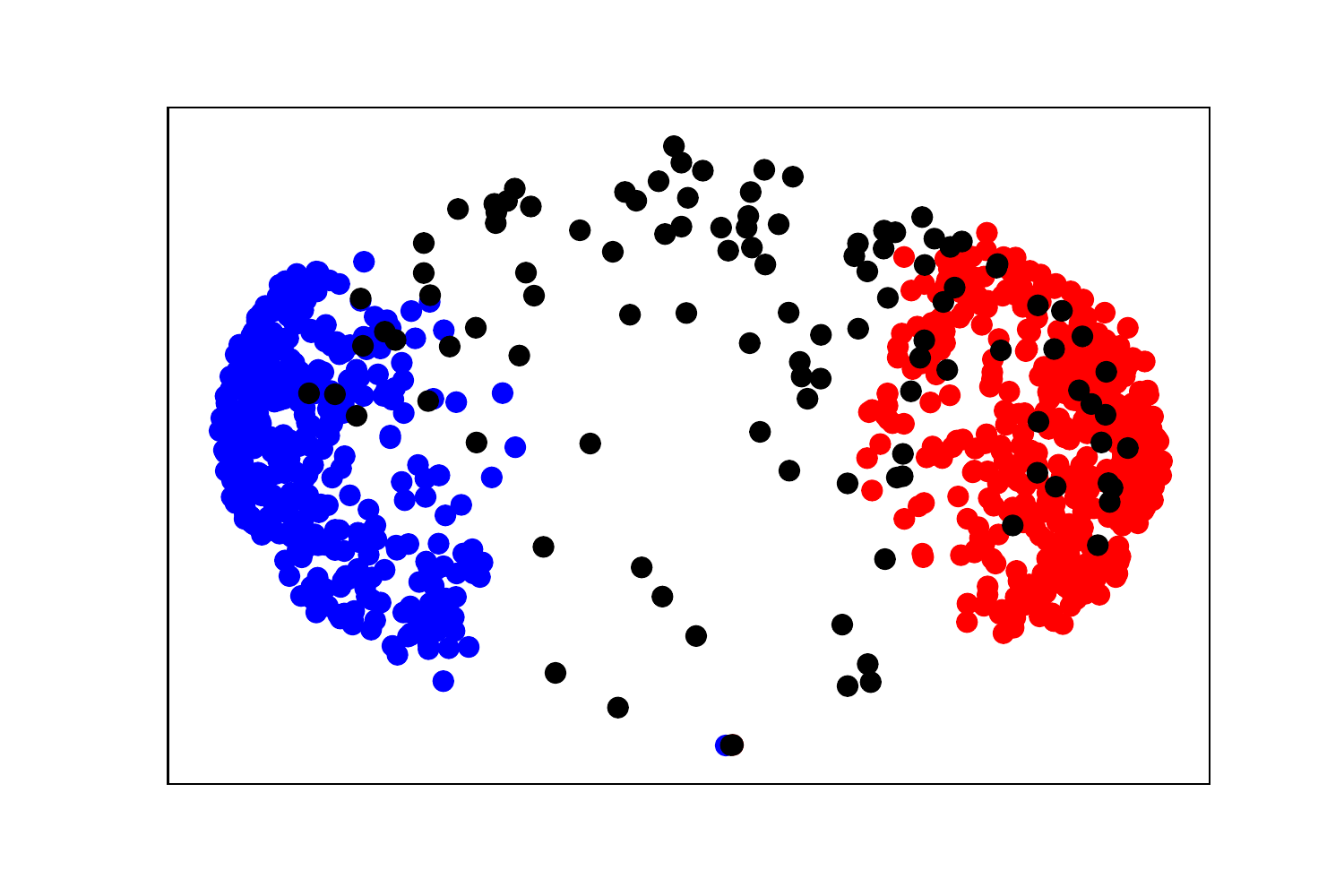}}&
\subfloat[Disputed 6]{%
       \includegraphics[scale=0.2]{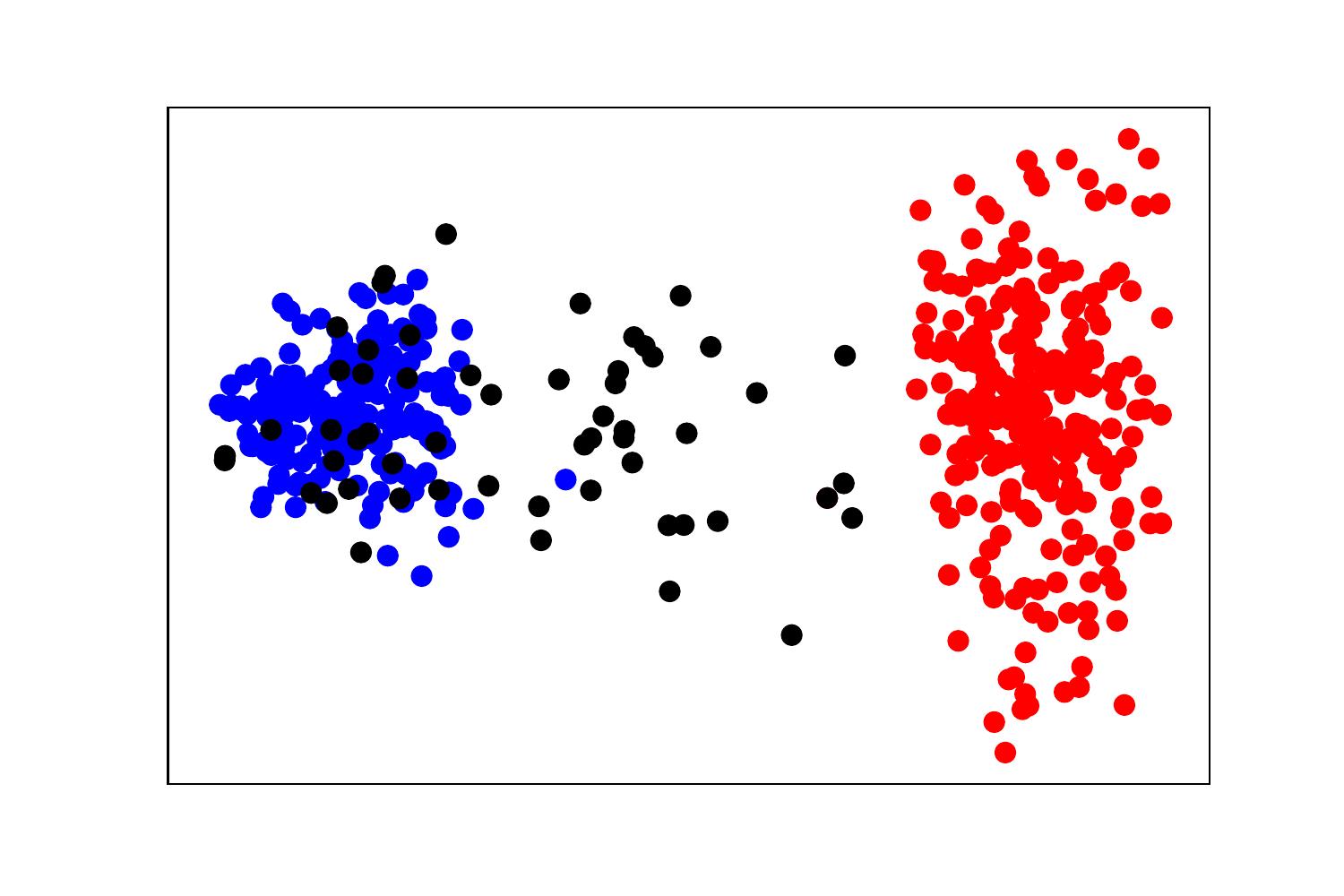}}&
\subfloat[Disputed 7]{%
       \includegraphics[scale=0.2]{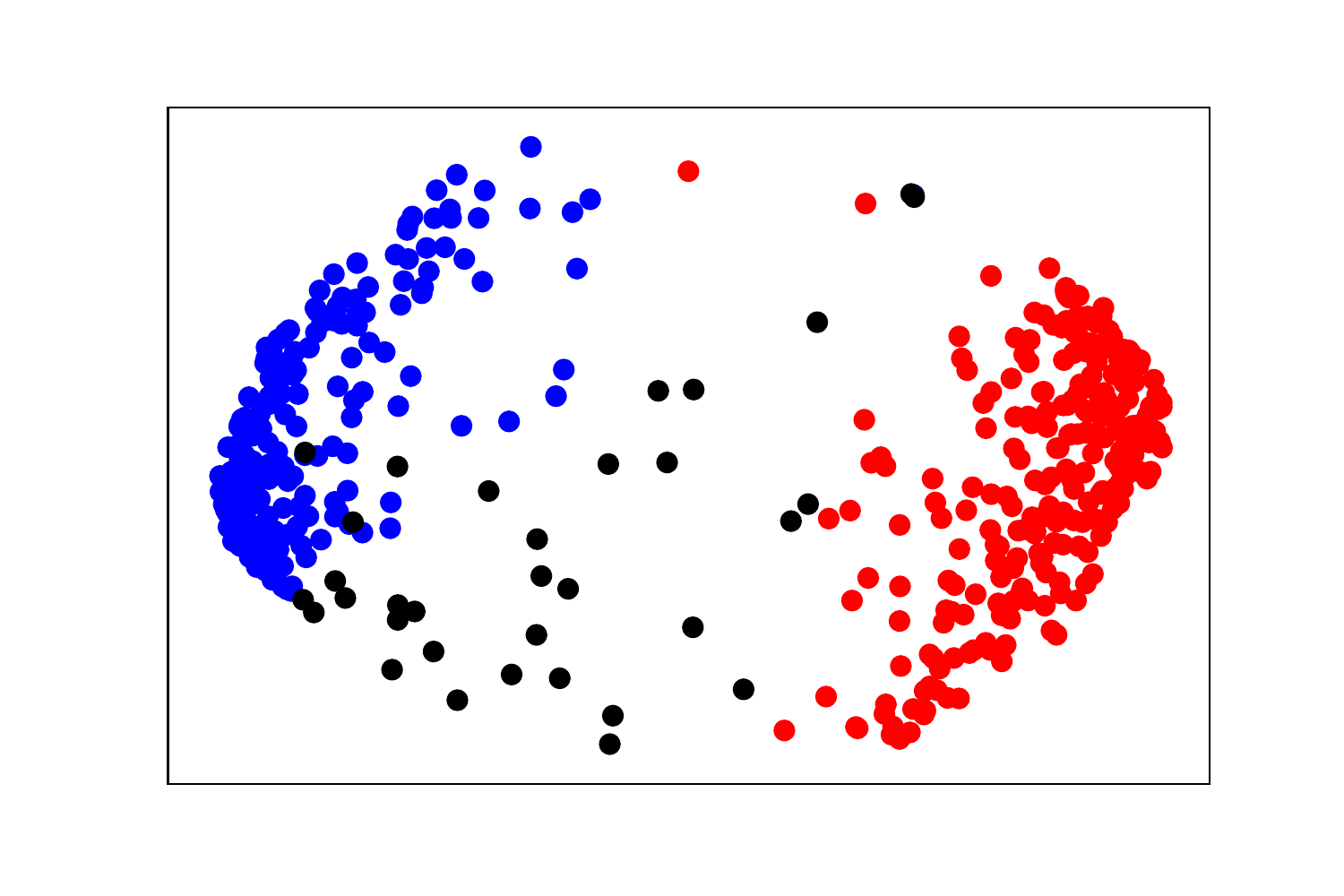}
     } &
\subfloat[Disputed 8]{%
       \includegraphics[scale=0.2]{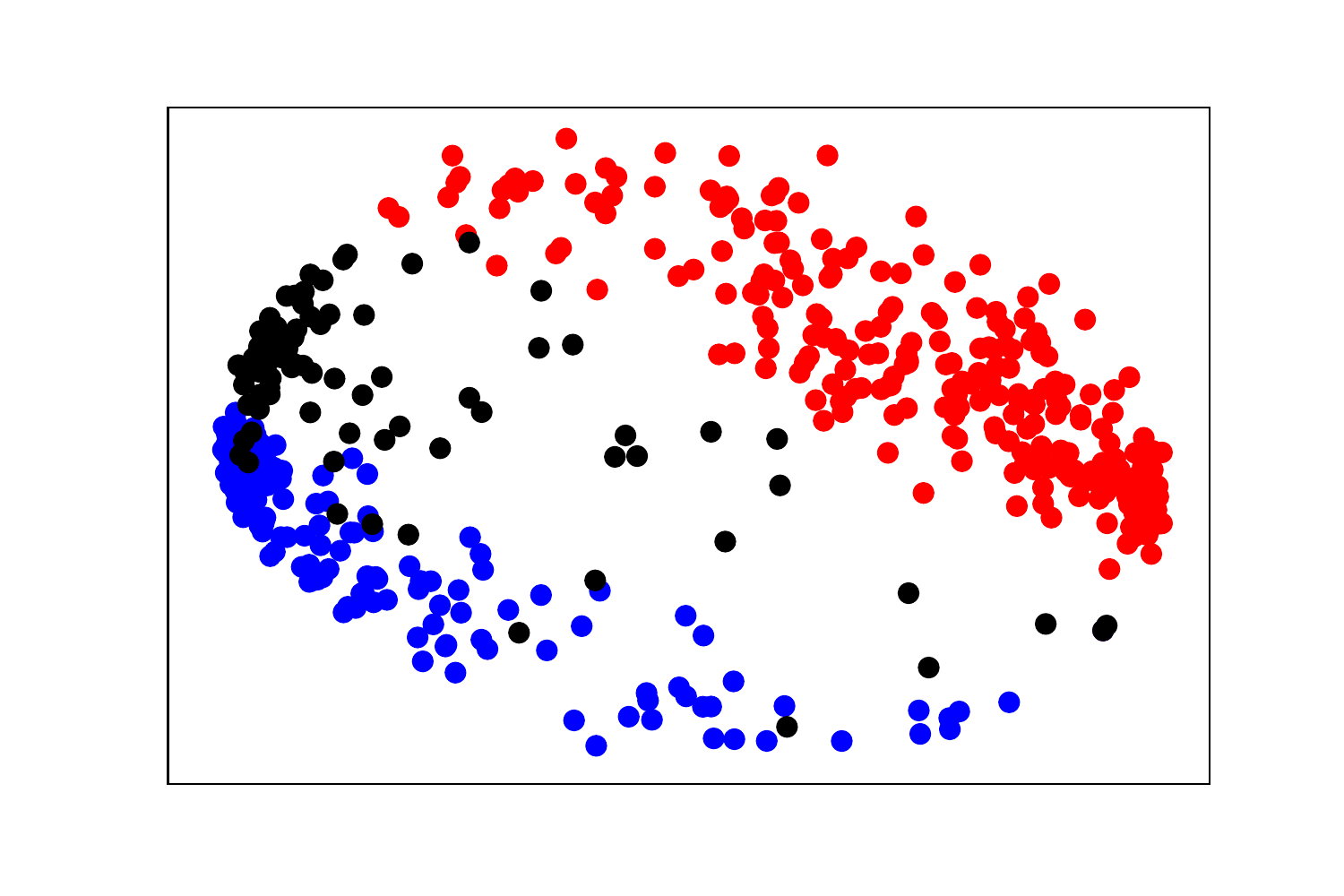}} \\
\subfloat[Disputed 9]{%
       \includegraphics[scale=0.2]{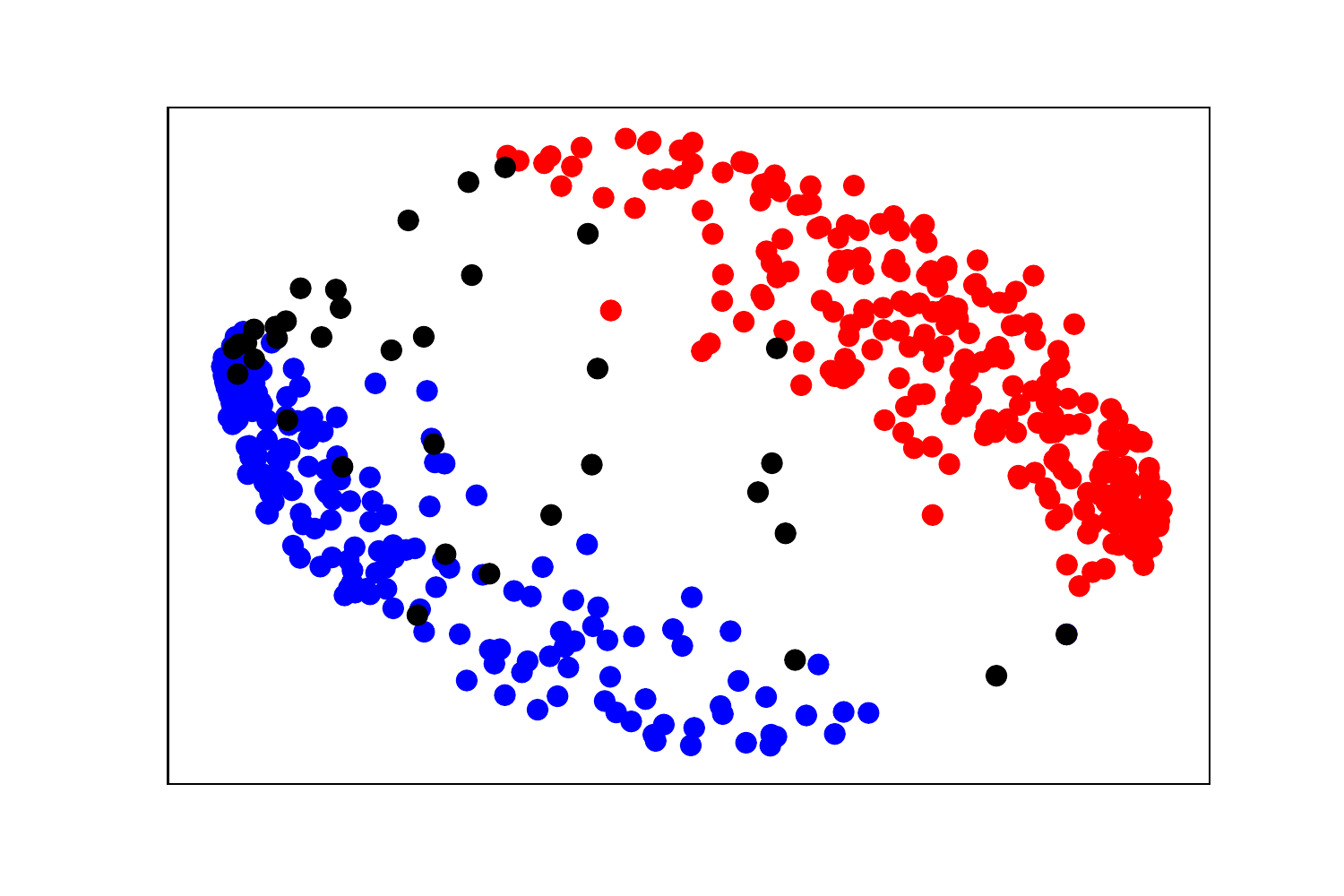}
     } &
\subfloat[Disputed 10]{%
       \includegraphics[scale=0.2]{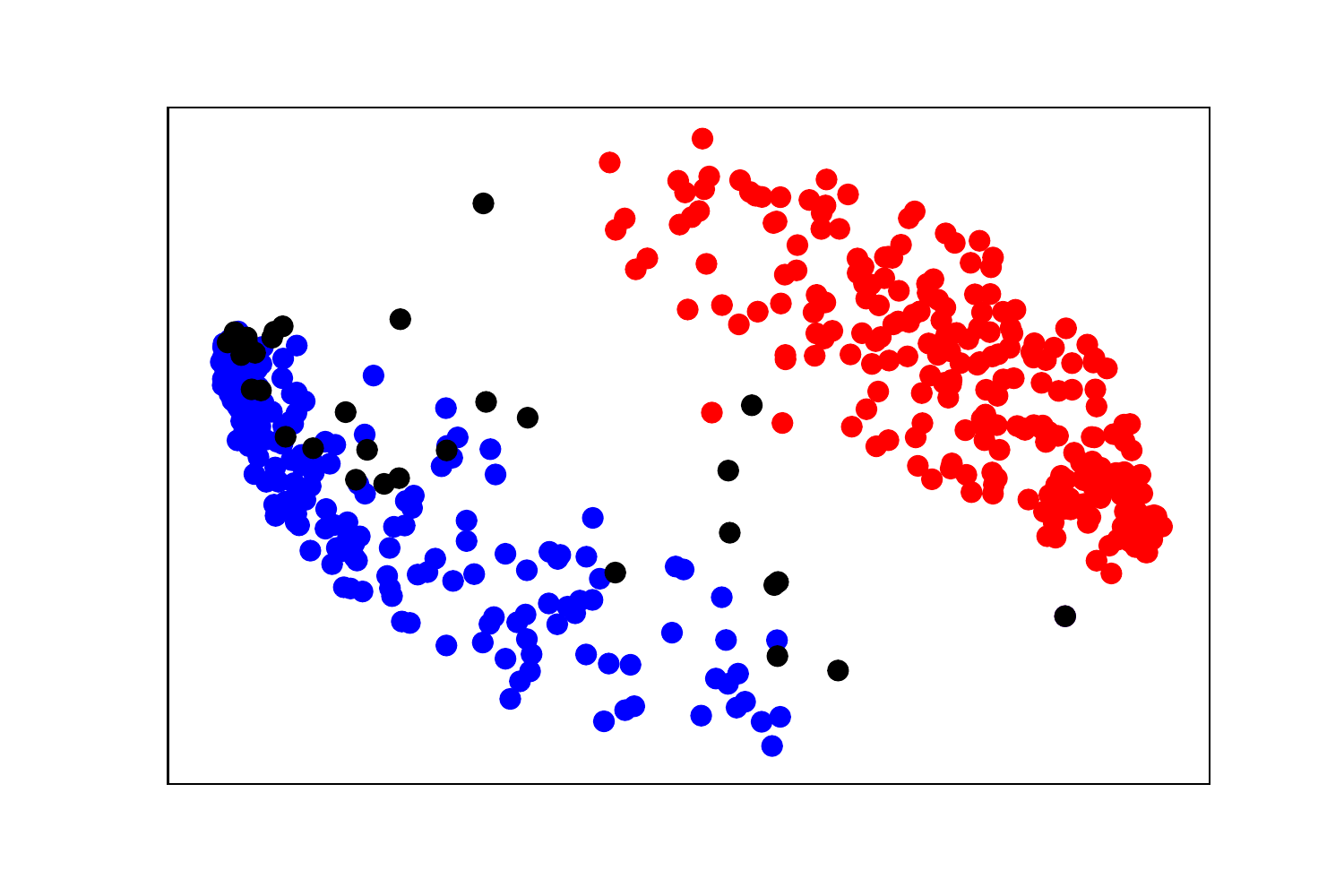}
     } &
\subfloat[Disputed 11]{%
       \includegraphics[scale=0.2]{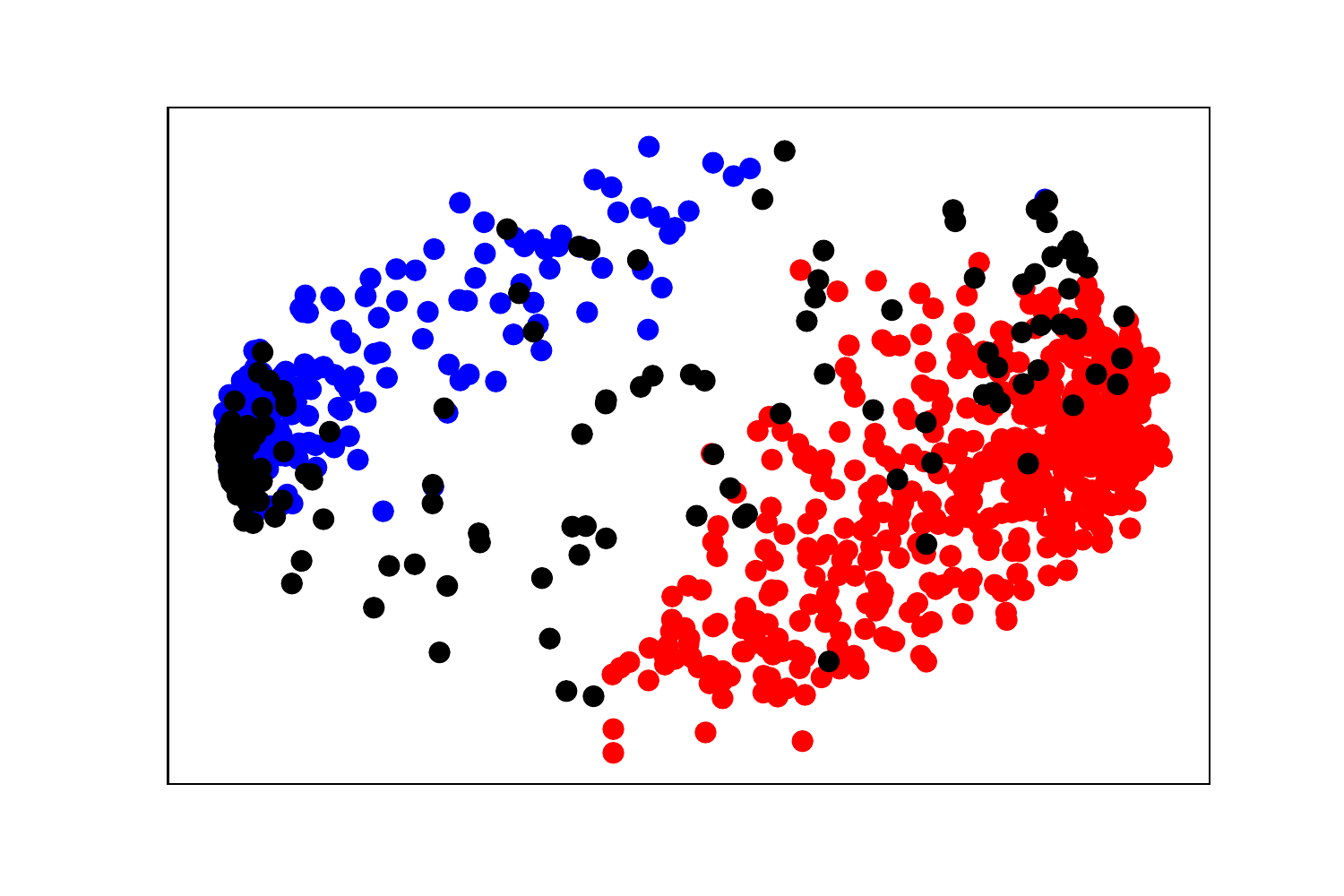}
     } \\
\end{tabular}
\caption{\textbf{First layer inspection for range 300-1300.} The Disputed id follows the mapping given in Figure \ref{fig:modern_cross_validation}. The plot is generated passing the text segments in the training set as input of the Neural Network and projecting the output of the first layer in a bi-dimensional space using PCA. We can see how the first layer learns to split the text segments in different groups: the red points represent the Russian-French group, while the blue dots represent the contemporary French-French group. The points in black are the text segments extracted from the Disputed text at that cross-validation round. We can see that the black points generally overlap with the points of the corresponding group. \label{fig:modern_first_layer_300_1300}}
\end{figure}

The Principal Component Analysis allows us to look at the points on the graph from a different angle, avoiding their overlap. We use PCA to visualize the hidden layer of the Neural Network in two dimensions. In most cases, the Network lets us see that the texts written by the same group of authors are grouped (red denotes the Russian-French group, and blue denotes the French-French group). We present graphs for the Neural Network trained on the ranges from 300 to 1300 words for the classic and contemporary authors in Figures~\ref{fig:first_layer_300_1300} and~\ref{fig:modern_first_layer_300_1300} respectively.

\newpage
\newpage
\section{Conclusion}

We conducted an experimental stylometric analysis using the following authorship attribution methods: Chi-squared, Ridge, SVM, KNN, and Neural Network. 

As a result, novels by French writers of Russian origins Andreï Makine, Valéry Afanassiev, Vladimir Fédorovski, Iegor Gran, Luba Jurgenson are similar in lexicon and differ from French ``classic" writers M. Proust, G. Flaubert, É. Zola, H. de Balzac, as well as from French ``contemporary" writers P. Modiano, P. Quignard, A. Ernaux, M. Duras, M. Houellebecq. This is due to the differences between them, although it is difficult to determine which ones. Perhaps there are some general similarities, invisible ``Russian traces", in the novels written in French by Russian-French authors. 

 We were able to discover one case of interference (the predominance of demonstratives over definite articles) in the texts by Russian-French bilingual authors. The demonstrative ``ces" tilted Ridge and SVM classifications in favor of the Russian-French group, and the frequency calculations revealed that the demonstratives were more frequent with the Russian-French authors, while the definite articles were more frequent with the French-French authors. Russian of bilingual authors influenced their writing in French. 
 
 This work opens the door to numerous future directions. In particular, it would be interesting to expand the corpus and to use more powerful computational resources to conduct the experiment. Future improvements could also be aimed at increasing the accuracy of author classification in addition to studying sporadic cases of interference. It is perhaps possible to detect syntactic interference using machine learning methods on texts pre-processed with part-of-speech tagging. 

The study of differences between styles of bilingual/multilingual and non-bilingual/non-multilingual authors can be useful in stylistics, in the study of the writer's vocabulary, in translation theory and practice, in literary studies, and in language teaching. 

\newpage
\bibliography{neurips}
\bibliographystyle{plain}

\end{document}